\documentclass[10pt,journal,compsoc]{IEEEtran}
\usepackage{amsfonts}
\usepackage{bm}

\ifCLASSINFOpdf
\else
\fi

\usepackage{amsmath, amssymb}
\usepackage{amsthm}
\usepackage{xcolor}

\usepackage{algorithm}
\usepackage{algorithmic}
\makeatletter
\newcommand\fs@blackruled{%
  \def\@fs@cfont{\bfseries}\let\@fs@capt\floatc@ruled
  \def\@fs@pre{\color{black}\hrule height.8pt depth0pt\kern2pt}%
  \def\@fs@post{\kern2pt\color{black}\hrule\relax}%
  \def\@fs@mid{\kern2pt\color{black}\hrule\kern2pt}%
  \let\@fs@iftopcapt\iftrue}
\floatstyle{blackruled}
\restylefloat{algorithm}
\makeatother

\usepackage{stfloats}

\usepackage{titletoc}
\usepackage{tikz}
\usetikzlibrary{positioning}
\usepackage{pgfplots}
\pgfplotsset{compat=1.18}
\usepackage[framemethod=tikz]{mdframed}
\usepackage{thm-restate}
\usepackage{booktabs}
\usepackage{tabularx}
\usepackage{placeins}
\usepackage{float}

\usepackage{colortbl}
\usepackage{hyperref}
\usepackage{subfigure} 
\usepackage{multirow}
\usepackage{ragged2e}
\usepackage{soul}
\usepackage{tcolorbox}
\usepackage[numbers,sort]{natbib}
\usepackage{setspace} 
\usepackage{adjustbox}
\newcommand{\fitTable}[1]{\begin{adjustbox}{max width=\linewidth}#1\end{adjustbox}}
\usepackage{enumitem}
\hypersetup{
	colorlinks=true,
	linkcolor=blue,
	urlcolor=orange,
	citecolor=orange,
}

\newtheorem{theorem}{Theorem}
\newtheorem{lemma}{Lem.}
\newtheorem{definition}{Definition}

\newtheorem{remark}{Remark}
\newtheorem{corollary}{Corollary}

\newtheorem{assumption}{Assump.}

\definecolor{Top2}{RGB}{102, 171, 221}
\definecolor{Top1}{RGB}{245, 137, 112}
\definecolor{DiagPastel}{RGB}{105, 151, 184}
\definecolor{OffDiagPastel}{RGB}{196, 126, 132}
\definecolor{BestCell}{RGB}{220, 237, 248}
\definecolor{RunnerCell}{RGB}{250, 237, 211}
\definecolor{GapZeroCell}{RGB}{236, 244, 250}
\definecolor{GapLowCell}{RGB}{233, 232, 247}
\definecolor{GapMidCell}{RGB}{249, 232, 213}
\definecolor{GapHighCell}{RGB}{247, 220, 216}
\definecolor{DefinitionBack}{RGB}{244, 240, 249}
\definecolor{DefinitionFrame}{RGB}{184, 166, 205}
\definecolor{TheoremBack}{RGB}{235, 244, 250}
\definecolor{TheoremFrame}{RGB}{142, 181, 207}
\definecolor{CorollaryBack}{RGB}{252, 244, 225}
\definecolor{CorollaryFrame}{RGB}{213, 181, 119}
\definecolor{RemarkBack}{RGB}{247, 247, 247}
\definecolor{RemarkFrame}{RGB}{178, 178, 178}
\definecolor{RemarkText}{RGB}{92, 92, 92}

\usepackage{pifont}
\newcommand{\cmark}{\ding{51}}%
\newcommand{\xmark}{\ding{55}}%
\usepackage{bbding}
\newcommand{\bmark}{\ding{51} \kern-1.7ex\raisebox{0.6ex}{\rotatebox[origin=c]{125}{\textbf{--}}}}

\tcbuselibrary{skins,breakable}
\tcolorboxenvironment{definition}{
  enhanced jigsaw, breakable, frame hidden, colback=DefinitionBack,
  boxrule=0pt, sharp corners, borderline west={2.2pt}{0pt}{DefinitionFrame},
  left=5pt, right=4pt, top=4pt, bottom=4pt,
  before skip=6pt, after skip=6pt}
\tcolorboxenvironment{theorem}{
  enhanced jigsaw, breakable, frame hidden, colback=TheoremBack,
  boxrule=0pt, sharp corners, borderline west={2.2pt}{0pt}{TheoremFrame},
  left=5pt, right=4pt, top=4pt, bottom=4pt,
  before skip=6pt, after skip=6pt}
\tcolorboxenvironment{lemma}{
  enhanced jigsaw, breakable, frame hidden, colback=TheoremBack,
  boxrule=0pt, sharp corners, borderline west={2.2pt}{0pt}{TheoremFrame},
  left=5pt, right=4pt, top=4pt, bottom=4pt,
  before skip=6pt, after skip=6pt}
\tcolorboxenvironment{assumption}{
  enhanced jigsaw, breakable, frame hidden, colback=TheoremBack,
  boxrule=0pt, sharp corners, borderline west={2.2pt}{0pt}{TheoremFrame},
  left=5pt, right=4pt, top=4pt, bottom=4pt,
  before skip=6pt, after skip=6pt}
\tcolorboxenvironment{proposition}{
  enhanced jigsaw, breakable, frame hidden, colback=TheoremBack,
  boxrule=0pt, sharp corners, borderline west={2.2pt}{0pt}{TheoremFrame},
  left=5pt, right=4pt, top=4pt, bottom=4pt,
  before skip=6pt, after skip=6pt}
\tcolorboxenvironment{corollary}{
  enhanced jigsaw, breakable, frame hidden, colback=CorollaryBack,
  boxrule=0pt, sharp corners, borderline west={2.2pt}{0pt}{CorollaryFrame},
  left=5pt, right=4pt, top=4pt, bottom=4pt,
  before skip=6pt, after skip=6pt}
\tcolorboxenvironment{remark}{
  enhanced jigsaw, breakable, frame hidden, colback=RemarkBack, coltext=RemarkText,
  boxrule=0pt, sharp corners, borderline west={2.2pt}{0pt}{RemarkFrame},
  left=5pt, right=4pt, top=4pt, bottom=4pt,
  before skip=6pt, after skip=6pt}
\newtcolorbox{theoremstatementbox}{
  enhanced jigsaw, breakable, frame hidden, colback=TheoremBack,
  boxrule=0pt, sharp corners, borderline west={2.2pt}{0pt}{TheoremFrame},
  left=5pt, right=4pt, top=4pt, bottom=4pt,
  before skip=6pt, after skip=6pt}
\newtcolorbox{corollarystatementbox}{
  enhanced jigsaw, breakable, frame hidden, colback=CorollaryBack,
  boxrule=0pt, sharp corners, borderline west={2.2pt}{0pt}{CorollaryFrame},
  left=5pt, right=4pt, top=4pt, bottom=4pt,
  before skip=6pt, after skip=6pt}

\newenvironment{qbox-tight}
	{\begin{tcolorbox}[enhanced jigsaw, drop shadow=black!50!white,colback=white, width=\linewidth, center, left=2pt,right=2pt,top=1pt,bottom=1pt,halign=center]}
	{\end{tcolorbox}}

\usepackage{listings}
\usepackage{color}
\definecolor{dkgreen}{rgb}{0,0.6,0}
\definecolor{gray}{rgb}{0.5,0.5,0.5}
\definecolor{mauve}{rgb}{0.58,0,0.82}
\begin{document}
  \title{Semantic Mismapping in Generalized Category Discovery: Evaluation and Mitigation}

\author{Zitai~Wang,
        Qianqian~Xu*,~\IEEEmembership{Senior ~Member,~IEEE,}
        Cong Hua, 
        Xi Chen,
        Zhiyong~Yang, \\
        Xiaochun~Cao,~\IEEEmembership{Senior ~Member,~IEEE,}
        and~Qingming~Huang*,~\IEEEmembership{Fellow,~IEEE}%
        \IEEEcompsocitemizethanks{
            \IEEEcompsocthanksitem Zitai Wang is with the State Key Laboratory of AI Safety, Institute of Computing Technology, Chinese Academy of Sciences, Beijing 100190, China (email: \texttt{wangzitai@ict.ac.cn}). 
            \IEEEcompsocthanksitem Qianqian Xu is with the State Key Laboratory of AI Safety, Institute of Computing Technology, Chinese Academy of Sciences, Beijing 100190, China, and also with Beijing Academy of Artificial Intelligence (BAAI), Beijing 100083, China (email: \texttt{xuqianqian@ict.ac.cn}). 
            \IEEEcompsocthanksitem Cong Hua is with the State Key Laboratory of AI Safety, Institute of Computing Technology, Chinese Academy of Sciences, Beijing 100190, China, and also with the School of Computer Science and Technology, University of Chinese Academy of Sciences, Beijing 101408, China (email: \texttt{huacong23z@ict.ac.cn}).
            \IEEEcompsocthanksitem Xi Chen is with the State Key Laboratory of AI Safety, Institute of Computing Technology, Chinese Academy of Sciences, Beijing 100190, China, and also with the School of Computer Science and Technology, University of Chinese Academy of Sciences, Beijing 101408, China (email: \texttt{chenxi26@mails.ucas.ac.cn}).
            \IEEEcompsocthanksitem Zhiyong Yang is with the School of Computer Science and Technology, University of Chinese Academy of Sciences, Beijing 100049, China (email: \texttt{yangzhiyong21@ucas.ac.cn}).
            \IEEEcompsocthanksitem Xiaochun Cao is with the School of Cyber Science and Technology, Shenzhen Campus of Sun Yat-sen University, Shenzhen 518107, China (email: \texttt{caoxiaochun@mail.sysu.edu.cn}).
            \IEEEcompsocthanksitem Qingming Huang is with the School of Computer Science and Technology, University of Chinese Academy of Sciences, Beijing 101408, China, and also with the State Key Laboratory of AI Safety, Institute of Computing Technology, Chinese Academy of Sciences, Beijing 100190, China (email: \texttt{qmhuang@ucas.ac.cn}).
            \IEEEcompsocthanksitem *Corresponding authors.
        }
}

\markboth{}%
{Wang \MakeLowercase{\textit{et al.}}: Semantic Mismapping in GCD}

\IEEEtitleabstractindextext{%
    \begin{abstract}
        \justifying
        Generalized Category Discovery (GCD) aims to perform old-category recognition and novel-category discovery in the same unlabeled set. Early methods address both objectives through clustering and evaluate the exchangeable predicted indices with global Hungarian (GH) matching. Recent methods adopt trainable parametric heads, using supervision for old-category recognition and discovery-oriented objectives for new categories. In this way, the output space becomes partially exchangeable: old indices are semantically anchored, while new indices remain exchangeable. However, these methods still apply global matching, which can reassign anchored old indices to other categories and hide some prediction errors. Empirically, we are surprised to observe that this reassignment can induce an overestimation of new-category accuracy by up to 25.63 points. We term this phenomenon \emph{semantic mismapping} and further derive a permutation-cycle condition that characterizes when semantic mismapping can increase the global matching score. To fix this issue, we propose Semantics-Preserving Evaluation (SPE), which preserves the supervised semantics of old indices and restricts Hungarian matching to the new region. Furthermore, we propose Mismapping-Mitigating Optimization (MMO) to alleviate the prediction errors that can give rise to semantic mismapping, combining a region-balance loss (RBL) for old/new routing with a novel-diversity loss (NDL) that encourages orthogonality among new prototypes. Theoretically, we show that minimizing the two losses can tighten an upper bound on spurious accuracy gains from semantic mismapping, with controllable residual terms. Empirical results under both GH and SPE demonstrate the effectiveness of our proposed method.
    \end{abstract}
    
}
 
  \maketitle
  \IEEEdisplaynontitleabstractindextext
  \IEEEpeerreviewmaketitle

  \IEEEraisesectionheading{\section{Introduction}\label{sec:introduction}}
\IEEEPARstart{T}{raditional} recognition tasks typically follow a closed-world assumption: the categories encountered at test time are covered by the labeled training set. However, in real-world scenarios, a deployed recognition system may receive samples from known categories together with samples from categories that have never been labeled before~\cite{hendrycks_ood_2017,xu_interneg_2026}. In view of this, Generalized Category Discovery (GCD) considers a more challenging open-world setting, where a model is trained with labeled samples from old categories and unlabeled samples drawn from both old and new categories \cite{DBLP:conf/cvpr/VazeHVZ22}. This setting gives rise to two objectives that typically belong to different learning paradigms: \textbf{(O1)} recognizing old categories in the unlabeled samples (\textit{i.e.}, classification) and \textbf{(O2)} discovering underlying clusters for new categories (\textit{i.e.}, clustering). Hence, it becomes a core question in GCD: \textit{how to unify the two distinct objectives into a single framework?}

\begin{figure*}[t]
    \centering
    \includegraphics[width=\textwidth]{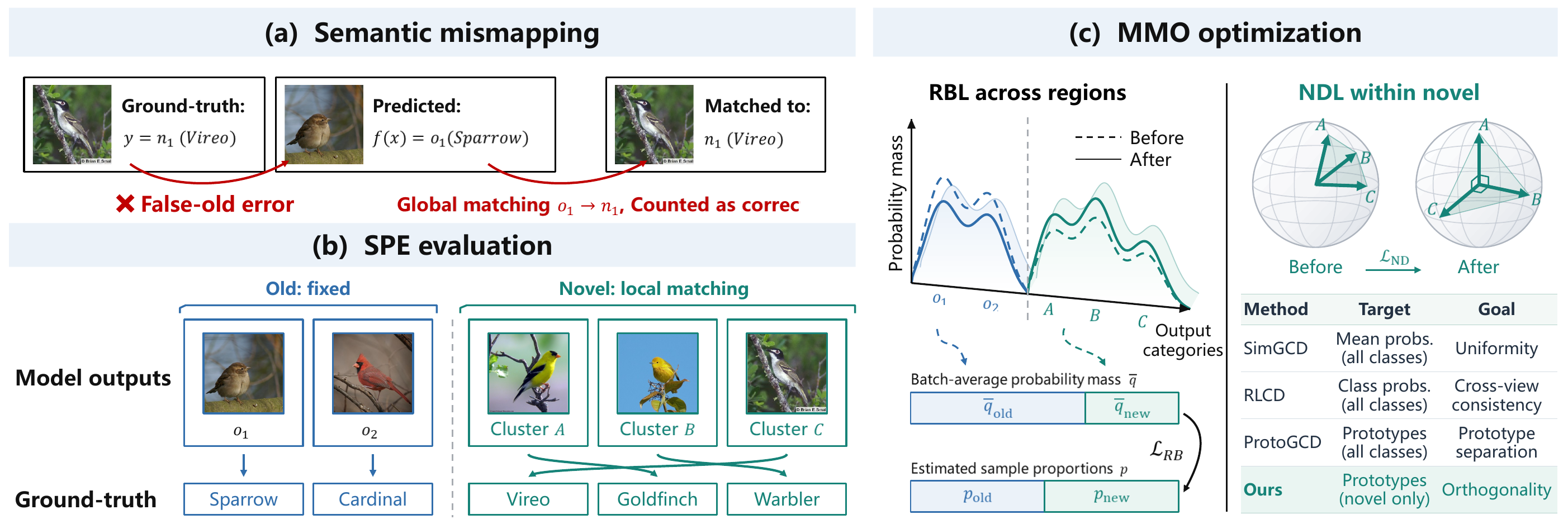}
    \caption{\textbf{(a)} When evaluating parametric GCD methods, global matching could reassign an old output to a new category (\textit{e.g.}, $o_1 \to n_1$), making a false-old prediction appear correct. \textbf{(b)} Our SPE protocol {preserves the semantics of old outputs} and performs local matching within the new region. \textbf{(c)} {Our mismapping-mitigating optimization (MMO) combines region-balance loss (RBL), which matches old/new prediction mass to estimated sample proportions, with novel-diversity loss (NDL), which encourages orthogonality among new prototypes.}}
    \label{fig:main_overview}
\end{figure*}

Early GCD methods adopt a unified clustering-based framework, learning visual features from both labeled and unlabeled data and then applying non-parametric clustering \cite{DBLP:conf/cvpr/VazeHVZ22,DCCL,PromptCAL}. The induced cluster indices are exchangeable as their semantics have been lost. Hence, the standard GCD evaluation protocol aligns these indices with ground-truth categories through Hungarian matching \cite{kuhn1955hungarian} and then calculates accuracy.

Recent advances, represented by SimGCD \cite{DBLP:conf/iccv/WenZQ23} and its follow-ups \cite{DBLP:conf/nips/VazeVZ23,DBLP:conf/cvpr/CaoZWYS0L024,wang2025get}, turn to a divide-and-conquer parametric framework, achieving superior performance. These methods replace the non-parametric clustering procedure with trainable classifier or prototype heads. For old categories (\textbf{O1}), the parametric head is effectively trained with the supervision from labeled samples. For new categories (\textbf{O2}), the corresponding indices are not tied to fixed categories and are trained with discovery-oriented objectives, such as self-distillation \cite{DBLP:conf/iccv/WenZQ23}, pseudo-labeling \cite{DBLP:conf/nips/VazeVZ23}, and contrastive learning \cite{DBLP:conf/cvpr/ChoiKC24}. Overall, the output space becomes partially exchangeable: old indices are semantically anchored, while new indices remain exchangeable.

Although effective, this divide-and-conquer framework can exhibit notorious \textit{prediction bias}. Since old categories receive more direct supervision, the parametric head tends to produce an imbalanced distribution across old and new categories \cite{DBLP:conf/iccv/WenZQ23}. A typical failure is \textit{false-old}, where many new samples are predicted as old categories. To address this bias, prior work has designed a series of training objectives and regularizers \cite{DBLP:conf/iccv/WenZQ23,DBLP:conf/cvpr/CaoZWYS0L024,ma2025protogcd,liu2025generalized}.

Despite their progress, parametric methods still apply Hungarian matching to the entire output space. However, we find that such global matching can reassign semantically anchored old indices to new categories, making some false-old predictions appear correct. {As illustrated in Fig.\ref{fig:main_overview}(a), a new bird with an incorrect old-category prediction is reassigned to its ground-truth new category.} In view of this, we call any assignment that overrides an anchored index \textit{semantic mismapping}, which may occur both across the old/new boundary and within the old region. Theoretically, we further derive a permutation-cycle condition that characterizes when semantic mismapping can increase the global matching score. For fine-grained datasets, such as Herbarium 19, this mismapping can induce an overestimation of new accuracy by 25.63 points, whereas for coarse-grained datasets, such as CIFAR-10 and ImageNet-100, its effect becomes negligible. This observation raises a natural question:

\begin{qbox-tight}
How should parametric GCD methods be evaluated to avoid semantic mismapping?
\end{qbox-tight}

To answer this question, we propose a divide-and-conquer protocol named \textbf{Semantics-Preserving Evaluation (SPE)}. {As illustrated in Fig.\ref{fig:main_overview}(b), SPE preserves the semantics of old outputs and performs local matching within the new region, while predictions crossing the old/new boundary remain errors.} This design directly measures old-category recognition for \textbf{(O1)} while preserving the permutation invariance required for novel-category discovery under \textbf{(O2)}. We further extend SPE to non-parametric GCD methods and settings with an estimated number of new categories.

Motivated by our analysis, we propose Mismapping-Mitigating Optimization (MMO), which combines two plug-and-play regularizers following the divide-and-conquer principle, as summarized in Fig.\ref{fig:main_overview}(c). On the one hand, a region-balance loss (RBL) explicitly balances prediction mass across the old and new regions. On the other hand, a novel-diversity loss (NDL) promotes new-category discrimination by encouraging orthogonality among new prototypes. We further derive two bounds showing how the two losses can suppress spurious accuracy gains when residual errors are controlled. These bounds provide sufficient conditions for GH and SPE to yield the same overall accuracy. Finally, extensive experiments on several challenging fine-grained benchmarks validate the necessity of SPE and the effectiveness of the proposed regularizers. Overall, our contributions are summarized as follows:
\begin{itemize}
    \item \textbf{New insight.} We identify semantic mismapping in parametric GCD and theoretically characterize when it produces spurious accuracy gains under GH.
    \item \textbf{New protocol.} Building on this insight, we propose SPE to prevent semantic mismapping from hiding prediction errors, enabling faithful evaluation of old-category recognition and novel-category discovery.
    \item \textbf{New method.} To mitigate the underlying prediction errors, we propose MMO with two plug-and-play losses, supported by theoretical bounds and extensive experiments.
\end{itemize}

\section{Related Work}
\label{sec:related_work}

\subsection{Category-Discovery Settings}
Category discovery aims to identify previously unseen categories in unlabeled data by transferring knowledge from labeled old categories \cite{DBLP:conf/iclr/HsuLK18,DBLP:conf/iclr/HsuLSOK19,DBLP:conf/cvpr/ZhongZLL0S21}.

As one of the most representative settings, Novel Category Discovery (NCD) provides labeled old-category data alongside an unlabeled discovery set containing only new categories \cite{DBLP:conf/iccv/HanVZ19,DBLP:conf/iclr/HanREVZ20,DBLP:conf/iclr/ChiLYLL00ZS22,DBLP:conf/cvpr/YangWDZ23,DBLP:conf/cvpr/LiuCJQ0P24}. Mainstream NCD methods combine self-supervised learning with old-category supervision to improve the clustering of unlabeled samples \cite{DBLP:conf/iclr/HanREVZ20,zhong2021neighborhood,DBLP:conf/iccv/JiaHZG21,DBLP:journals/pami/HanREVZ22,DBLP:conf/nips/ZhaoH21}. 
Class-incremental NCD extends this setting by first learning on labeled old-category data and then receiving unlabeled data for novel-category discovery \cite{joseph_ncdwf_2022,roy_class_incremental_ncd_2022}.
Generalized Category Discovery (GCD) is a more challenging setting, where old and new categories coexist in the unlabeled set \cite{DBLP:conf/cvpr/VazeHVZ22,DBLP:journals/corr/abs-2305-10420}. 
Subsequent studies investigate fine-grained discovery \cite{DBLP:conf/bmvc/FeiZYZ22,rastegar_selex_2024,zheng_generalized_2025}, domain shifts \cite{wang_hilo_2025}, and continuous discovery \cite{kim_proxy_2023}. OpenGCD integrates generalized category discovery into open-world recognition~\cite{wang_openauc_2022,hua_openworldauc_2025} to assist in grouping unlabeled samples for subsequent incremental learning~\cite{gao_opengcd_2023}. 

This paper focuses on the standard GCD setting, studying how to faithfully evaluate and jointly improve old-category recognition and novel-category discovery.

\subsection{Evaluation Protocols for Category Discovery}
\label{sec:related_evaluation_protocols}
Evaluation protocols for the aforementioned settings mainly differ in how they emphasize old-category recognition and novel-category discovery. Early NCD studies focus on clustering accuracy for new categories \cite{DBLP:conf/iccv/HanVZ19,DBLP:conf/iclr/HanREVZ20}. Later NCD studies evaluate both old and new categories, distinguishing task-aware and task-agnostic protocols according to whether the model is told if each sample belongs to an old or a new category \cite{DBLP:conf/iccv/FiniSLZN021,yang_comex_2022,DBLP:conf/cvpr/LiFHG23}. Class-incremental NCD also measures how well old-category recognition is maintained as the model learns new categories \cite{joseph_ncdwf_2022,roy_class_incremental_ncd_2022}. Notably, FRoST also identifies old--new errors hidden by Hungarian matching and preserves old-category semantics during NCD evaluation \cite{roy_class_incremental_ncd_2022}. However, its new-category mapping is established on new-only training data, leaving evaluation on the mixed unlabeled set in GCD unaddressed.

To handle the coexistence of old and new categories, GCD extends NCD evaluation by applying a global Hungarian matching to the whole unlabeled set \cite{DBLP:conf/cvpr/VazeHVZ22}. This clustering-based protocol remains widely used by parametric GCD methods, providing a common basis for comparison with non-parametric approaches \cite{DBLP:conf/iccv/WenZQ23,ma2025protogcd}. $\mu$GCD also observes that Hungarian matching can report nonzero new-category accuracy even when no samples are predicted to new outputs \cite{DBLP:conf/nips/VazeVZ23}. However, its evaluation still follows global matching, without preserving anchored old-output semantics. We systematically study this mismatch in standard GCD, characterize matching gains through permutation cycles, quantify their effects on empirical performance and method rankings, and propose SPE for faithful evaluation.

Recent work also highlights the distinction between clustering accuracy and semantic classification. SECOS uses category names and vision-language knowledge to classify both old and new categories, without post-hoc matching \cite{liu2026secos}. In standard GCD, however, new categories are expected to be discovered without their names being given. Hence, our work focuses on evaluation in the standard GCD setting.

\subsection{GCD Methods and Prediction Bias}
Early GCD approaches learn transferable features from both labeled and unlabeled data, followed by non-parametric clustering \cite{DBLP:conf/cvpr/VazeHVZ22,DCCL,PromptCAL}. Further studies improve feature adaptation through spatial prompts or textual knowledge \cite{DBLP:conf/iclr/WangV024,textgcd,wang2025get}. Parametric methods replace non-parametric clustering with trainable classifiers or prototypes \cite{DBLP:conf/iccv/WenZQ23,DBLP:conf/nips/VazeVZ23,ma2025protogcd}. This divide-and-conquer paradigm achieves state-of-the-art performance, as the parametric classifier leverages old-category supervision more effectively.

However, this parametric paradigm can also lead to prediction bias, as old and new categories rely on different learning signals. Direct supervision can bias predictions toward old categories \cite{DBLP:conf/iccv/WenZQ23}, while overconfidence in new categories can also hinder old-category recognition accuracy \cite{guo2025degcd}. To mitigate this bias, existing methods apply constraints at three levels. At the prediction level, existing methods use entropy and class-wise regularization \cite{DBLP:conf/iccv/WenZQ23,liu2025generalized}, consistency constraints \cite{DBLP:conf/cvpr/CaoZWYS0L024}, or auxiliary debiased classifiers \cite{liu2025debgcd,guo2025degcd} to balance old- and new-category predictions. At the feature level, existing methods use prototype separation \cite{ma2025protogcd}, feature-geometry regularization \cite{tang2026bia}, or relational distillation \cite{liu2026closer} to improve category discrimination. At the gradient level, EAGC coordinates supervised and unsupervised gradients to reduce conflicts between old-category recognition and novel-category discovery \cite{zheng2026eagc}.

Our work instead examines prediction bias at the evaluation level, revealing how global matching can hide the resulting errors. Guided by this insight, we design regularizers at the prediction and feature levels.

\section{Preliminaries}
\label{sec:preliminaries}

\subsection{Generalized Category Discovery}
Let $\mathcal{D}_l=\{(\boldsymbol{x}_i,y_i)\}_{i=1}^{N_l}$ and $\mathcal{D}_u=\{\boldsymbol{x}_j\}_{j=1}^{N_u}$ denote the labeled and unlabeled training sets, respectively. The labels in $\mathcal{D}_l$ belong to a set of old categories $\mathcal{Y}_{\mathrm{old}}$, whereas $\mathcal{D}_u$ contains samples from both $\mathcal{Y}_{\mathrm{old}}$ and a set of new categories $\mathcal{Y}_{\mathrm{new}}$.
The two category sets are disjoint, and together form the target space $\mathcal{Y}=\mathcal{Y}_{\mathrm{old}}\cup\mathcal{Y}_{\mathrm{new}}$.
Given $\mathcal{D}_l$ and $\mathcal{D}_u$, GCD aims to learn a model $f:\mathcal{X}\rightarrow\widehat{\mathcal{Y}}$, where $\widehat{\mathcal{Y}}$ denotes the predicted output indices. Unless otherwise specified, we assume that the total number of target categories is given and set $|\widehat{\mathcal{Y}}|=|\mathcal{Y}|$, following the common practice \cite{DBLP:conf/cvpr/VazeHVZ22,DBLP:conf/iccv/WenZQ23}. When this number is unknown, $|\widehat{\mathcal{Y}}|$ can be estimated via off-the-shelf methods \cite{DBLP:conf/cvpr/VazeHVZ22,GPC}.
During evaluation, $f$ is applied to each sample in $\mathcal{D}_u$, recognizing old-category samples and assigning new-category samples to proper category indices.

Early GCD methods follow a \emph{unified clustering-based framework} \cite{DBLP:conf/cvpr/VazeHVZ22,DCCL,PromptCAL}. Specifically, the model can be formulated as $f=\psi\circ\phi$, where $\phi: \mathcal{X}\rightarrow\mathbb{R}^{d}$ is a feature extractor and $\psi:\mathbb{R}^{d}\rightarrow\widehat{\mathcal{Y}}$ is a non-parametric clustering procedure. In this framework, $\widehat{\mathcal{Y}}$ is fully exchangeable and will be aligned with the ground-truth categories during evaluation.

Recent methods instead adopt a \emph{divide-and-conquer parametric framework} \cite{DBLP:conf/iccv/WenZQ23,DBLP:conf/nips/VazeVZ23,DBLP:conf/cvpr/CaoZWYS0L024,wang2025get}. These methods replace $\psi$ with a trainable classifier or prototype head $h: \mathbb{R}^{d}\rightarrow\widehat{\mathcal{Y}}$, so that the model can be formulated as $f=h\circ\phi$. In this case, the output space is partitioned as $\widehat{\mathcal{Y}}=\widehat{\mathcal{Y}}_{\mathrm{old}}\cup\widehat{\mathcal{Y}}_{\mathrm{new}}$ and is only partially exchangeable: the old-category dimensions in $\widehat{\mathcal{Y}}_{\mathrm{old}}$ are semantically anchored by the supervision from $\mathcal{D}_l$, while the new-category dimensions in $\widehat{\mathcal{Y}}_{\mathrm{new}}$ remain exchangeable and are learned through discovery-oriented objectives defined on $\mathcal{D}_u$.

\subsection{Standard GCD Evaluation Protocol}
\label{sec:metrics}
The standard GCD protocol treats every predicted index as exchangeable and finds one assignment over $\mathcal{D}_u$ with the Hungarian algorithm \cite{kuhn1955hungarian,DBLP:conf/cvpr/VazeHVZ22}. Let $\Pi$ be the set of one-to-one assignments from $\widehat{\mathcal{Y}}$ to $\mathcal{Y}$. The evaluator has access to the ground-truth category $y(\boldsymbol{x})$ and obtains
\begin{equation}
    \mathcal{A}_{\mathrm{GH}}
    \in
    \arg\max_{\mathcal{A}\in\Pi}
    \sum_{\boldsymbol{x}\in\mathcal{D}_u}
    \mathbb{I}\{\mathcal{A}(f(\boldsymbol{x}))=y(\boldsymbol{x})\}.
\end{equation}
This single assignment is then reused for every reported subset. For $r\in\{\mathrm{old},\mathrm{new},\mathrm{all}\}$, define $\mathcal{D}_{u,r}:=\{\boldsymbol{x}\in\mathcal{D}_u:y(\boldsymbol{x})\in\mathcal{Y}_r\}$, where $\mathcal{Y}_{\mathrm{all}}=\mathcal{Y}$. The corresponding accuracy is
\begin{equation}
    \texttt{Acc}_{\mathrm{GH}}^{r}(f)
    =
    \frac{1}{|\mathcal{D}_{u,r}|}
    \sum_{\boldsymbol{x}\in\mathcal{D}_{u,r}}
    \mathbb{I}\{\mathcal{A}_{\mathrm{GH}}(f(\boldsymbol{x}))=y(\boldsymbol{x})\}.
\end{equation}
For conciseness, we write these three scores as $\texttt{Old}_{\mathrm{GH}}$, $\texttt{New}_{\mathrm{GH}}$, and $\texttt{All}_{\mathrm{GH}}$, respectively. 

This protocol is appropriate when all output indices are exchangeable. However, a parametric GCD head is only partially exchangeable. This mismatch can induce a problem we call \emph{semantic mismapping}, which we will analyze in the next section.

\section{Semantic Mismapping and Semantics-Preserving Evaluation}
\label{sec:semantic_mismapping}
\subsection{Semantic Mismapping under Global Matching}
\begin{figure}[t]
\centering
\vspace{-0.35em}
\includegraphics[width=\linewidth]{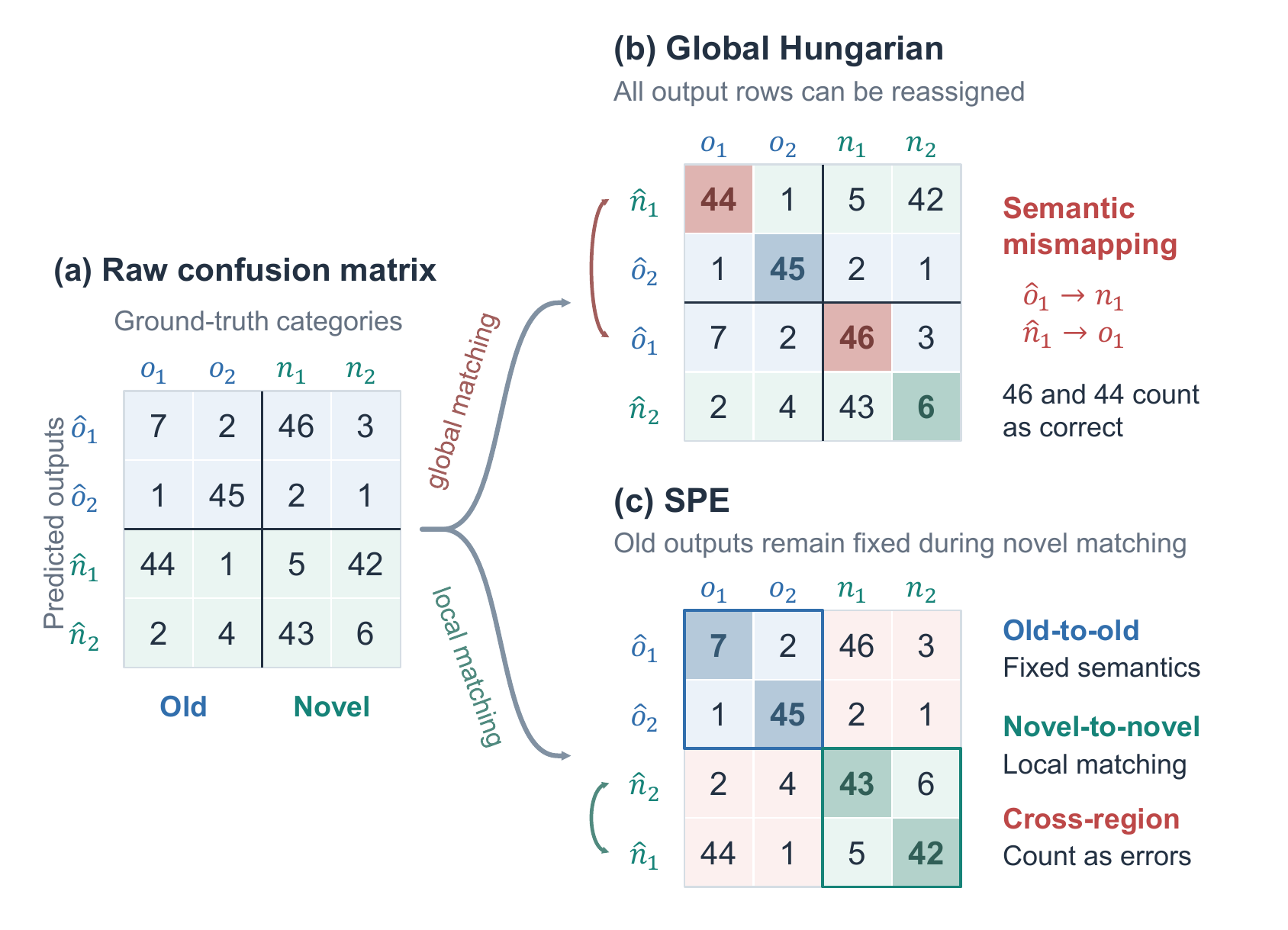}
\vspace{-0.55em}
\caption{Comparison of GH and SPE using the same predictions. \textbf{(a)} The raw confusion matrix has predicted outputs as rows and ground-truth categories as columns, with old and new categories denoted by $o$ and $n$. \textbf{(b)} GH reassigns the anchored old output $\hat{o}_1$ to $n_1$ and the new output $\hat{n}_1$ to $o_1$, counting 46 false-old predictions and 44 old-to-new errors as correct. \textbf{(c)} SPE {preserves the semantics of old outputs} and applies local matching within the new region. In this way, predictions in either cross-region block remain errors, including the 46 and 44 samples hidden by GH.}
\label{fig:intro_bias_metric}
\vspace{-0.4em}
\end{figure}

To illustrate semantic mismapping, we first partition the confusion matrix into four blocks, with predicted output groups as rows and ground-truth category groups as columns:
\begin{equation*}
    \renewcommand{\arraystretch}{1.3}
    \begin{array}{c|cc}
        & \mathcal{Y}_{\mathrm{old}}
        & \mathcal{Y}_{\mathrm{new}} \\ \hline
        \widehat{\mathcal{Y}}_{\mathrm{old}}
        & \textcolor{DiagPastel}{\mathcal{Y}_{\mathrm{old}}\!\times\!\widehat{\mathcal{Y}}_{\mathrm{old}}}
        & \textcolor{OffDiagPastel}{\mathcal{Y}_{\mathrm{new}}\!\times\!\widehat{\mathcal{Y}}_{\mathrm{old}}} \\
        \widehat{\mathcal{Y}}_{\mathrm{new}}
        & \textcolor{OffDiagPastel}{\mathcal{Y}_{\mathrm{old}}\!\times\!\widehat{\mathcal{Y}}_{\mathrm{new}}}
        & \textcolor{DiagPastel}{\mathcal{Y}_{\mathrm{new}}\!\times\!\widehat{\mathcal{Y}}_{\mathrm{new}}}
    \end{array}
\end{equation*}
The diagonal blocks, \textit{i.e.}, $\textcolor{DiagPastel}{\mathcal{Y}_{\mathrm{old}}\!\times\!\widehat{\mathcal{Y}}_{\mathrm{old}}}$ and $\textcolor{DiagPastel}{\mathcal{Y}_{\mathrm{new}}\!\times\!\widehat{\mathcal{Y}}_{\mathrm{new}}}$, represent old-to-old and new-to-new predictions, respectively. In contrast, the off-diagonal blocks, \textit{i.e.}, $\textcolor{OffDiagPastel}{\mathcal{Y}_{\mathrm{old}}\!\times\!\widehat{\mathcal{Y}}_{\mathrm{new}}}$ and $\textcolor{OffDiagPastel}{\mathcal{Y}_{\mathrm{new}}\!\times\!\widehat{\mathcal{Y}}_{\mathrm{old}}}$, represent old-to-new and new-to-old errors, respectively.

How does global matching induce semantic mismapping? Next, we take the predictions in Fig.\ref{fig:intro_bias_metric} as an example. In this case, keeping $\hat{o}_1\mapsto o_1$ and $\hat{n}_1\mapsto n_1$ only yields $7+5=12$ correct predictions, whereas swapping the two mappings yields $46+44=90$ seemingly correct ones. Thus, global matching will choose $\mathcal{A}(\hat{o}_1)=n_1$ and $\mathcal{A}(\hat{n}_1)=o_1$, counting both groups of cross-boundary errors as correct, as shown in Fig.\ref{fig:intro_bias_metric}(b). 
Inspired by this example, we have the following definition of semantic mismapping.

\begin{definition}[Semantic mismapping]
\label{def:semantic_mismapping}
For each old category $y_o\in\mathcal{Y}_{\mathrm{old}}$, let $\hat{y}_o\in\widehat{\mathcal{Y}}_{\mathrm{old}}$ denote its output index in the parametric head $h$. Given an assignment $\mathcal{A}\in\Pi$, semantic mismapping occurs at $\hat{y}_o$ if $\mathcal{A}(\hat{y}_o)\neq y_o$.
\end{definition}

Next, we ask when semantic mismapping can increase the global matching score. To answer this question, we generalize the two-category swap in Fig.\ref{fig:intro_bias_metric}(b) to arbitrary permutations of the output indices. Suppose the outputs are initially assigned as follows: $\hat{y}_1\mapsto y_1$, $\hat{y}_2\mapsto y_2$, and $\hat{y}_3\mapsto y_3$. Global matching may replace these mappings with the cycle $\hat{y}_1\mapsto y_2$, $\hat{y}_2\mapsto y_3$, and $\hat{y}_3\mapsto y_1$, while leaving all other outputs unchanged. Since any reassignment can be decomposed into such cycles, its change in matching score can be measured by the correct predictions lost and gained along each cycle. The following theorem, proved in App.\ref{app:proof_mismapping_margins}, formalizes this comparison.

\begin{theorem}[Semantic mismapping margin]
\label{thm:semantic_mismapping_margin}
Let $C_{y\rightarrow\hat{y}}$ denote the number of samples from true category $y$ predicted as output index $\hat{y}\in\widehat{\mathcal{Y}}$. We define the highest-scoring assignment that preserves the supervised old-output meanings by
\begin{equation*}
\begin{aligned}
\mathcal{A}_0
&\in
\arg\max_{\mathcal{A}\in\Pi}
\sum_{\hat{y}\in\widehat{\mathcal{Y}}}
C_{\mathcal{A}(\hat{y})\rightarrow\hat{y}} \\
\textnormal{s.t.}\quad
&\mathcal{A}(\hat{y}_o)=y_o,
\quad \forall y_o\in\mathcal{Y}_{\mathrm{old}},\\
&\mathcal{A}(\widehat{\mathcal{Y}}_{\mathrm{new}})
=\mathcal{Y}_{\mathrm{new}}.
\end{aligned}
\end{equation*}
For distinct categories $y_1,\ldots,y_k$ with $k\geq2$, set $\hat{y}_i=\mathcal{A}_0^{-1}(y_i)$, $y_{k+1}=y_1$, and $\mathcal{Y}_{\gamma}:=\{y_1,\ldots,y_k\}$. The permutation cycle $\gamma=(y_1,\ldots,y_k)$ defines an assignment $\mathcal{A}_{\gamma}$ that maps $\hat{y}_i$ to $y_{i+1}$ and agrees with $\mathcal{A}_0$ elsewhere. Define the cycle margin as
\begin{equation}
    m_{\gamma}
    :=
    \sum_{i=1}^{k}
    \left(
    C_{y_i\rightarrow\hat{y}_i}
    -C_{y_{i+1}\rightarrow\hat{y}_i}
    \right).
\end{equation}
The assignment $\mathcal{A}_{\gamma}$ improves the global matching score if and only if $m_{\gamma}<0$, and ties when $m_{\gamma}=0$. In other words, define the semantic-mismapping margin by
\begin{equation}
    m_{\mathrm{sem}}
    :=
    \min_{\gamma:\,\mathcal{Y}_{\gamma}\cap\mathcal{Y}_{\mathrm{old}}\neq\varnothing}
    m_{\gamma}.
\end{equation}
Then, $\mathcal{A}_0$ is globally optimal over the entire category space if and only if $m_{\mathrm{sem}}\geq0$.
\end{theorem}

\begingroup
\color{black}
\begin{algorithm}[!t]
\centering
\begin{minipage}{0.96\linewidth}
\noindent\textbf{Input:} Samples $\boldsymbol{x}\in\mathcal{D}_{u}$, labels $y$, predictions $\hat{y}=f(\boldsymbol{x})$, old categories $\mathcal{Y}_{\mathrm{old}}$, new categories $\mathcal{Y}_{\mathrm{new}}$
\par\noindent\textbf{Output:} $\texttt{Old}_{\mathrm{SPE}}$, $\texttt{New}_{\mathrm{SPE}}$, $\texttt{All}_{\mathrm{SPE}}$, and $\texttt{H}_{\mathrm{SPE}}$
\vspace{3pt}
\hrule height 0.4pt
\vspace{3pt}
\begin{algorithmic}[1]
\STATE Split $\mathcal{D}_{u}$ into $\mathcal{D}_{u,\mathrm{old}}$ and $\mathcal{D}_{u,\mathrm{new}}$.
\STATE Compute $\texttt{Old}_{\mathrm{SPE}}$ using the old outputs.
\STATE Build the confusion matrix between $\mathcal{Y}_{\mathrm{new}}$ and $\widehat{\mathcal{Y}}_{\mathrm{new}}$ using samples $\{\boldsymbol{x}: y \in \mathcal{Y}_{\mathrm{new}}, \hat{y} \in \widehat{\mathcal{Y}}_{\mathrm{new}}\}$.
\STATE Obtain the local Hungarian matching $\mathcal{A}_{\mathrm{new}}^{*}$ using Eq.\eqref{eq:spe_local_matching}.
\STATE Compute $\texttt{New}_{\mathrm{SPE}}$ and $\texttt{All}_{\mathrm{SPE}}$ using Eq.\eqref{eq:spe_accuracy}, and $\texttt{H}_{\mathrm{SPE}}$ using Eq.\eqref{eq:spe_harmonic}.
\end{algorithmic}
\end{minipage}
\caption{\textcolor{black}{Semantics-Preserving Evaluation for parametric GCD.}}
\label{alg:spe_protocol}
\end{algorithm}
\endgroup

\begin{remark}[Mismatch types]
\label{rem:mismatch_type}
We distinguish two types: Type \textnormal{(I)}, \emph{within-old semantic mismapping}, occurs when $\mathcal{A}(\hat{y}_o)\in\mathcal{Y}_{\mathrm{old}}\setminus\{y_o\}$. Type \textnormal{(II)}, \emph{cross-boundary semantic mismapping}, occurs when $\mathcal{A}(\hat{y}_o)\in\mathcal{Y}_{\mathrm{new}}$. Accordingly, an old-only cycle with $\mathcal{Y}_{\gamma}\subseteq\mathcal{Y}_{\mathrm{old}}$ produces only type \textnormal{(I)}, whereas a mixed cycle satisfying $\mathcal{Y}_{\gamma}\cap\mathcal{Y}_{\mathrm{old}}\neq\varnothing$ and $\mathcal{Y}_{\gamma}\cap\mathcal{Y}_{\mathrm{new}}\neq\varnothing$ may additionally produce type \textnormal{(II)}.
\end{remark}

\begin{table*}[!t]
    \centering
    \caption{Comparison of existing GCD evaluation protocols. The first property indicates whether a predicted output can be repeatedly assigned to different categories. The latter two properties correspond to the prevention of Type~\textnormal{(I)} and Type~\textnormal{(II)} semantic mismapping, respectively.}
    \label{tab:protocol_comparison}
    \begin{tabular}{llccc}
        \toprule
        Protocol & Matching rule & \shortstack{Each output\\assigned only once?} & \shortstack{Within-old semantics\\preserved?} & \shortstack{Cross-region\\errors detected?} \\
        \midrule
        Legacy NCD & Repeated local matching & \xmark & \xmark & \xmark \\
        Global Hungarian (GH) & Single global matching & \cmark & \xmark & \xmark \\
        SPE & Single local matching & \cmark & \cmark & \cmark \\
        \bottomrule
    \end{tabular}
\end{table*}

\subsection{Semantics-Preserving Evaluation}
\label{sec:spe_protocol}
So far, we have shown that global matching can overestimate the performance of a parametric GCD model. To mitigate this issue, we propose Semantics-Preserving Evaluation (SPE), an evaluation protocol aligned with the divide-and-conquer framework of parametric GCD. Specifically, SPE evaluates the different output regions as follows:
\begin{itemize}
    \item \textbf{Old output region.} Since each old output represents a known category, SPE directly checks whether the prediction is correct.
    \item \textbf{New output region.} Since new outputs do not have predefined categories, SPE applies local Hungarian matching between new outputs and new categories.
    \item \textbf{Cross-region predictions.} An old sample predicted in the new output region or a new sample predicted in the old output region is directly counted as an error, as SPE no longer allows the matching algorithm to correct such cross-region predictions.
\end{itemize}

Formally, SPE keeps the original category meanings of the old outputs in the classifier head $h$. For new outputs, let $\Pi_{\mathrm{new}}$ be the one-to-one mappings from $\widehat{\mathcal{Y}}_{\mathrm{new}}$ to $\mathcal{Y}_{\mathrm{new}}$. Then, the local matching can be denoted as
\begin{equation}
    \label{eq:spe_local_matching}
    \mathcal{A}_{\mathrm{new}}^{*}
    \in
    \arg\max_{\mathcal{A}_{\mathrm{new}}\in\Pi_{\mathrm{new}}}
    \sum_{\hat{y}\in\widehat{\mathcal{Y}}_{\mathrm{new}}}
    C_{\mathcal{A}_{\mathrm{new}}(\hat{y})\rightarrow\hat{y}}.
\end{equation}
Overall, the SPE mapping is defined as:
\begin{equation}
    \mathcal{A}_{\mathrm{SPE}}(\hat{y})
    =
    \begin{cases}
    y_o,
        & \hat{y}=\hat{y}_o,\ y_o\in\mathcal{Y}_{\mathrm{old}},\\
    \mathcal{A}_{\mathrm{new}}^{*}(\hat{y}),
        & \hat{y}\in\widehat{\mathcal{Y}}_{\mathrm{new}}.
    \end{cases}
\end{equation}
Using the same subsets $\mathcal{D}_{u,r}$ as in Sec.\ref{sec:metrics}, the SPE accuracies are defined as
\begin{equation}
    \label{eq:spe_accuracy}
    \texttt{Acc}_{\mathrm{SPE}}^{r}(f)
    =
    \frac{1}{|\mathcal{D}_{u,r}|}
    \sum_{\boldsymbol{x}\in\mathcal{D}_{u,r}}
    \mathbb{I}\{\mathcal{A}_{\mathrm{SPE}}(f(\boldsymbol{x}))=y(\boldsymbol{x})\},
\end{equation}
where $r\in\{\mathrm{old},\mathrm{new},\mathrm{all}\}$. For conciseness, we denote the three accuracies by $\texttt{Old}_{\mathrm{SPE}}$, $\texttt{New}_{\mathrm{SPE}}$, and $\texttt{All}_{\mathrm{SPE}}$, respectively.
Overall, the SPE procedure is summarized in Alg.\ref{alg:spe_protocol}. For a single summary of the balance between old-category recognition and novel-category discovery, we additionally report their harmonic mean:
\begin{equation}
    \label{eq:spe_harmonic}
    \texttt{H}_{\mathrm{SPE}}
    =
    \frac{2\texttt{Old}_{\mathrm{SPE}}\cdot\texttt{New}_{\mathrm{SPE}}}
    {\texttt{Old}_{\mathrm{SPE}}+\texttt{New}_{\mathrm{SPE}}}.
\end{equation}

\textbf{SPE vs. Existing GCD Evaluation Protocols.} In Tab.\ref{tab:protocol_comparison}, we compare existing GCD evaluation protocols through three properties. Legacy NCD uses ground-truth categories to pre-partition samples into old and new subsets and performs two separate local matchings. Consequently, the same predicted output can be reused across the two subsets and assigned to different categories, as discussed in \cite{DBLP:conf/cvpr/VazeHVZ22}. Although GH does not pre-partition samples, its global matching neither preserves anchored old semantics nor reliably detects cross-region errors, making it possible to overestimate model performance. In contrast, SPE satisfies all three properties. First, unlike Legacy NCD, SPE only performs a single local matching for the new outputs. Second, SPE directly calculates the classification accuracy of the old outputs, excluding matching procedures and thus preventing Type~\textnormal{(I)} semantic mismapping. Third, SPE directly counts cross-region predictions as errors, preventing their correction through reassignment and thus preventing Type~\textnormal{(II)} semantic mismapping. Besides, the local matching among new outputs {preserves} the permutation invariance required for category discovery. 

\subsection{Extensions of SPE Protocol}
\label{sec:nonparametric_extension}
In this part, we extend SPE beyond parametric GCD methods and the default equal-cardinality setting, \textit{i.e.}, $|\widehat{\mathcal{Y}}|=|\mathcal{Y}|$. 

\textbf{Extension to Non-Parametric GCD.}
SPE requires a partially exchangeable output space and thus cannot be directly applied to non-parametric GCD methods. Fortunately, we can extend SPE to non-parametric methods by identifying the old-category indices based on the labeled set $\mathcal{D}_l$. 
To this end, let $\Pi_l$ denote the set of one-to-one maps from $\mathcal{Y}_{\mathrm{old}}$ to $\widehat{\mathcal{Y}}$. We define the old-output mapping by
\begin{equation}
    \label{eq:nonparametric_calibration}
    b_l^{*}
    \in
    \arg\max_{b\in\Pi_l}
    \sum_{(\boldsymbol{x},y)\in\mathcal{D}_l}
    \mathbb{I}\{f(\boldsymbol{x})=b(y)\}.
\end{equation}
Then, the output space of $f=\psi\circ\phi$ is partitioned as
\begin{equation}
    \widehat{\mathcal{Y}}_{\mathrm{old}}
    :=b_l^{*}(\mathcal{Y}_{\mathrm{old}}),
    \qquad
    \widehat{\mathcal{Y}}_{\mathrm{new}}
    :=\widehat{\mathcal{Y}}\setminus\widehat{\mathcal{Y}}_{\mathrm{old}}.
\end{equation}
On top of this partition, we can apply SPE to evaluate the non-parametric GCD model. Notably, this extension is applicable to most non-parametric GCD methods, which use semi-supervised $k$-means as the non-parametric clustering procedure \cite{DBLP:conf/cvpr/VazeHVZ22,DCCL,PromptCAL}.

\textbf{Extension to Estimated New Category Number.}
When $|\widehat{\mathcal{Y}}|$ is estimated and thus not equal to $|\mathcal{Y}|$, standard GCD evaluation performs global matching over $\widehat{\mathcal{Y}}$ and $\mathcal{Y}$ after padding the smaller set with null elements \cite{DBLP:conf/cvpr/VazeHVZ22}. Following this practice, we extend SPE by applying the padding only to the new region. Formally, if $|\widehat{\mathcal{Y}}_{\mathrm{new}}|<|\mathcal{Y}_{\mathrm{new}}|$, we add $|\mathcal{Y}_{\mathrm{new}}|-|\widehat{\mathcal{Y}}_{\mathrm{new}}|$ dummy outputs; if $|\widehat{\mathcal{Y}}_{\mathrm{new}}|>|\mathcal{Y}_{\mathrm{new}}|$, we add $|\widehat{\mathcal{Y}}_{\mathrm{new}}|-|\mathcal{Y}_{\mathrm{new}}|$ dummy categories. The resulting new sets are
\begin{equation}
    \widehat{\mathcal{Y}}_{\mathrm{new}}^{D}
    :=\widehat{\mathcal{Y}}_{\mathrm{new}}
    \cup\widehat{\mathcal{Y}}^{D},
    \qquad
    \mathcal{Y}_{\mathrm{new}}^{D}
    :=\mathcal{Y}_{\mathrm{new}}
    \cup\mathcal{Y}^{D}.
\end{equation}
Matching a real output to a dummy category leaves the output unmatched, while matching a dummy output to a real category leaves the category unmatched. Hence, every confusion-matrix entry involving a dummy is set to zero, and the problem is converted into an equal-cardinality assignment. Accordingly, Thm.\ref{thm:semantic_mismapping_margin} yields the following corollary.

\begin{corollary}[Cycle margin with estimated cardinalities]
\label{cor:unequal_cycle_margin}
Define the zero-padded confusion matrix by
\begin{equation}
    C^{D}_{y\rightarrow\hat{y}}
    :=
    \begin{cases}
        C_{y\rightarrow\hat{y}},
        & y\notin\mathcal{Y}^{D},\ \hat{y}\notin\widehat{\mathcal{Y}}^{D},\\
        0, & \textnormal{otherwise}.
    \end{cases}
\end{equation}
Let $\mathcal{A}_0^{D}$ be the highest-scoring one-to-one assignment from $\widehat{\mathcal{Y}}_{\mathrm{old}}\cup\widehat{\mathcal{Y}}_{\mathrm{new}}^{D}$ to $\mathcal{Y}_{\mathrm{old}}\cup\mathcal{Y}_{\mathrm{new}}^{D}$ that preserves the old-output meanings. For a cycle $\gamma=(y_1,\ldots,y_k)$ with $k\geq2$, let $\hat{y}_i$ denote the output assigned to $y_i$ by $\mathcal{A}_0^{D}$, and set $y_{k+1}=y_1$ and $\mathcal{Y}_{\gamma}:=\{y_1,\ldots,y_k\}$. The margin is
\begin{equation}
    m_{\gamma}^{D}
    :=
    \sum_{i=1}^{k}
    \left(
    C^{D}_{y_i\rightarrow\hat{y}_i}
    -C^{D}_{y_{i+1}\rightarrow\hat{y}_i}
    \right).
\end{equation}
Define
\begin{equation}
    m_{\mathrm{sem}}^{D}
    :=
    \min_{\gamma:\,\mathcal{Y}_{\gamma}\cap\mathcal{Y}_{\mathrm{old}}\neq\varnothing}
    m_{\gamma}^{D}.
\end{equation}
Global matching achieves a higher score than $\mathcal{A}_0^{D}$ if and only if $m_{\mathrm{sem}}^{D}<0$. Equivalently, $\mathcal{A}_0^{D}$ is globally optimal on the padded sets if and only if $m_{\mathrm{sem}}^{D}\geq0$.
\end{corollary}

\section{MMO}
\label{sec:optimization}
SPE prevents semantic mismapping during evaluation by preserving anchored old-output semantics. During training, we propose MMO to boost existing parametric GCD methods with two plug-and-play losses.

\subsection{Objectives of MMO}
\label{sec:mmo_losses}
The two losses follow the divide-and-conquer principle. On the one hand, Region-Balance Loss (RBL) regulates prediction mass across the old and new regions to improve old/new routing. On the other hand, Novel-Diversity Loss (NDL) strengthens discrimination within the new region by encouraging new outputs to be diverse and well-separated.

\textbf{Region-Balance Loss.}
To alleviate prediction bias, we constrain the aggregate probability assigned to each region on unlabeled data. To this end, we encourage the batch-average probability mass $\bar{q}^{\,r}$ assigned to each region $r\in\{\mathrm{old},\mathrm{new}\}$ to match its estimated sample proportion $p_r$ in the unlabeled training set.

Formally, for sample $i$, let $s_{i,j}$ denote the logit of output $j$. We define the probability mass assigned to region $r$ as 
\begin{equation}
    q_i^r
    =
    \frac{
        \sum_{j \in \widehat{\mathcal{Y}}_r}
        \exp(s_{i,j}/\tau)
    }{
        \sum_{j \in \widehat{\mathcal{Y}}}
        \exp(s_{i,j}/\tau)
    },
    \qquad
    r \in \{\mathrm{old}, \mathrm{new}\},
    \label{eq:pao_region_probability}
\end{equation}
where $\tau$ is the temperature hyperparameter and
$q_i^{\mathrm{old}} + q_i^{\mathrm{new}} = 1$.
For an unlabeled batch $\mathcal{B}_u$, we compute the average regional probabilities as
\begin{equation}
    \bar{q}^{\,r}
    =
    \frac{1}{|\mathcal{B}_u|}
    \sum_{i \in \mathcal{B}_u} q_i^r.
\end{equation}

\noindent Then, let $p_r$ denote the estimated proportion of region $r$ in the unlabeled training set, with $p_{\mathrm{old}}+p_{\mathrm{new}}=1$. We constrain the regional probability allocation through the following cross-entropy loss:
\begin{equation}
    \mathcal{L}_{\mathrm{RB}}
    =\sum_{r\in\{\mathrm{old},\mathrm{new}\}}
    - p_r\log\bar q^{\,r}.
    \label{eq:pao_balance_loss}
\end{equation}
In practice, the old/new sample proportions in each minibatch may differ from the overall proportions $p_r$. To address this issue, we propose a tolerance-based variant that accommodates such differences, as detailed in App.\ref{app:rbl_implementation}. We use this variant in our experiments.

\begin{remark}[Estimating the regional prior]
\label{rem:rbl_prior}
We consider two cases, with $p_{\mathrm{new}}=1-p_{\mathrm{old}}$ in both.
\textbf{(1) Balanced category sizes.} If samples are approximately uniformly distributed across categories before splitting, let $C_r=|\mathcal{Y}_r|$. Then
\begin{equation}
    p_{\mathrm{old}}
    = \frac{C_{\mathrm{old}}(N_l+N_u)/(C_{\mathrm{old}}+C_{\mathrm{new}})-N_l}{N_u}.
    \label{eq:rbl_balanced_prior}
\end{equation}
\textbf{(2) Imbalanced category sizes.} If the fraction of old-category samples selected for labeling, denoted by $\rho$, is known, we use
\begin{equation}
    p_{\mathrm{old}}=\frac{(1-\rho)N_l}{\rho N_u}.
    \label{eq:pao_region_prior}
\end{equation}
Otherwise, we estimate $\rho$ with TIcE~\cite{BekkerDavis2018TIcE}, assuming that old-category samples are selected uniformly at random for labeling. App.\ref{app:rbl_prior_estimation} describes the procedure.
\end{remark}

\begin{table*}[!t]
    \centering
    \footnotesize
    \caption{Basic information about the datasets used in the benchmark.}
    \fitTable{
      \begin{tabular}{ccccc|ccccc}
      \toprule
      \multirow{2}[4]{*}{Dataset} & \multicolumn{2}{c}{Labeled $\mathcal{D}_{l}$} & \multicolumn{2}{c|}{Unlabeled $\mathcal{D}_{u}$} & \multirow{2}[4]{*}{Dataset} & \multicolumn{2}{c}{Labeled $\mathcal{D}_{l}$} & \multicolumn{2}{c}{Unlabeled $\mathcal{D}_{u}$} \\
      \cmidrule{2-5}\cmidrule{7-10} & Images & Classes & Images & Classes & & Images & Classes & Images & Classes \\
      \midrule
      CIFAR-10 & 12.5k & 5 & 37.5k & 10 & CUB & 1.5k & 100 & 4.5k & 200 \\
      CIFAR-100 & 20.0k & 80 & 30.0k & 100 & Stanford Cars & 2.0k & 98 & 6.1k & 196 \\
      ImageNet-100 & 31.9k & 50 & 95.3k & 100 & FGVC-Aircraft & 1.7k & 50 & 5.0k & 100 \\
      ImageNet-1K & 321k & 500 & 960k & 1000 & Herbarium 19 & 8.9k & 341 & 25.4k & 683 \\
      \bottomrule
      \end{tabular}
    }
    \label{tab:dataset}
\end{table*}

\begin{table*}[!t]
    \centering
    \footnotesize
    \caption{Comparison between GH and SPE across benchmark families for seven method--backbone configurations. \colorbox{black!8}{Gray shading} highlights configurations whose ranks differ between the two protocols. The score gaps $\Delta_r, r \in \{\mathrm{new},\mathrm{all}\}$ and rank changes reveal GH-induced performance overestimation, as the answer to \textbf{RQ1}.}
    \label{tab:protocol_benchmark}
    \setlength{\tabcolsep}{2.8pt}
    \renewcommand{\arraystretch}{1.10}
    \definecolor{ProtocolGapPositive}{RGB}{232,151,76}
    \definecolor{ProtocolGapNegative}{RGB}{92,160,211}
    \newcommand{\ProtocolGap}[1]{%
        \ifdim#1pt<0pt
            \def\ProtocolGapFill{ProtocolGapNegative}%
            \dimen0=-#1pt\relax
        \else
            \def\ProtocolGapFill{ProtocolGapPositive}%
            \dimen0=#1pt\relax
        \fi
        \ifdim\dimen0<1pt \def\ProtocolGapShade{10}%
        \else\ifdim\dimen0<3pt \def\ProtocolGapShade{23}%
        \else\ifdim\dimen0<7pt \def\ProtocolGapShade{38}%
        \else\ifdim\dimen0<15pt \def\ProtocolGapShade{55}%
        \else \def\ProtocolGapShade{75}\fi\fi\fi\fi
        \ifdim#1pt>0pt\ifdim\dimen0<15pt\else\def\ProtocolGapShade{60}\fi\fi
        \ifdim\dimen0=0pt
            \def\ProtocolGapFill{ProtocolGapNegative}\def\ProtocolGapShade{5}%
        \fi
        \edef\ProtocolGapPaint{\noexpand\cellcolor{\ProtocolGapFill!\ProtocolGapShade!white}}%
        \ProtocolGapPaint\textcolor{black}{#1}%
    }
    \makeatletter
    \newcommand{\ProtocolRankEnd}[1]{%
        \tikz[remember picture,overlay]{\coordinate (protocol-rank-#1) at (0,0);}%
    }
    \newcommand{\ProtocolRankBridge}[1]{%
        \tikz[remember picture,overlay]{%
            \fill[black!8] ([yshift=-\dp\@arstrutbox]protocol-rank-#1)
                rectangle (0,\ht\@arstrutbox);%
        }%
    }
    \makeatother
    \begin{tabular*}{\textwidth}{@{\extracolsep{\fill}}lllrrrrrr@{}}
        \toprule
        Dataset & Method & Backbone
        & $\texttt{New}_{\mathrm{GH}}$ & $\texttt{New}_{\mathrm{SPE}}$ & $\Delta_{\mathrm{new}}$
        & $\texttt{Old}_{\mathrm{SPE}}$ & $\texttt{All}_{\mathrm{SPE}}$ & $\Delta_{\mathrm{all}}$ \\
        \midrule
        \multicolumn{9}{l}{\textit{(a) Coarse-grained benchmarks}} \\
        \multirow{7}{*}{CIFAR-10} & LegoGCD & DINO & 97.93 & 97.93 & \ProtocolGap{0.00} & 95.76 & 97.21 & \ProtocolGap{0.00} \\
         & SimGCD-BIA & DINO & 98.74 & 98.74 & \ProtocolGap{0.00} & 84.25 & 93.91 & \ProtocolGap{0.00} \\
         & ProtoGCD & DINO & 98.94 & 98.94 & \ProtocolGap{0.00} & 81.94 & 93.27 & \ProtocolGap{0.00} \\
         & SimGCD & DINO & 98.48 & 98.48 & \ProtocolGap{0.00} & 93.71 & 96.89 & \ProtocolGap{0.00} \\
         & SimGCD & CLIP & 98.11 & 98.11 & \ProtocolGap{0.00} & 94.75 & 96.99 & \ProtocolGap{0.00} \\
         & GET-TAR & CLIP & 98.42 & 98.42 & \ProtocolGap{0.00} & 94.70 & 97.18 & \ProtocolGap{0.00} \\
         & GET & CLIP & 98.36 & 98.36 & \ProtocolGap{0.00} & 94.87 & 97.19 & \ProtocolGap{0.00} \\
        \cmidrule(lr){2-9}
        \multirow{7}{*}{CIFAR-100} & LegoGCD & DINO & 85.99 & 85.99 & \ProtocolGap{0.00} & 81.71 & 83.13 & \ProtocolGap{0.00} \\
         & SimGCD-BIA & DINO & 78.60 & 78.60 & \ProtocolGap{0.00} & 75.50 & 76.53 & \ProtocolGap{0.00} \\
         & ProtoGCD & DINO & 82.49 & 82.49 & \ProtocolGap{0.00} & 77.82 & 79.38 & \ProtocolGap{0.00} \\
         & SimGCD & DINO & 74.02 & 74.02 & \ProtocolGap{0.00} & 81.69 & 79.13 & \ProtocolGap{0.00} \\
         & SimGCD & CLIP & 73.75 & 71.68 & \ProtocolGap{2.07} & 84.90 & 80.50 & \ProtocolGap{0.12} \\
         & GET-TAR & CLIP & 71.69 & 71.69 & \ProtocolGap{0.00} & 85.68 & 81.01 & \ProtocolGap{0.00} \\
         & GET & CLIP & 71.12 & 71.12 & \ProtocolGap{0.00} & 85.69 & 80.83 & \ProtocolGap{0.00} \\
        \cmidrule(lr){2-9}
        \multirow{7}{*}{ImageNet-100} & LegoGCD & DINO & 77.76 & 77.76 & \ProtocolGap{0.00} & 94.88 & 83.49 & \ProtocolGap{0.00} \\
         & SimGCD-BIA & DINO & 78.70 & 78.70 & \ProtocolGap{0.00} & 86.05 & 81.16 & \ProtocolGap{0.00} \\
         & ProtoGCD & DINO & 78.36 & 78.36 & \ProtocolGap{0.00} & 91.00 & 82.59 & \ProtocolGap{0.00} \\
         & SimGCD & DINO & 81.26 & 81.26 & \ProtocolGap{0.00} & 92.58 & 85.04 & \ProtocolGap{0.00} \\
         & SimGCD & CLIP & 87.80 & 87.80 & \ProtocolGap{0.00} & 94.76 & 90.13 & \ProtocolGap{0.00} \\
         & GET-TAR & CLIP & 88.28 & 88.28 & \ProtocolGap{0.00} & 95.43 & 90.67 & \ProtocolGap{0.00} \\
         & GET & CLIP & 92.02 & 92.02 & \ProtocolGap{0.00} & 95.69 & 93.25 & \ProtocolGap{0.00} \\
        \midrule
        \multicolumn{3}{l}{\textit{(b) Fine-grained benchmarks}} & (rank) & (rank) & & & & \\
        \multirow{7}{*}{Herbarium 19} & LegoGCD & DINO & \cellcolor{black!8}37.48 (4)\ProtocolRankEnd{1} & \cellcolor{black!8}\ProtocolRankBridge{1}17.98 (2) & \ProtocolGap{19.50} & 69.02 & 35.83 & \ProtocolGap{8.99} \\
         & SimGCD-BIA & DINO & \cellcolor{black!8}35.65 (7)\ProtocolRankEnd{2} & \cellcolor{black!8}\ProtocolRankBridge{2}16.34 (6) & \ProtocolGap{19.31} & 65.47 & 33.53 & \ProtocolGap{9.00} \\
         & ProtoGCD & DINO & \cellcolor{black!8}35.92 (6)\ProtocolRankEnd{3} & \cellcolor{black!8}\ProtocolRankBridge{3}18.52 (1) & \ProtocolGap{17.40} & 68.04 & 35.84 & \ProtocolGap{7.90} \\
         & SimGCD & DINO & \cellcolor{black!8}36.72 (5)\ProtocolRankEnd{4} & \cellcolor{black!8}\ProtocolRankBridge{4}17.66 (3) & \ProtocolGap{19.06} & 67.09 & 34.95 & \ProtocolGap{9.03} \\
         & SimGCD & CLIP & \cellcolor{black!8}38.26 (2)\ProtocolRankEnd{5} & \cellcolor{black!8}\ProtocolRankBridge{5}12.63 (7) & \ProtocolGap{25.63} & 77.34 & 35.27 & \ProtocolGap{12.13} \\
         & GET-TAR & CLIP & \cellcolor{black!8}37.77 (3)\ProtocolRankEnd{6} & \cellcolor{black!8}\ProtocolRankBridge{6}16.46 (5) & \ProtocolGap{21.31} & 78.59 & 38.20 & \ProtocolGap{9.13} \\
         & GET & CLIP & \cellcolor{black!8}41.11 (1)\ProtocolRankEnd{7} & \cellcolor{black!8}\ProtocolRankBridge{7}17.10 (4) & \ProtocolGap{24.01} & 78.52 & 38.59 & \ProtocolGap{10.44} \\
        \cmidrule(lr){2-9}
        \multirow{7}{*}{Stanford Cars} & LegoGCD & DINO & 43.83 (7) & 40.72 (7) & \ProtocolGap{3.11} & 79.41 & 53.32 & \ProtocolGap{0.70} \\
         & SimGCD-BIA & DINO & \cellcolor{black!8}45.16 (5)\ProtocolRankEnd{8} & \cellcolor{black!8}\ProtocolRankBridge{8}42.19 (6) & \ProtocolGap{2.97} & 74.91 & 52.85 & \ProtocolGap{0.80} \\
         & ProtoGCD & DINO & 45.21 (4) & 43.59 (4) & \ProtocolGap{1.62} & 69.07 & 51.89 & \ProtocolGap{0.40} \\
         & SimGCD & DINO & \cellcolor{black!8}44.61 (6)\ProtocolRankEnd{9} & \cellcolor{black!8}\ProtocolRankBridge{9}42.70 (5) & \ProtocolGap{1.91} & 74.11 & 52.93 & \ProtocolGap{0.60} \\
         & SimGCD & CLIP & 64.06 (3) & 59.38 (3) & \ProtocolGap{4.68} & 89.01 & 69.03 & \ProtocolGap{1.43} \\
         & GET-TAR & CLIP & 72.80 (2) & 60.51 (2) & \ProtocolGap{12.29} & 95.75 & 71.99 & \ProtocolGap{3.97} \\
         & GET & CLIP & 74.58 (1) & 70.75 (1) & \ProtocolGap{3.83} & 92.30 & 77.77 & \ProtocolGap{1.19} \\
        \cmidrule(lr){2-9}
        \multirow{7}{*}{CUB} & LegoGCD & DINO & 57.69 (7) & 55.72 (7) & \ProtocolGap{1.97} & 73.05 & 61.50 & \ProtocolGap{0.69} \\
         & SimGCD-BIA & DINO & 62.00 (4) & 60.89 (4) & \ProtocolGap{1.11} & 60.91 & 60.90 & \ProtocolGap{0.71} \\
         & ProtoGCD & DINO & \cellcolor{black!8}59.13 (5)\ProtocolRankEnd{10} & \cellcolor{black!8}\ProtocolRankBridge{10}58.09 (6) & \ProtocolGap{1.04} & 64.78 & 60.32 & \ProtocolGap{0.65} \\
         & SimGCD & DINO & \cellcolor{black!8}58.09 (6)\ProtocolRankEnd{11} & \cellcolor{black!8}\ProtocolRankBridge{11}58.39 (5) & \ProtocolGap{-0.30} & 65.04 & 60.61 & \ProtocolGap{0.33} \\
         & SimGCD & CLIP & 65.10 (3) & 64.40 (3) & \ProtocolGap{0.70} & 76.12 & 68.31 & \ProtocolGap{0.22} \\
         & GET-TAR & CLIP & 73.01 (2) & 72.07 (2) & \ProtocolGap{0.94} & 78.79 & 74.31 & \ProtocolGap{0.29} \\
         & GET & CLIP & 74.97 (1) & 74.74 (1) & \ProtocolGap{0.23} & 76.85 & 75.44 & \ProtocolGap{0.03} \\
        \cmidrule(lr){2-9}
        \multirow{7}{*}{FGVC-Aircraft} & LegoGCD & DINO & \cellcolor{black!8}50.52 (6)\ProtocolRankEnd{12} & \cellcolor{black!8}\ProtocolRankBridge{12}47.53 (7) & \ProtocolGap{2.99} & 64.71 & 53.25 & \ProtocolGap{1.14} \\
         & SimGCD-BIA & DINO & \cellcolor{black!8}50.40 (7)\ProtocolRankEnd{13} & \cellcolor{black!8}\ProtocolRankBridge{13}49.84 (5) & \ProtocolGap{0.56} & 56.30 & 51.99 & \ProtocolGap{0.10} \\
         & ProtoGCD & DINO & 54.60 (3) & 54.42 (3) & \ProtocolGap{0.18} & 46.04 & 51.63 & \ProtocolGap{0.24} \\
         & SimGCD & DINO & \cellcolor{black!8}50.67 (5)\ProtocolRankEnd{14} & \cellcolor{black!8}\ProtocolRankBridge{14}49.66 (6) & \ProtocolGap{1.01} & 60.14 & 53.15 & \ProtocolGap{0.16} \\
         & SimGCD & CLIP & 53.61 (4) & 53.43 (4) & \ProtocolGap{0.18} & 59.54 & 55.47 & \ProtocolGap{0.12} \\
         & GET-TAR & CLIP & 58.26 (1) & 57.09 (1) & \ProtocolGap{1.17} & 66.39 & 60.19 & \ProtocolGap{0.18} \\
         & GET & CLIP & 58.17 (2) & 55.71 (2) & \ProtocolGap{2.46} & 67.65 & 59.69 & \ProtocolGap{0.62} \\
        \bottomrule
    \end{tabular*}
\end{table*}

\textbf{Novel-Diversity Loss.} 
SPE evaluates novel-category discovery through local matching within the new output region. To improve discrimination among these categories, we encourage orthogonality among new prototypes.

Formally, let $\mathbf{w}_u\in\mathbb{R}^d$ denote the prototype of new output $u$, which is typically instantiated by a learnable classifier weight vector \cite{DBLP:conf/iccv/WenZQ23,ma2025protogcd}. We normalize each nonzero vector as
\begin{equation}
    \widetilde{\mathbf{w}}_u
    =\frac{\mathbf{w}_u}{\|\mathbf{w}_u\|_2},
    \qquad u=1,\ldots,C_{\mathrm{new}}.
\end{equation}
To encourage mutual orthogonality among new-category prototypes, we define the novel-diversity loss as
\begin{equation}
    \mathcal{L}_{\mathrm{ND}}
    =\frac{1}{C_{\mathrm{new}}(C_{\mathrm{new}}-1)}
    \sum_{u\ne v}
    \left(\widetilde{\mathbf{w}}_u^{\top}\widetilde{\mathbf{w}}_v\right)^2.
    \label{eq:pao_novel_diversity}
\end{equation}
Let $\widetilde{\mathbf{W}}_{\mathrm{new}}=[\widetilde{\mathbf{w}}_1,\ldots,\widetilde{\mathbf{w}}_{C_{\mathrm{new}}}]^{\top}\in\mathbb{R}^{C_{\mathrm{new}}\times d}$. Then, the loss can equivalently be written as

\begin{equation}
    \mathcal{L}_{\mathrm{ND}}
    =\frac{\left\|
        \widetilde{\mathbf{W}}_{\mathrm{new}}
        \widetilde{\mathbf{W}}_{\mathrm{new}}^{\top}
        -\mathbf{I}_{C_{\mathrm{new}}}
    \right\|_F^2}{C_{\mathrm{new}}(C_{\mathrm{new}}-1)}.
\end{equation}
Notably, normalization fixes the diagonal entries at one, so the loss penalizes only off-diagonal inner products. Exact orthogonality is attainable when $C_{\mathrm{new}}\leq d$; otherwise, NDL still reduces correlations between new-category prototypes without requiring zero loss. 

\textbf{Overall Objective.}
Overall, the training objective can be summarized as
\begin{equation}
    \mathcal{L}_{\mathrm{train}}
    =
    \mathcal{L}_{\mathrm{base}}
    +
    \lambda_{\mathrm{RB}} \mathcal{L}_{\mathrm{RB}}
    +
    \lambda_{\mathrm{ND}} \mathcal{L}_{\mathrm{ND}},
    \label{eq:pao_training_objective}
\end{equation}
where $\mathcal{L}_{\mathrm{base}}$ is the original parametric GCD objective, and
$\lambda_{\mathrm{RB}}, \lambda_{\mathrm{ND}} \ge 0$
control the two regularizers.
Together, these objectives address cross-region probability allocation and within-new prototype diversity, while preserving the baseline method's supervised and discovery-oriented learning objectives.

\subsection{Theoretical Analysis}
\label{sec:mmo_theory}
To understand how MMO mitigates semantic mismapping, we examine the spurious performance gains obtained by GH. To this end, we bound $\Delta_{\mathrm{all}}:=\texttt{All}_{\mathrm{GH}}-\texttt{All}_{\mathrm{SPE}}$ through $\mathcal{L}_{\mathrm{RB}}$ and $\mathcal{L}_{\mathrm{ND}}$.

Intuitively, with $p$ fixed, RBL minimizes $\mathbb{KL}_{\mathrm{RB}}(p\Vert\bar q)$ and thus encourages $\bar q^{\,\mathrm{old}}$ to match $p_{\mathrm{old}}$, where $p:=(p_{\mathrm{old}},p_{\mathrm{new}})$ and $\bar q:=(\bar q^{\,\mathrm{old}},\bar q^{\,\mathrm{new}})$. In this way, RBL penalizes deviations from the estimated old/new proportions, helping mitigate cross-boundary mismapping. On the other hand, NDL encourages orthogonal new prototypes to reduce confusion between new categories. Reassigning outputs can recover incorrect predictions but also sacrifices correct predictions of the affected categories. The following bound accounts for both effects.

The following theorem formalizes these contributions, with the residual $\epsilon_{\mathrm{res}}$ accounting for additional errors, such as old-category misclassification and inaccurate prior estimates. With $\epsilon_{\mathrm{res}}$ controlled, reducing the two losses can tighten the upper bound on the matching gain and raise the lower bound on semantic cycle margins. The full definition and proof are provided in App.\ref{app:mmo_bound}.

\begin{theorem}[Upper bound on semantic mismapping]
\label{thm:mmo_mismapping_bound}
Under the conditions in App.\ref{app:mmo_bound}, define
\begin{equation}
    B:=\sqrt{\frac{\mathbb{KL}_{\mathrm{RB}}(p\Vert\bar q)}{2}}
    +\alpha\mathcal{L}_{\mathrm{ND}}+\epsilon_{\mathrm{res}},
    \label{eq:mmo_error_bound}
\end{equation}
where $\alpha:=\pi_{\mathrm{new}}^{*}C_{\mathrm{new}}(C_{\mathrm{new}}-1)$, $\pi_{\mathrm{new}}^{*}:=\max_{y\in\mathcal{Y}_{\mathrm{new}}}\pi_y$, and $\pi_y$ denotes the sample proportion of category $y$.
Then, the matching gain satisfies
\begin{equation}
    \Delta_{\mathrm{all}}\leq\min\bigl\{B,[2B-\pi_*]_+\bigr\},
    \label{eq:mmo_gap_bound}
\end{equation}
where $[x]_+:=\max\{x,0\}$ and
\[
    \pi_*:=\min_{c\in\mathcal{Y}_{\mathrm{old}},\,y\in\mathcal{Y}\setminus\{c\}}(\pi_c+\pi_y).
\]
Furthermore, for every permutation cycle $\gamma$, we have
\begin{equation}
    \frac{m_\gamma}{N_u}
    \geq\sum_{y\in\mathcal{Y}_\gamma}\pi_y-2B.
    \label{eq:mmo_cycle_bound}
\end{equation}
\end{theorem}

\begin{remark}
Eq.\eqref{eq:mmo_gap_bound} provides a global guarantee: GH and SPE yield the same overall accuracy when $2B\leq\pi_*$. Eq.\eqref{eq:mmo_cycle_bound} provides a local guarantee for a given cycle $\gamma$ involving old categories: the reassignment along this cycle cannot improve accuracy when $2B\leq\sum_{y\in\mathcal{Y}_\gamma}\pi_y$. Since minimizing the two losses can reduce $B$, both results show how MMO helps suppress the potential gains from semantic mismapping.
\end{remark}

\section{Experiments}
\label{sec:benchmark}

    \definecolor{SPEGainFill}{RGB}{232,151,76}
    \definecolor{SPELossFill}{RGB}{92,160,211}
    \definecolor{SPEGainText}{RGB}{119,57,14}
    \definecolor{SPELossText}{RGB}{24,69,111}
    \newcommand{\SPEgain}[2]{%
        \ifdim#2pt<0pt
            \def\SPEfill{SPELossFill}\def\SPEink{SPELossText}%
            \dimen0=-#2pt\relax
        \else
            \def\SPEfill{SPEGainFill}\def\SPEink{SPEGainText}%
            \dimen0=#2pt\relax
        \fi
        \ifdim\dimen0<1pt \def\SPEshade{10}%
        \else\ifdim\dimen0<3pt \def\SPEshade{23}%
        \else\ifdim\dimen0<7pt \def\SPEshade{38}%
        \else\ifdim\dimen0<15pt \def\SPEshade{55}%
        \else \def\SPEshade{75}\fi\fi\fi\fi
        \ifdim#2pt>0pt\ifdim\dimen0<15pt\else\def\SPEshade{60}\fi\fi
        \ifdim\dimen0=0pt
            \def\SPEfill{SPELossFill}\def\SPEshade{5}\def\SPEink{SPELossText}%
        \fi
        \edef\SPEpaint{\noexpand\cellcolor{\SPEfill!\SPEshade!white}}%
        \SPEpaint
        \mbox{\textcolor{black}{#1}\,{\normalfont\fontsize{6}{7}\selectfont\color{\SPEink}(#2)}}%
    }

\begin{table*}[!t]
    \centering
    \footnotesize
    \setlength{\abovecaptionskip}{3pt}
    \caption{Empirical validation of the cycle-margin analysis in Thm.\ref{thm:semantic_mismapping_margin} using SimGCD. The summed cycle margins $\sum_{\gamma\in\Gamma(\mathcal A_{\mathrm{GH}},\mathcal A_0)}m_\gamma$ exactly recover the observed accuracy gaps $\Delta_{\mathrm{all}}$ through Eq.\eqref{eq:cycle_gap_identity}. The last two columns report the numbers of old outputs involved in Type~\textnormal{(I)} within-old and Type~\textnormal{(II)} cross-boundary semantic mismapping, respectively, as defined in Rem.\ref{rem:mismatch_type}.}
    \label{tab:cycle_validation}
    \setlength{\tabcolsep}{6pt}
    \renewcommand{\arraystretch}{1.15}
    \begin{tabular}{@{}llrrrr@{}}
        \toprule
        Dataset (Backbone) & $\Delta_{\mathrm{all}}$ & $N_u$
        & \shortstack{Total cycle margin\\$\sum_{\gamma\in\Gamma(\mathcal A_{\mathrm{GH}},\mathcal A_0)}m_{\gamma}$}
        & \shortstack{Old$\rightarrow$wrong-old\\mismapping (Type I)}
        & \shortstack{Old$\rightarrow$new\\mismapping (Type II)} \\
        \midrule
        CIFAR-10 (DINO) & 0.00 (Zero) & 37{,}500 & $0$        & 0 & 0   \\
        CUB (CLIP) & 0.22 (Low) & 4{,}496  & $-10$      & 1 & 3   \\
        Stanford Cars (CLIP) & 1.43 (Medium) & 6{,}144  & $-88$      & 0 & 6   \\
        Herbarium 19 (CLIP) & 12.13 (High) & 25{,}356 & $-3{,}076$ & 4 & 119 \\
        \bottomrule
    \end{tabular}
\end{table*}

\begin{table*}[!t]
    \centering
    \scriptsize
    \caption{Comprehensive comparison on four fine-grained benchmarks under GH and SPE. Adding MMO (\textit{i.e.}, \textbf{+Ours}) helps improve new-category recognition and old--new balance under both GH and SPE, with particularly large gains on Herbarium 19. Moreover, MMO yields larger average gains under SPE than under GH, indicating better alignment with our semantics-preserving evaluation, as the answer to \textbf{RQ2}.}
    \label{tab:main_results}
    \setlength{\tabcolsep}{1.8pt}
    \renewcommand{\arraystretch}{1.10}
    \begin{tabularx}{\textwidth}{@{}l@{\hspace{6pt}}ll*{2}{*{2}{>{\hsize=0.65\hsize\linewidth=\hsize\centering\arraybackslash}X}*{2}{>{\hsize=1.35\hsize\linewidth=\hsize\centering\arraybackslash}X}}@{}}
        \toprule
        \multirow{2}{*}{Dataset} & \multirow{2}{*}{Method} & \multirow{2}{*}{Backbone}
        & \multicolumn{4}{c}{GH}
        & \multicolumn{4}{c}{SPE} \\
        \cmidrule(lr){4-7}\cmidrule(lr){8-11}
        & & & $\texttt{All}_{\mathrm{GH}}$ & $\texttt{Old}_{\mathrm{GH}}$ & $\texttt{New}_{\mathrm{GH}}$ & $\texttt{H}_{\mathrm{GH}}$
        & $\texttt{All}_{\mathrm{SPE}}$ & $\texttt{Old}_{\mathrm{SPE}}$ & $\texttt{New}_{\mathrm{SPE}}$ & $\texttt{H}_{\mathrm{SPE}}$ \\
        \midrule
        & LegoGCD & DINO & 62.19 & 71.18 & 57.69  & 63.73 & 61.50 & 73.05 & 55.72 & 63.22 \\
        \rowcolor{black!2}\cellcolor{white}& \quad\textbf{+ Ours} & DINO & \textbf{63.88} & \textbf{74.85} & \SPEgain{\textbf{58.39}}{+0.70} & \SPEgain{\textbf{65.60}}{+1.87} & \textbf{63.63} & \textbf{74.72} & \SPEgain{\textbf{58.09}}{+2.37} & \SPEgain{\textbf{65.36}}{+2.14} \\ 
        & SimGCD-BIA & DINO & 61.61 & 60.84 & \textbf{62.00}  & 61.41 & 60.90 & 60.91 & \textbf{60.89} & 60.90 \\
        \rowcolor{black!2}\cellcolor{white}& \quad\textbf{+ Ours} & DINO & \textbf{61.81} & \textbf{62.17} & \SPEgain{61.63}{-0.37} & \SPEgain{\textbf{61.90}}{+0.49} & \textbf{61.17} & \textbf{62.91} & \SPEgain{60.29}{-0.60} & \SPEgain{\textbf{61.57}}{+0.67} \\
        & ProtoGCD & DINO & 60.97 & 64.64 & 59.13 & 61.76 & 60.32 & 64.78 & 58.09 & 61.25 \\
        \rowcolor{black!2}\cellcolor{white}& \quad\textbf{+ Ours} & DINO & \textbf{61.08} & \textbf{64.84} & \SPEgain{\textbf{59.19}}{+0.06} & \SPEgain{\textbf{61.89}}{+0.13} & \textbf{60.50} & \textbf{64.98} & \SPEgain{\textbf{58.26}}{+0.17} & \SPEgain{\textbf{61.44}}{+0.19} \\ 
        & SimGCD & DINO & 60.94 & \textbf{66.64} & 58.09  & \textbf{62.07} & 60.61 & 65.04 & 58.39 & 61.54 \\
        \rowcolor{black!2}\cellcolor{white}& \quad\textbf{+ Ours} & DINO & \textbf{61.43} & 63.91 & \SPEgain{\textbf{60.19}}{+2.10} & \SPEgain{61.99}{-0.08} & \textbf{61.14} & \textbf{65.64} & \SPEgain{\textbf{58.89}}{+0.50} & \SPEgain{\textbf{62.08}}{+0.54} \\ 
        & SimGCD & CLIP & 68.53 & 75.38 & 65.10  & 69.86 & 68.31 & \textbf{76.12} & 64.40 & 69.77 \\
        \rowcolor{black!2}\cellcolor{white}& \quad\textbf{+ Ours} & CLIP & \textbf{72.69} & \textbf{76.72} & \SPEgain{\textbf{70.67}}{+5.57} & \SPEgain{\textbf{73.57}}{+3.71} & \textbf{72.26} & 75.78 & \SPEgain{\textbf{70.50}}{+6.10} & \SPEgain{\textbf{73.04}}{+3.27} \\ 
        & GET-TAR & CLIP & 74.60 & \textbf{77.79} & 73.01  & \textbf{75.32} & 74.31 & \textbf{78.79} & 72.07 & \textbf{75.28} \\
        \rowcolor{black!2}\cellcolor{white}& \quad\textbf{+ Ours} & CLIP & \textbf{74.89} & 74.32 & \SPEgain{\textbf{75.18}}{+2.17} & \SPEgain{74.75}{-0.57} & \textbf{74.62} & 75.78 & \SPEgain{\textbf{74.04}}{+1.97} & \SPEgain{74.90}{-0.38} \\ 
        & GET & CLIP & 75.47 & \textbf{76.45} & 74.97  & 75.70 & 75.44 & \textbf{76.85} & 74.74 & 75.78 \\
        \rowcolor{black!2}\cellcolor{white}\multirow{-14}{*}{\textbf{CUB}} & \quad\textbf{+ Ours} & CLIP & \textbf{76.89} & 75.72 & \SPEgain{\textbf{77.48}}{+2.51} & \SPEgain{\textbf{76.59}}{+0.89} & \textbf{76.18} & 76.78 & \SPEgain{\textbf{75.88}}{+1.14} & \SPEgain{\textbf{76.33}}{+0.55} \\
        \midrule
        & LegoGCD & DINO & \textbf{54.39} & \textbf{62.12} & \textbf{50.52}  & \textbf{55.72} & \textbf{53.25} & \textbf{64.71} & \textbf{47.53} & \textbf{54.81} \\
        \rowcolor{black!2}\cellcolor{white}& \quad\textbf{+ Ours} & DINO & 52.51 & 58.88 & \SPEgain{49.33}{-1.19} & \SPEgain{53.68}{-2.04} & 51.79 & 61.40 & \SPEgain{46.99}{-0.54} & \SPEgain{53.24}{-1.57} \\ 
        & SimGCD-BIA & DINO & 52.09 & 55.46 & 50.40  & 52.81 & 51.99 & 56.30 & 49.84 & 52.87 \\
        \rowcolor{black!2}\cellcolor{white}& \quad\textbf{+ Ours} & DINO & \textbf{54.15} & \textbf{58.70} & \SPEgain{\textbf{51.87}}{+1.47} & \SPEgain{\textbf{55.07}}{+2.26} & \textbf{53.83} & \textbf{60.50} & \SPEgain{\textbf{50.49}}{+0.65} & \SPEgain{\textbf{55.04}}{+2.17} \\ 
        & ProtoGCD & DINO & 51.87 & 46.40 & \textbf{54.60} & 50.17 & 51.63 & 46.04 & \textbf{54.42} & 49.88 \\
        \rowcolor{black!2}\cellcolor{white}& \quad\textbf{+ Ours} & DINO & \textbf{53.43} & \textbf{59.72} & \SPEgain{50.28}{-4.32} & \SPEgain{\textbf{54.59}}{+4.42} & \textbf{53.35} & \textbf{61.76} & \SPEgain{49.15}{-5.27} & \SPEgain{\textbf{54.74}}{+4.86} \\ 
        & SimGCD & DINO & 53.31 & 58.58 & 50.67  & 54.34 & 53.15 & \textbf{60.14} & 49.66 & 54.40 \\
        \rowcolor{black!2}\cellcolor{white}& \quad\textbf{+ Ours} & DINO & \textbf{54.55} & \textbf{59.84} & \SPEgain{\textbf{51.90}}{+1.23} & \SPEgain{\textbf{55.59}}{+1.25} & \textbf{54.39} & 59.96 & \SPEgain{\textbf{51.60}}{+1.94} & \SPEgain{\textbf{55.47}}{+1.07} \\ 
        & SimGCD & CLIP & 55.59 & 59.54 & 53.61  & 56.42 & 55.47 & 59.54 & 53.43 & 56.32 \\
        \rowcolor{black!2}\cellcolor{white}& \quad\textbf{+ Ours} & CLIP & \textbf{60.41} & \textbf{60.92} & \SPEgain{\textbf{60.15}}{+6.54} & \SPEgain{\textbf{60.53}}{+4.11} & \textbf{59.61} & \textbf{62.42} & \SPEgain{\textbf{58.20}}{+4.77} & \SPEgain{\textbf{60.24}}{+3.92} \\ 
        & GET-TAR & CLIP & 60.37 & \textbf{64.59} & 58.26  & \textbf{61.26} & 60.19 & \textbf{66.39} & 57.09 & \textbf{61.39} \\
        \rowcolor{black!2}\cellcolor{white}& \quad\textbf{+ Ours} & CLIP & \textbf{60.47} & 62.00 & \SPEgain{\textbf{59.70}}{+1.44} & \SPEgain{60.83}{-0.43} & \textbf{60.31} & 62.85 & \SPEgain{\textbf{59.04}}{+1.95} & \SPEgain{60.89}{-0.50} \\ 
        & GET & CLIP & 60.31 & 64.59 & 58.17  & 61.21 & 59.69 & \textbf{67.65} & 55.71 & 61.10 \\
        \rowcolor{black!2}\cellcolor{white}\multirow{-14}{*}{\textbf{FGVC-Aircraft}} & \quad\textbf{+ Ours} & CLIP & \textbf{60.89} & \textbf{65.85} & \SPEgain{\textbf{58.41}}{+0.24} & \SPEgain{\textbf{61.91}}{+0.70} & \textbf{60.71} & 65.31 & \SPEgain{\textbf{58.41}}{+2.70} & \SPEgain{\textbf{61.67}}{+0.57} \\
        \midrule
        & LegoGCD & DINO & \textbf{44.82} & 58.47 & \textbf{37.48}  & \textbf{45.68} & 35.83 & 69.02 & 17.98 & 28.53 \\
        \rowcolor{black!2}\cellcolor{white}& \quad\textbf{+ Ours} & DINO & 44.51 & \textbf{58.87} & \SPEgain{36.79}{-0.69} & \SPEgain{45.28}{-0.40} & \textbf{36.53} & \textbf{69.49} & \SPEgain{\textbf{18.80}}{+0.82} & \SPEgain{\textbf{29.59}}{+1.06} \\ 
        & SimGCD-BIA & DINO & 42.53 & 55.30 & \textbf{35.65}  & 43.35 & 33.53 & \textbf{65.47} & 16.34 & 26.15 \\
        \rowcolor{black!2}\cellcolor{white}& \quad\textbf{+ Ours} & DINO & \textbf{42.77} & \textbf{56.27} & \SPEgain{35.51}{-0.14} & \SPEgain{\textbf{43.54}}{+0.19} & \textbf{34.10} & 65.37 & \SPEgain{\textbf{17.28}}{+0.94} & \SPEgain{\textbf{27.33}}{+1.18} \\ 
        & ProtoGCD & DINO & 43.74 & \textbf{58.28} & 35.92 & 44.45 & 35.84 & \textbf{68.04} & 18.52 & 29.12 \\
        \rowcolor{black!2}\cellcolor{white}& \quad\textbf{+ Ours} & DINO & \textbf{44.31} & 55.60 & \SPEgain{\textbf{38.24}}{+2.32} & \SPEgain{\textbf{45.31}}{+0.86} & \textbf{41.56} & 60.97 & \SPEgain{\textbf{31.12}}{+12.60} & \SPEgain{\textbf{41.21}}{+12.09} \\ 
        & SimGCD & DINO & \textbf{43.98} & \textbf{57.47} & 36.72  & \textbf{44.81} & 34.95 & \textbf{67.09} & 17.66 & 27.96 \\
        \rowcolor{black!2}\cellcolor{white}& \quad\textbf{+ Ours} & DINO & 43.74 & 47.34 & \SPEgain{\textbf{41.81}}{+5.09} & \SPEgain{44.40}{-0.41} & \textbf{42.89} & 48.57 & \SPEgain{\textbf{39.83}}{+22.17} & \SPEgain{\textbf{43.77}}{+15.81} \\ 
        & SimGCD & CLIP & 47.40 & \textbf{64.39} & 38.26  & 48.00 & 35.27 & \textbf{77.34} & 12.63 & 21.71 \\
        \rowcolor{black!2}\cellcolor{white}& \quad\textbf{+ Ours} & CLIP & \textbf{49.14} & 56.91 & \SPEgain{\textbf{44.96}}{+6.70} & \SPEgain{\textbf{50.23}}{+2.23} & \textbf{48.58} & 57.90 & \SPEgain{\textbf{43.56}}{+30.93} & \SPEgain{\textbf{49.72}}{+28.01} \\ 
        & GET-TAR & CLIP & 47.33 & \textbf{65.08} & 37.77  & 47.80 & 38.20 & \textbf{78.59} & 16.46 & 27.22 \\
        \rowcolor{black!2}\cellcolor{white}& \quad\textbf{+ Ours} & CLIP & \textbf{53.83} & 64.86 & \SPEgain{\textbf{47.90}}{+10.13} & \SPEgain{\textbf{55.10}}{+7.30} & \textbf{52.72} & 67.45 & \SPEgain{\textbf{44.80}}{+28.34} & \SPEgain{\textbf{53.84}}{+26.62} \\
        & GET & CLIP & 49.03 & \textbf{63.75} & 41.11  & 49.99 & 38.59 & \textbf{78.52} & 17.10 & 28.08 \\
        \rowcolor{black!2}\cellcolor{white}\multirow{-14}{*}{\textbf{Herbarium 19}} & \quad\textbf{+ Ours} & CLIP & \textbf{51.33} & 61.23 & \SPEgain{\textbf{46.00}}{+4.89} & \SPEgain{\textbf{52.53}}{+2.54} & \textbf{50.58} & 63.08 & \SPEgain{\textbf{43.86}}{+26.76} & \SPEgain{\textbf{51.74}}{+23.66} \\ 
        \midrule
        & LegoGCD & DINO & 54.02 & \textbf{75.11} & 43.83  & 55.36 & 53.32 & 79.41 & 40.72 & 53.83 \\
        \rowcolor{black!2}\cellcolor{white}& \quad\textbf{+ Ours} & DINO & \textbf{56.12} & 73.96 & \SPEgain{\textbf{47.50}}{+3.67} & \SPEgain{\textbf{57.85}}{+2.49} & \textbf{54.85} & \textbf{80.71} & \SPEgain{\textbf{42.36}}{+1.64} & \SPEgain{\textbf{55.56}}{+1.73} \\ 
        & SimGCD-BIA & DINO & 53.65 & \textbf{71.21} & 45.16  & 55.27 & 52.85 & \textbf{74.91} & 42.19 & 53.98 \\
        \rowcolor{black!2}\cellcolor{white}& \quad\textbf{+ Ours} & DINO & \textbf{55.11} & 70.86 & \SPEgain{\textbf{47.50}}{+2.34} & \SPEgain{\textbf{56.87}}{+1.60} & \textbf{54.72} & 72.26 & \SPEgain{\textbf{46.25}}{+4.06} & \SPEgain{\textbf{56.40}}{+2.42} \\ 
        & ProtoGCD & DINO & 52.29 & 66.97 & 45.21 & 53.98 & 51.89 & 69.07 & 43.59 & 53.45 \\
        \rowcolor{black!2}\cellcolor{white}& \quad\textbf{+ Ours} & DINO & \textbf{52.44} & \textbf{67.07} & \SPEgain{\textbf{45.38}}{+0.17} & \SPEgain{\textbf{54.13}}{+0.15} & \textbf{52.07} & \textbf{69.17} & \SPEgain{\textbf{43.81}}{+0.22} & \SPEgain{\textbf{53.64}}{+0.19} \\ 
        & SimGCD & DINO & 53.53 & 72.01 & 44.61  & 55.09 & 52.93 & 74.11 & 42.70 & 54.18 \\
        \rowcolor{black!2}\cellcolor{white}& \quad\textbf{+ Ours} & DINO & \textbf{55.34} & \textbf{73.86} & \SPEgain{\textbf{46.39}}{+1.78} & \SPEgain{\textbf{56.99}}{+1.90} & \textbf{55.16} & \textbf{75.76} & \SPEgain{\textbf{45.21}}{+2.51} & \SPEgain{\textbf{56.63}}{+2.45} \\ 
        & SimGCD & CLIP & 70.46 & \textbf{83.71} & 64.06  & 72.58 & 69.03 & \textbf{89.01} & 59.38 & 71.24 \\
        \rowcolor{black!2}\cellcolor{white}& \quad\textbf{+ Ours} & CLIP & \textbf{76.11} & 83.46 & \SPEgain{\textbf{72.56}}{+8.50} & \SPEgain{\textbf{77.63}}{+5.05} & \textbf{75.83} & 84.26 & \SPEgain{\textbf{71.76}}{+12.38} & \SPEgain{\textbf{77.51}}{+6.27} \\ 
        & GET-TAR & CLIP & 75.96 & 82.51 & 72.80  & 77.35 & 71.99 & \textbf{95.75} & 60.51 & 74.16 \\
        \rowcolor{black!2}\cellcolor{white}& \quad\textbf{+ Ours} & CLIP & \textbf{79.48} & \textbf{85.64} & \SPEgain{\textbf{76.51}}{+3.71} & \SPEgain{\textbf{80.82}}{+3.47} & \textbf{78.68} & 88.46 & \SPEgain{\textbf{73.96}}{+13.45} & \SPEgain{\textbf{80.56}}{+6.40} \\ 
        & GET & CLIP & 78.96 & 88.01 & 74.58  & 80.74 & 77.77 & \textbf{92.30} & 70.75 & 80.10 \\
        \rowcolor{black!2}\cellcolor{white}\multirow{-14}{*}{\textbf{Stanford Cars}} & \quad\textbf{+ Ours} & CLIP & \textbf{82.23} & \textbf{88.31} & \SPEgain{\textbf{79.29}}{+4.71} & \SPEgain{\textbf{83.56}}{+2.82} & \textbf{81.62} & 90.00 & \SPEgain{\textbf{77.58}}{+6.83} & \SPEgain{\textbf{83.33}}{+3.23} \\ 
        \bottomrule
    \end{tabularx}
\end{table*}

In this section, we conduct experiments to answer the following research questions:
\begin{itemize}
    \item \textbf{RQ1}: Can we empirically observe performance overestimation caused by GH in parametric GCD, and when is this overestimation more significant?
    \item \textbf{RQ2}: How effective is MMO under GH and SPE, and how do its loss components and coefficients affect performance?
    \item \textbf{RQ3}: Does SPE remain applicable when the new-category number is estimated and when the evaluated method is non-parametric?
\end{itemize}

\subsection{Experimental Setup}
\label{sec:experimental_setup}

\textbf{Datasets.}
We use eight standard GCD benchmarks, CIFAR-10 and CIFAR-100 \cite{cifar10}, ImageNet-100 and ImageNet-1K \cite{imagenet,DBLP:journals/ijcv/RussakovskyDSKS15}, CUB-200-2011 \cite{cub}, Stanford Cars \cite{stanfordcars}, FGVC-Aircraft \cite{fgvcair}, and Herbarium 19 \cite{HerbariumChallenge2019}. They cover coarse-grained, fine-grained, and large-scale recognition. All experiments follow the established GCD splits \cite{DBLP:conf/cvpr/VazeHVZ22,DBLP:conf/iccv/WenZQ23} as shown in Tab.\ref{tab:dataset}. Additional dataset details are provided in App.\ref{app:additional_datasets}.

\textbf{Methods.}
We evaluate and compare seven configurations under both GH and SPE. With DINO~\cite{DNIO} as the backbone, we evaluate four baselines: LegoGCD~\cite{DBLP:conf/cvpr/CaoZWYS0L024}, SimGCD-BIA~\cite{DBLP:conf/iccv/WenZQ23,tang2026bia}, ProtoGCD~\cite{ma2025protogcd}, and SimGCD~\cite{DBLP:conf/iccv/WenZQ23}. With CLIP~\cite{DBLP:conf/icml/RadfordKHRGASAM21} as the backbone, we evaluate SimGCD~\cite{DBLP:conf/iccv/WenZQ23}, GET~\cite{wang2025get}, and GET-TAR~\cite{yang2026tar}. \underline{\textbf{We apply the proposed MMO to all seven baselines}}. Descriptions of these methods are provided in App.\ref{app:additional_compared_methods}.

\textbf{Metrics.}
We report $\texttt{Old}_{\mathrm{GH}}$, $\texttt{New}_{\mathrm{GH}}$, and $\texttt{All}_{\mathrm{GH}}$ under Global Hungarian (GH), together with $\texttt{Old}_{\mathrm{SPE}}$, $\texttt{New}_{\mathrm{SPE}}$, $\texttt{All}_{\mathrm{SPE}}$, and $\texttt{H}_{\mathrm{SPE}}$ under our proposed SPE. For $r\in\{\mathrm{new},\mathrm{all}\}$, we use $\Delta_r:=\texttt{Acc}_{\mathrm{GH}}^r-\texttt{Acc}_{\mathrm{SPE}}^r$ to measure how much GH overestimates the model performance compared to SPE. Since GH optimizes over a superset of the assignments available to SPE, $\Delta_{\mathrm{all}} \geq 0$.

\textbf{Implementation details.}
We implement our experiments in PyTorch~\cite{paszke2019pytorch}. We follow the training pipeline of each baseline method and introduce MMO only through the auxiliary terms in Eq.\eqref{eq:pao_training_objective}. Specifically, \textbf{the SimGCD-based runs follow the original SimGCD optimization \cite{DBLP:conf/iccv/WenZQ23}}: batch size 128, 200 epochs, SGD with an initial learning rate of 0.1 and cosine decay. Its DINO instantiation uses a DINO-pretrained ViT-B/16~\cite{DBLP:conf/iclr/DosovitskiyB0WZ21} and fine-tunes only the final transformer block. SimGCD--CLIP uses the same SimGCD objective and optimization schedule, with the DINO backbone replaced by the CLIP visual encoder. \textbf{The GET-based runs follow the original two-stage CLIP ViT-B/16 pipeline \cite{wang2025get}}: each stage is trained for 200 epochs using SGD, a training batch size of 128, an initial learning rate of 0.1 with cosine annealing, and the original augmentation; evaluation uses a batch size of 256. Other baseline-specific hyperparameters follow the original implementations and are detailed in App.\ref{app:baseline_hyperparameters}. For MMO, the sensitivity of $\lambda_{\mathrm{RB}}$ and $\lambda_{\mathrm{ND}}$ is analyzed in Sec.\ref{sec:exp_optimization}. We use $\rho=0.5$ given by the GCD benchmark protocols and only the automatic-estimation experiment in Sec.\ref{sec:exp_optimization} uses the estimated value $\hat{\rho}$.

\subsection{Empirical Evidence of GH Overestimation (RQ1)}
\label{sec:exp_protocol_importance}
We first examine whether GH-induced performance overestimation is empirically observable, going beyond the illustrative example in Fig.\ref{fig:intro_bias_metric}. To this end, we evaluate the same predictions under GH and SPE, guaranteeing that the score differences arise only from semantic mismapping. As shown in Tab.\ref{tab:protocol_benchmark}, GH and SPE yield identical new and overall accuracies for all configurations on coarse-grained datasets, such as CIFAR-10 and ImageNet-100. On CIFAR-100, only SimGCD--CLIP shows a narrow gap, with $\Delta_{\mathrm{new}}=2.07$ and $\Delta_{\mathrm{all}}=0.12$. However, larger differences appear on fine-grained benchmarks, most clearly on Herbarium 19, where $\Delta_{\mathrm{new}}$ ranges from $17.40$ to $25.63$. What's worse, the ranking of these methods changes when evaluated under SPE. On Herbarium 19, switching from GH to SPE moves ProtoGCD--DINO from sixth to first, SimGCD--CLIP from second to seventh, and GET--CLIP from first to fourth. Similar rank reversals occur within the DINO configurations on CUB, Stanford Cars, and FGVC-Aircraft. All these results show that GH can significantly overestimate novel-category discovery, especially on fine-grained datasets, and that ignoring semantic mismapping can mislead the ranking of methods, as the answer to \textbf{RQ1}. Complete results under both protocols are provided in App. Tab.\ref{tab:appendix_gh_spe_complete}.

\textbf{Prediction-level cycle validation.}
To connect the observed overestimation to the analysis in Thm.\ref{thm:semantic_mismapping_margin}, we examine the final SimGCD predictions in four configurations with different GH--SPE gaps. Let $\Gamma(\mathcal A_{\mathrm{GH}},\mathcal A_0)$ denote the set of nontrivial permutation cycles relating the GH assignment $\mathcal A_{\mathrm{GH}}$ to the highest-scoring semantics-preserving assignment $\mathcal A_0$.
Tab.\ref{tab:cycle_validation} reports the summed cycle margins $\sum_{\gamma\in\Gamma(\mathcal A_{\mathrm{GH}},\mathcal A_0)}m_\gamma$ and the observed accuracy gaps $\Delta_{\mathrm{all}}$. In all four configurations, the negative margin $- m_{\gamma}$ summed over all the cycles $\gamma \in\Gamma$ exactly equals the number of additional samples counted as correct under GH compared with SPE, yielding
\begin{equation}
\Delta_{\mathrm{all}}=-\frac{100}{N_u}\sum_{\gamma\in\Gamma(\mathcal A_{\mathrm{GH}},\mathcal A_0)}m_{\gamma}.
\label{eq:cycle_gap_identity}
\end{equation}
This exact agreement provides direct empirical support for our cycle-margin analysis provided in Thm.\ref{thm:semantic_mismapping_margin}. The complete calculation procedure and individual cycle margins are provided in App.\ref{app:cycle_audit}.

Following the two types defined in Rem.\ref{rem:mismatch_type}, Tab.\ref{tab:cycle_validation} further reports how many anchored old outputs GH reassigns to wrong old categories, \textit{i.e.}, Type~\textnormal{(I)}, and to new categories, \textit{i.e.}, Type~\textnormal{(II)}. Neither type occurs in the zero-gap CIFAR-10 configuration. CUB exhibits both types, with $1$ and $3$ affected outputs, respectively, while Stanford Cars has $6$ Type~\textnormal{(II)} outputs and no Type~\textnormal{(I)} outputs. On Herbarium 19, Type~\textnormal{(II)} predominates, with $119$ affected outputs compared with $4$ for Type~\textnormal{(I)}, showing that most mismapped old outputs are reassigned across the old--new boundary.  

\subsection{Effectiveness and Analysis of MMO (RQ2)}
\label{sec:exp_optimization}
In this part, we evaluate whether MMO mitigates the prediction errors that can give rise to semantic mismapping. To this end, we assess its combined effect under both GH and SPE, then examine hyperparameter sensitivity and visual representations.

\begin{figure*}[!t]
    \centering
    \begin{minipage}{0.245\textwidth}
        \centering
        \includegraphics[width=\linewidth]{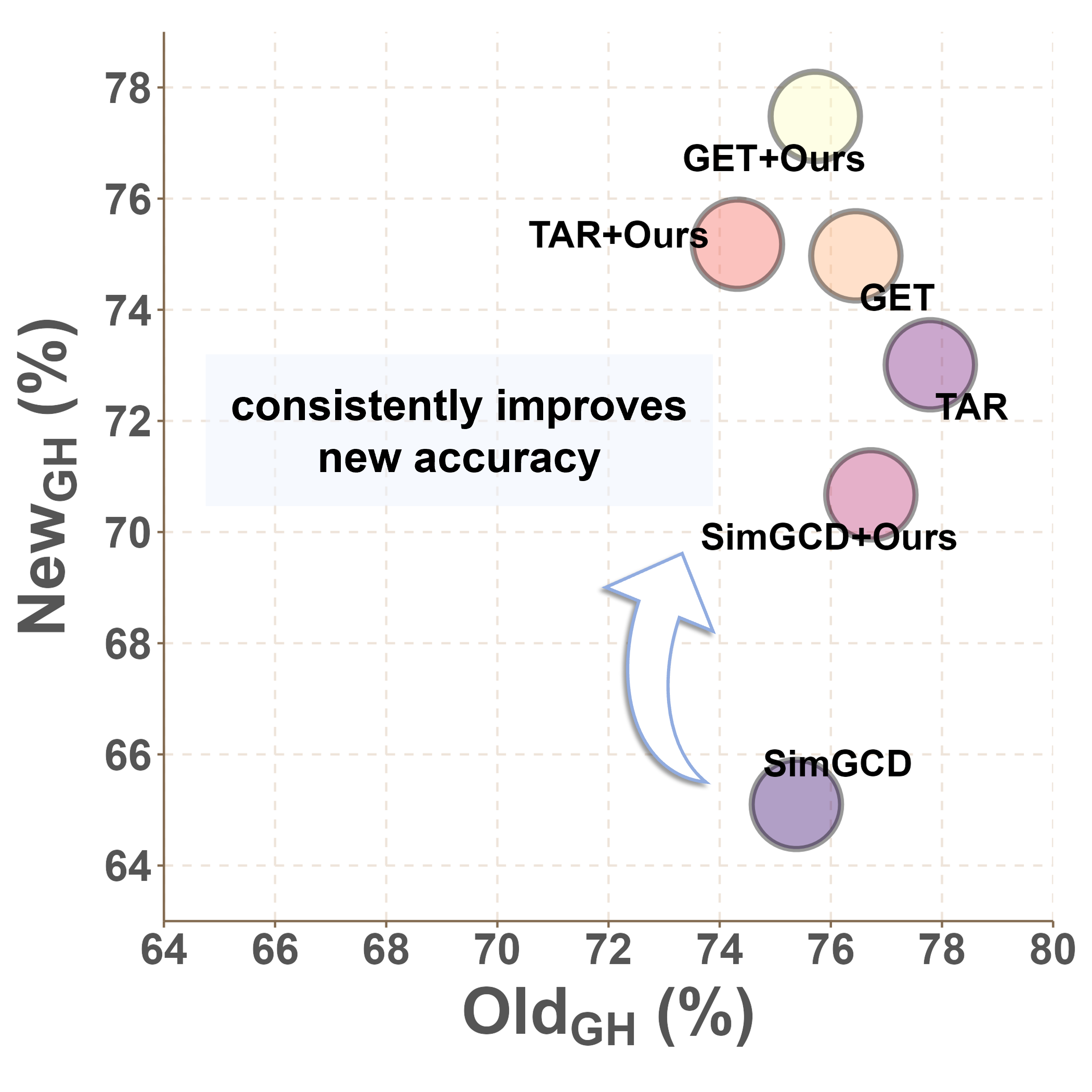}
        {\scriptsize (a) CUB (GH)}
    \end{minipage}\hfill
    \begin{minipage}{0.245\textwidth}
        \centering
        \includegraphics[width=\linewidth]{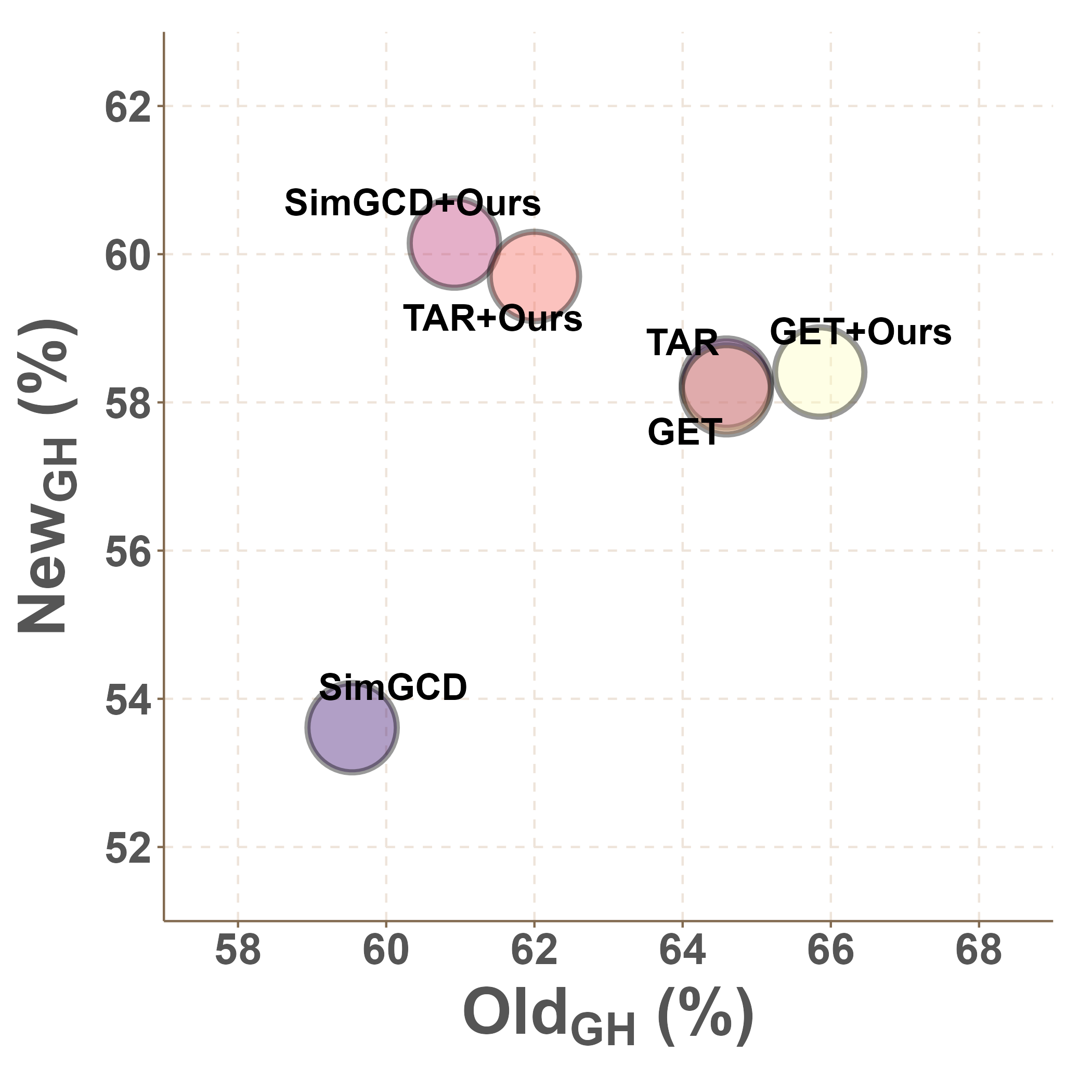}
        {\scriptsize (b) Aircraft (GH)}
    \end{minipage}\hfill
    \begin{minipage}{0.245\textwidth}
        \centering
        \includegraphics[width=\linewidth]{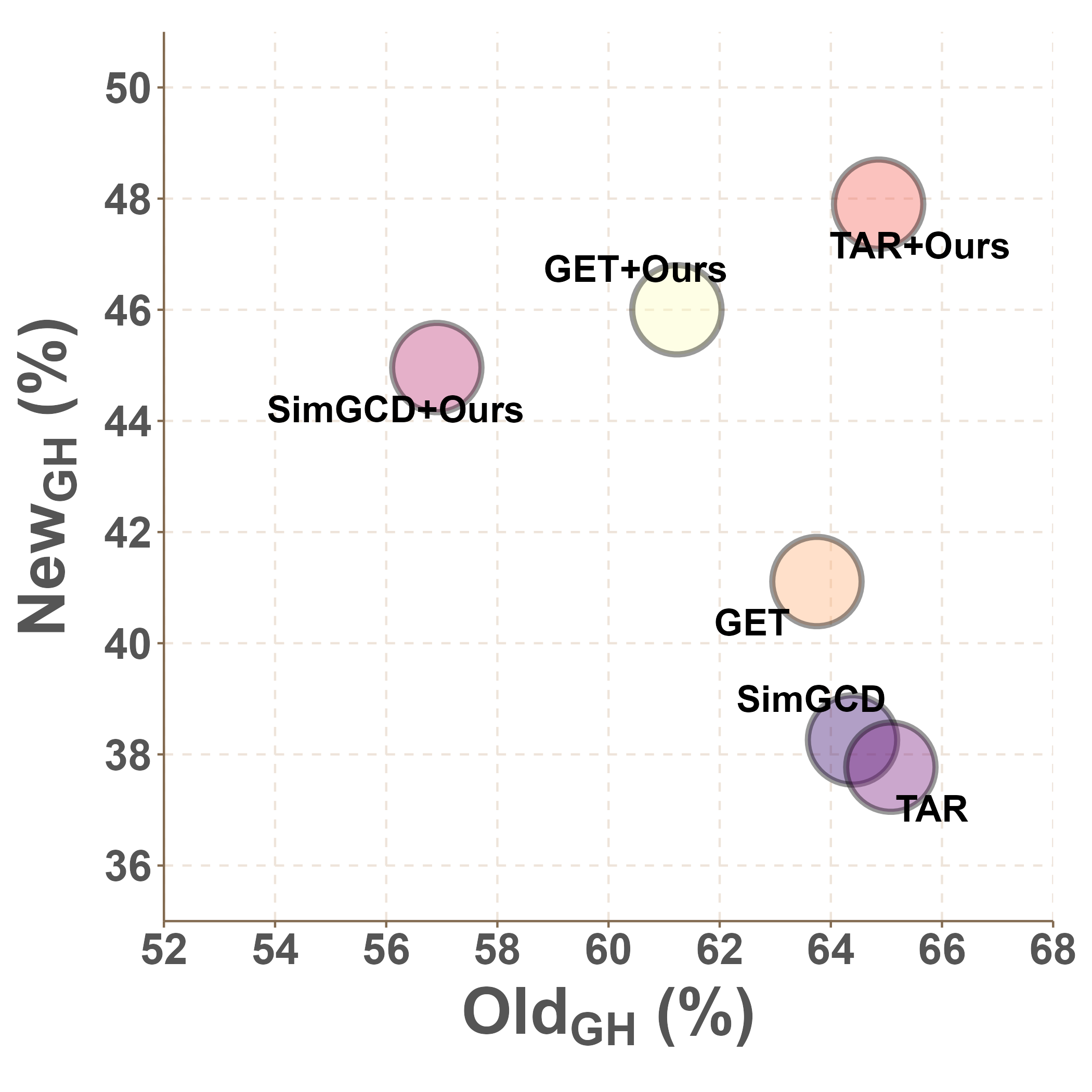}
        {\scriptsize (c) H19 (GH)}
    \end{minipage}\hfill
    \begin{minipage}{0.245\textwidth}
        \centering
        \includegraphics[width=\linewidth]{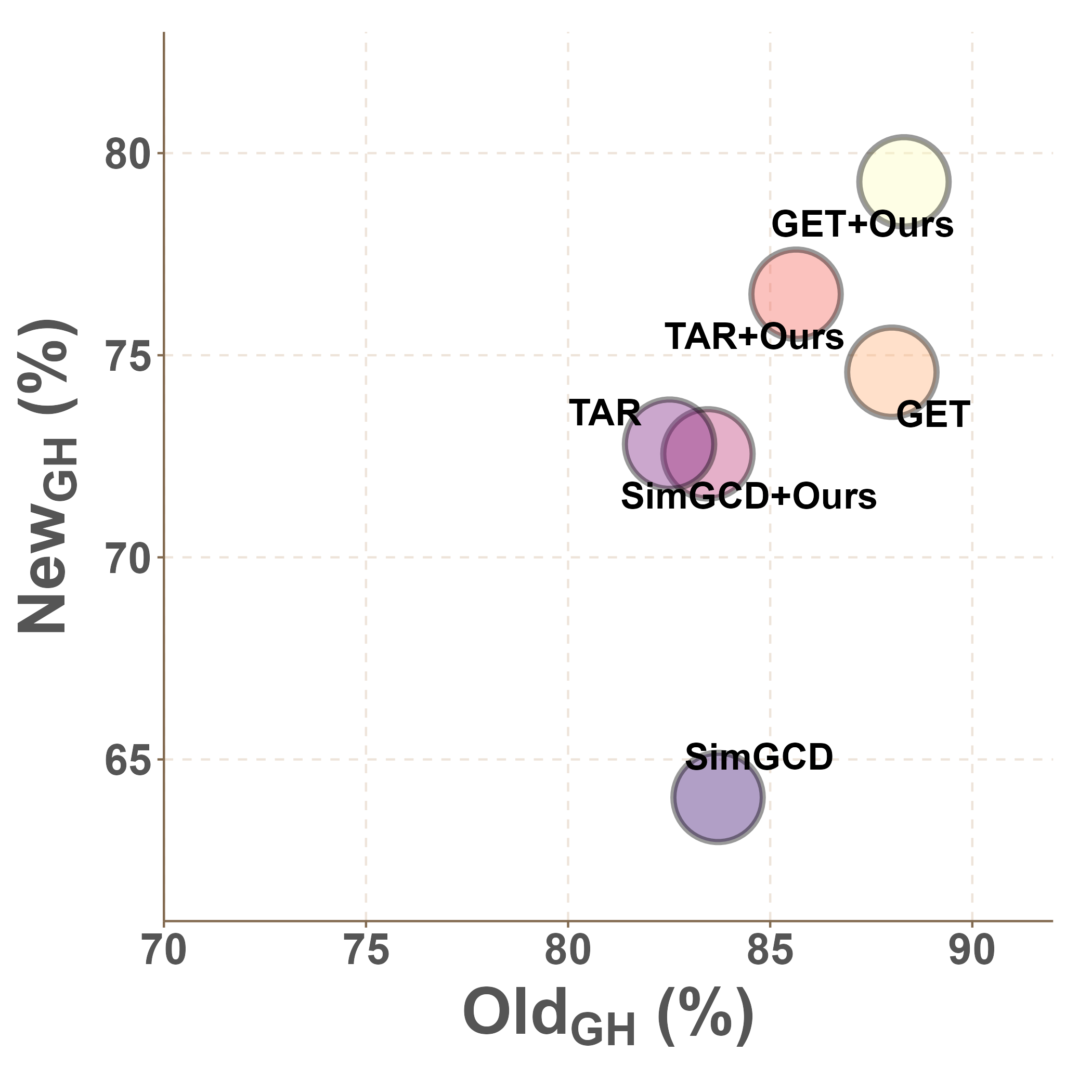}
        {\scriptsize (d) Cars (GH)}
    \end{minipage}
    \par\medskip
    \begin{minipage}{0.245\textwidth}
        \centering
        \includegraphics[width=\linewidth]{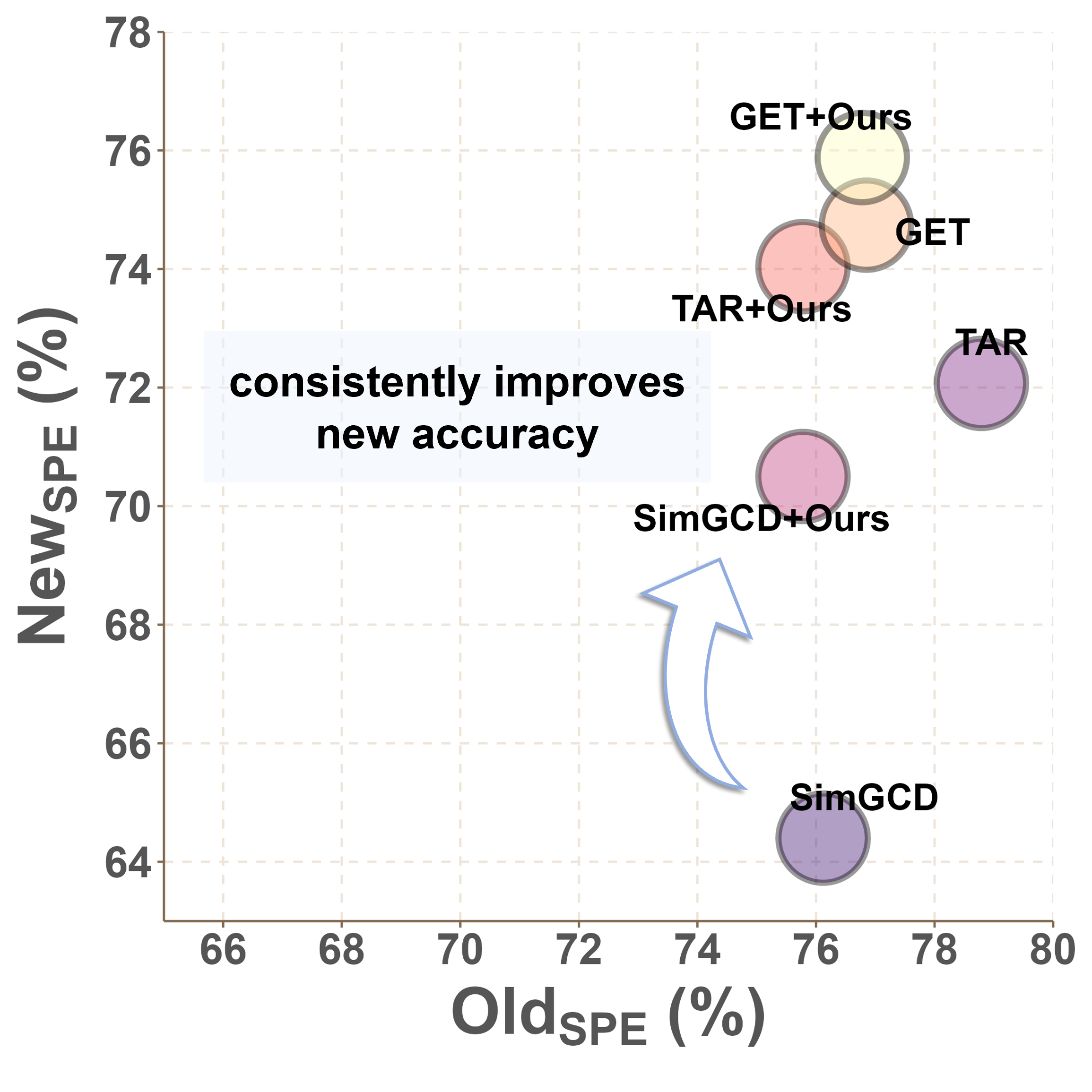}
        {\scriptsize (e) CUB (SPE)}
    \end{minipage}\hfill
    \begin{minipage}{0.245\textwidth}
        \centering
        \includegraphics[width=\linewidth]{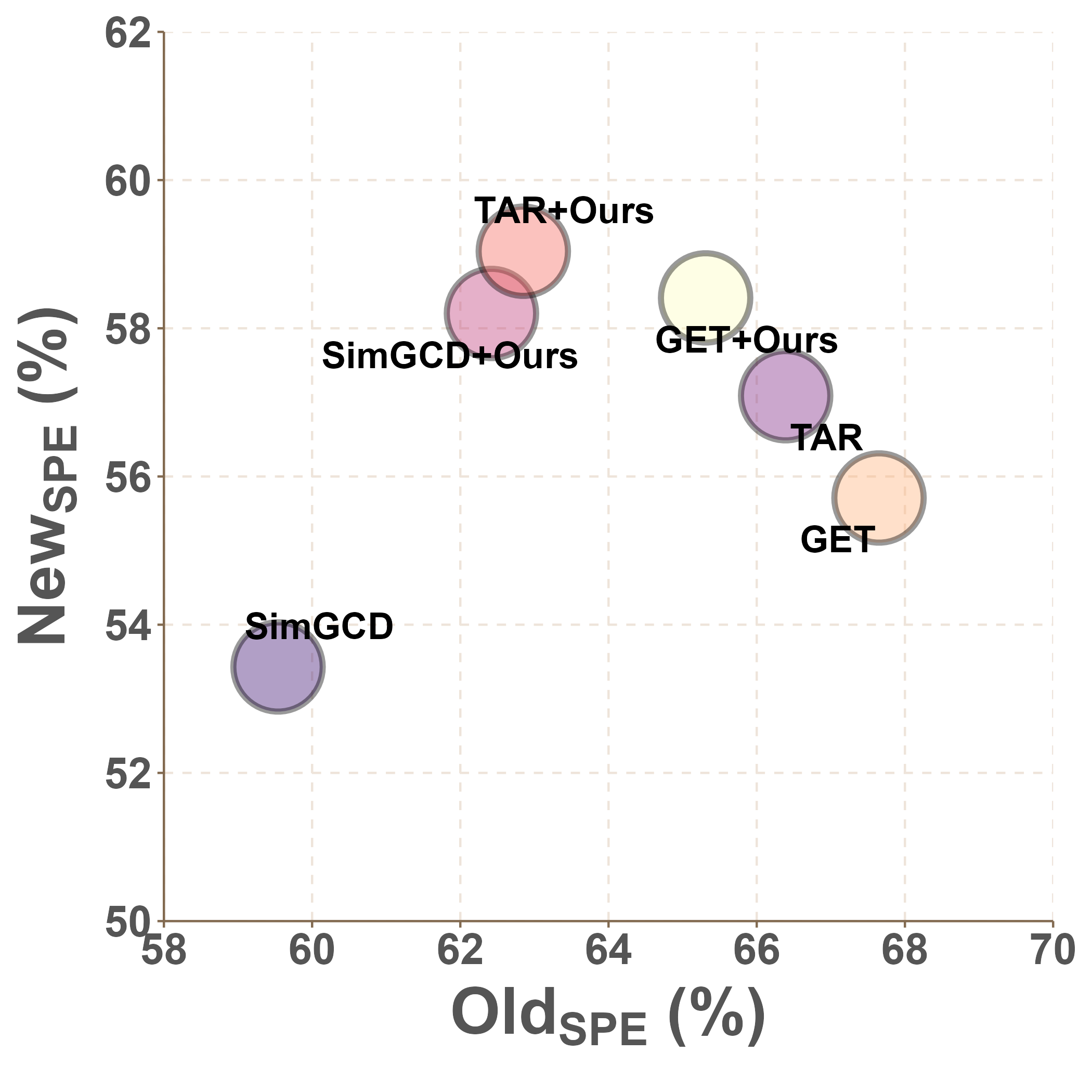}
        {\scriptsize (f) Aircraft (SPE)}
    \end{minipage}\hfill
    \begin{minipage}{0.245\textwidth}
        \centering
        \includegraphics[width=\linewidth]{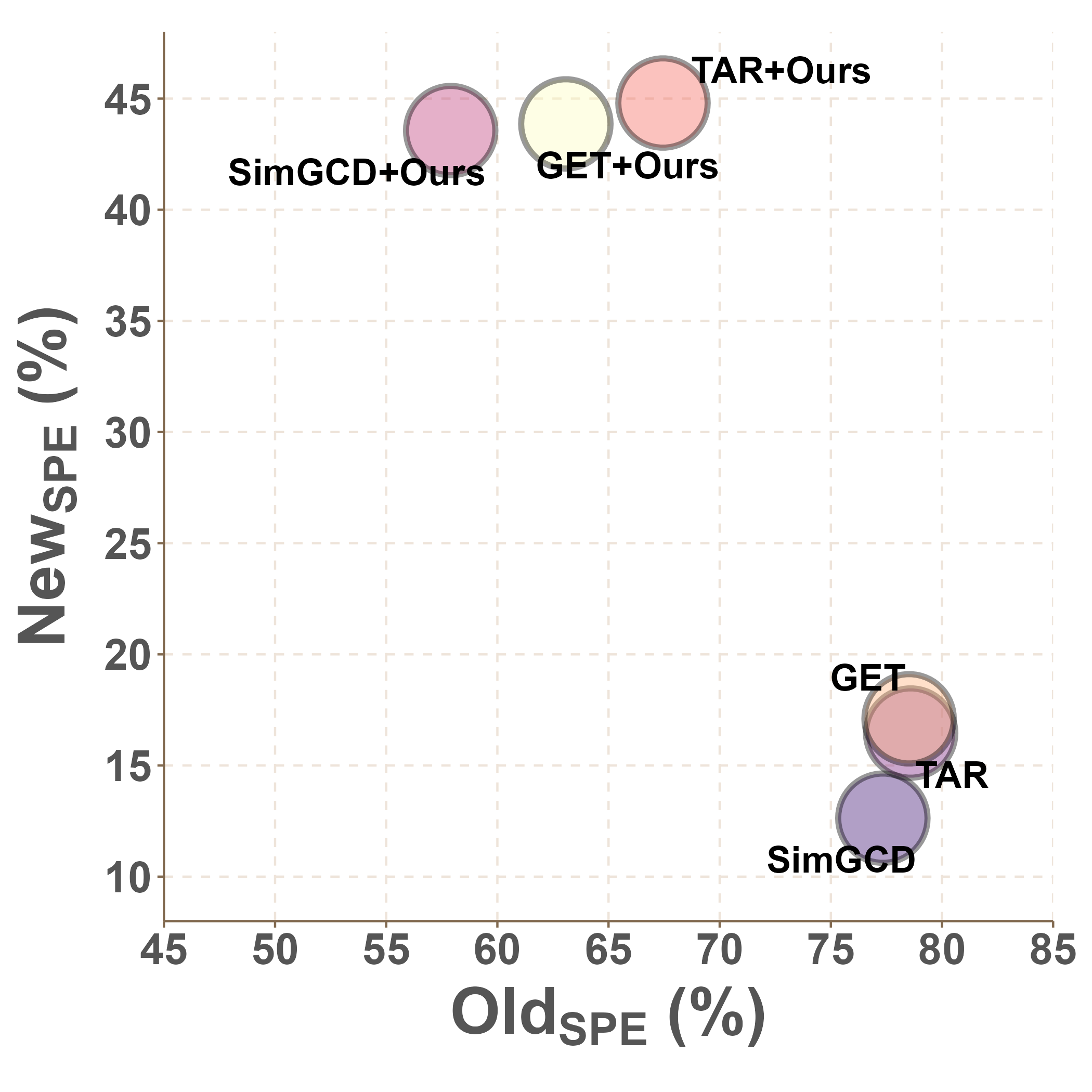}
        {\scriptsize (g) H19 (SPE)}
    \end{minipage}\hfill
    \begin{minipage}{0.245\textwidth}
        \centering
        \includegraphics[width=\linewidth]{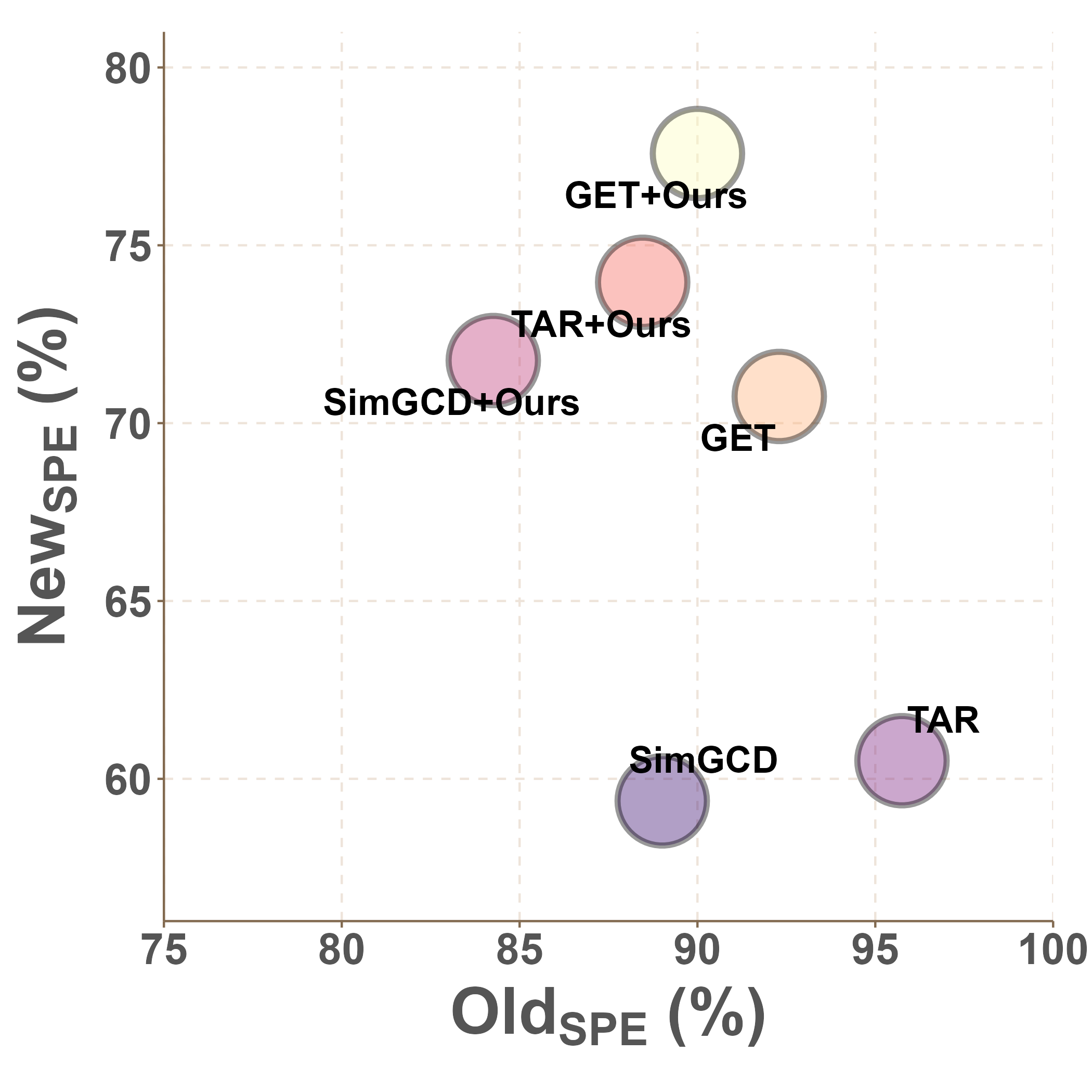}
        {\scriptsize (h) Cars (SPE)}
    \end{minipage}
    \caption{Old--new accuracy trade-offs for the CLIP-based methods in Tab.\ref{tab:main_results}. The top and bottom rows report GH and SPE, respectively; columns show CUB, FGVC-Aircraft, Herbarium 19, and Stanford Cars. TAR denotes GET-TAR, and \textbf{+Ours} denotes the corresponding method with MMO. }
    \label{fig:old_new_tradeoff}
\end{figure*}

\begin{figure*}[!t]
    \centering
    \includegraphics[width=0.247\textwidth]{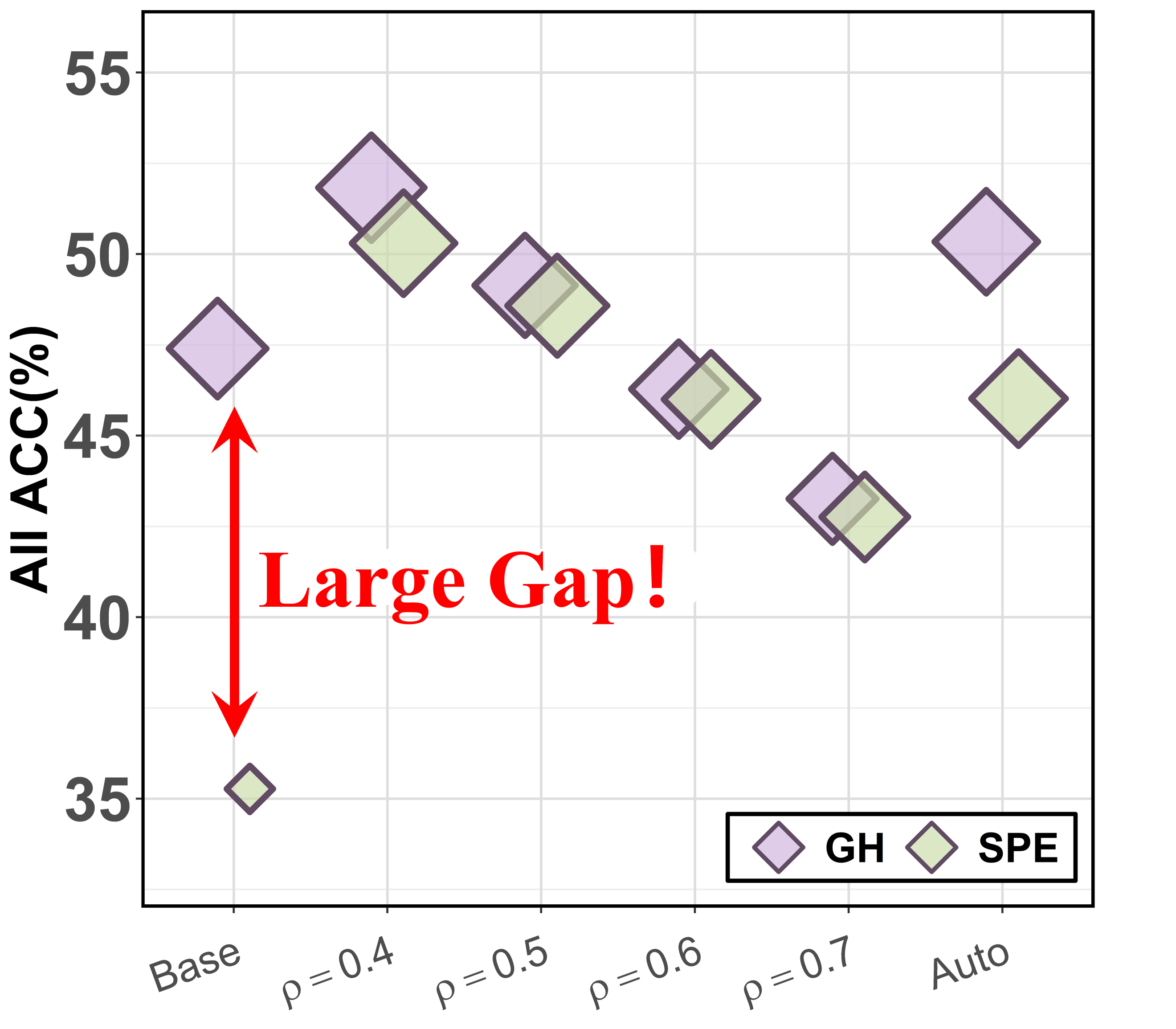}\hfill
    \includegraphics[width=0.247\textwidth]{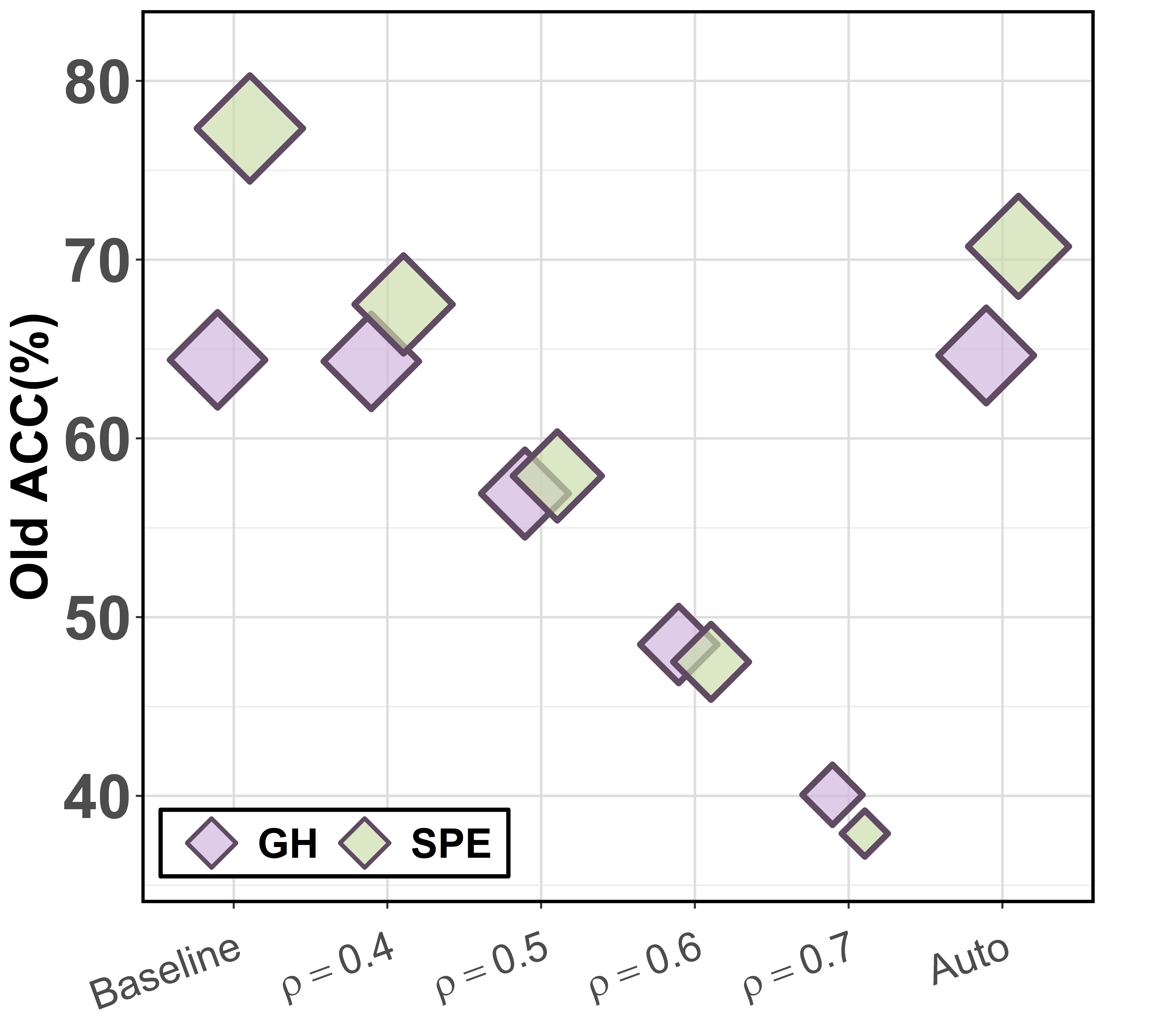}\hfill
    \includegraphics[width=0.247\textwidth]{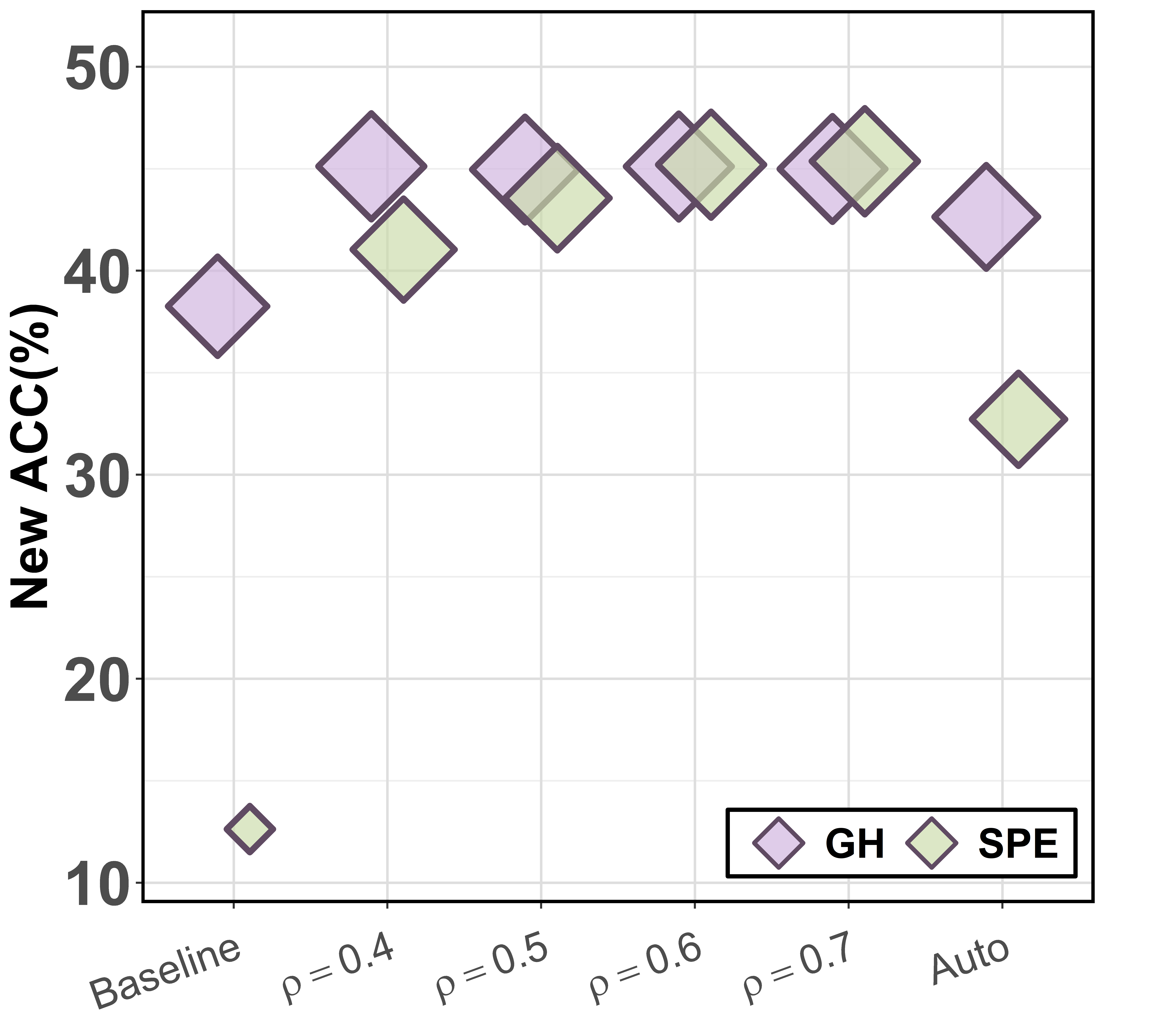}\hfill
    \includegraphics[width=0.247\textwidth]{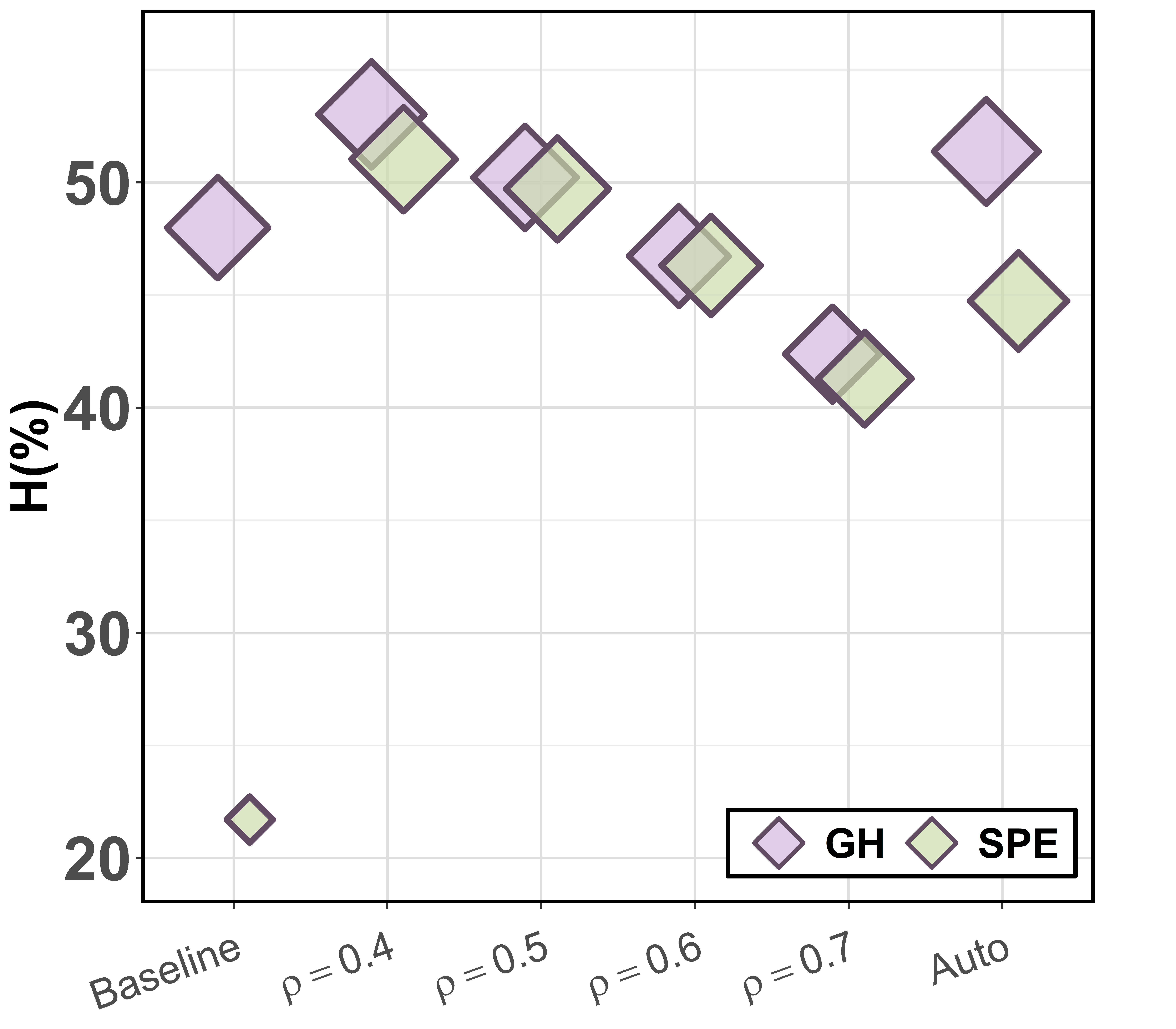}
    \caption{Sensitivity to the labeling fraction on the imbalanced Herbarium 19 benchmark with SimGCD--CLIP. From left to right, the panels show All, Old, New, and harmonic-mean ($H$) accuracy under GH and SPE. Baseline denotes the unmodified result, the middle points use manually specified $\widehat\rho$, and Auto uses the TIcE estimate. Complete numerical results are provided in App. Tab.\ref{tab:appendix_rho_sensitivity_full}.}
    \label{fig:rho_sensitivity}
\end{figure*}

\begin{figure}[!t]
    \centering
    \setlength{\tabcolsep}{0.5pt}
    \renewcommand{\arraystretch}{0.90}
    \begin{tabular}{@{}ccc@{}}
        \includegraphics[width=0.322\linewidth]{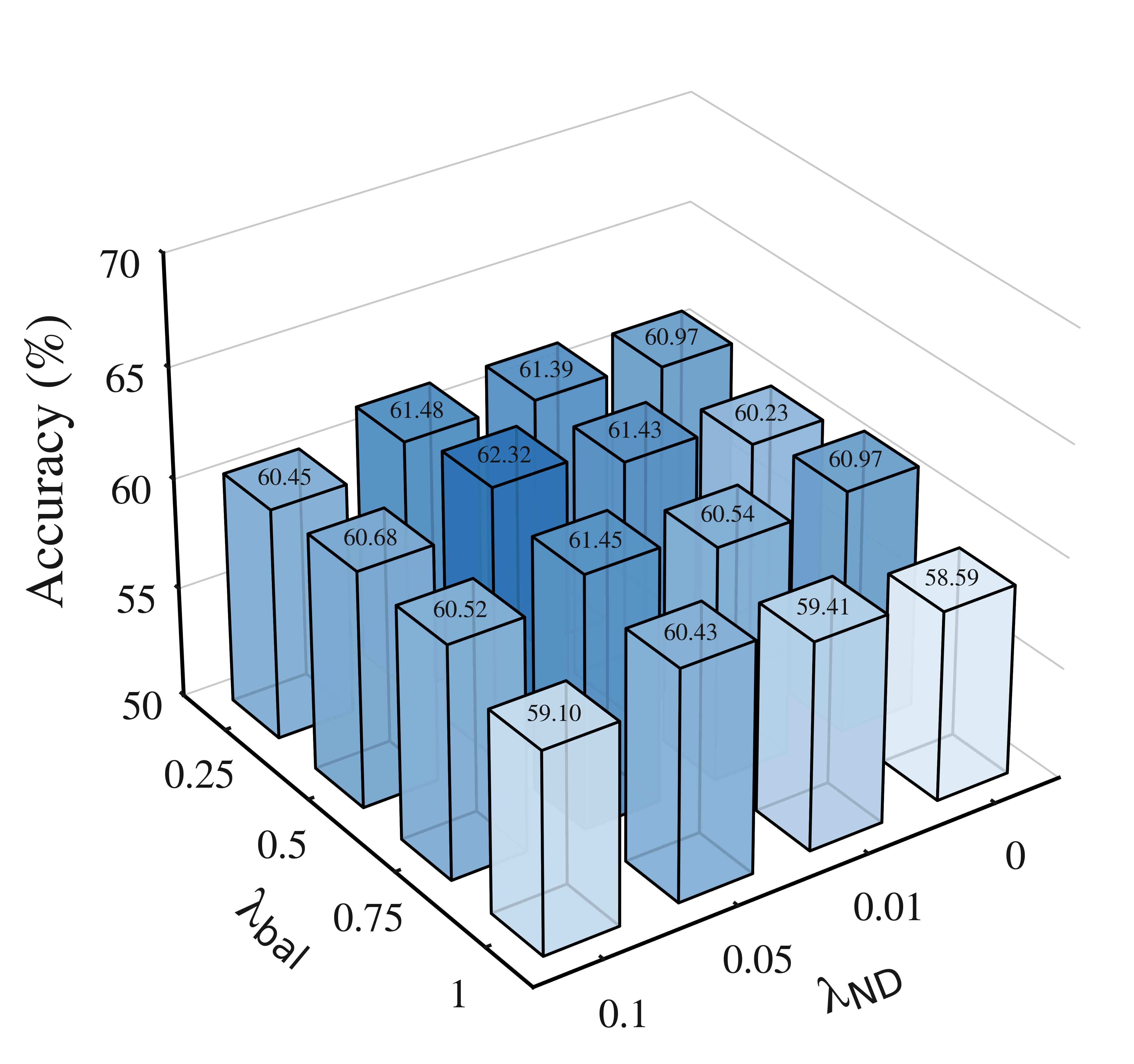} &
        \includegraphics[width=0.322\linewidth]{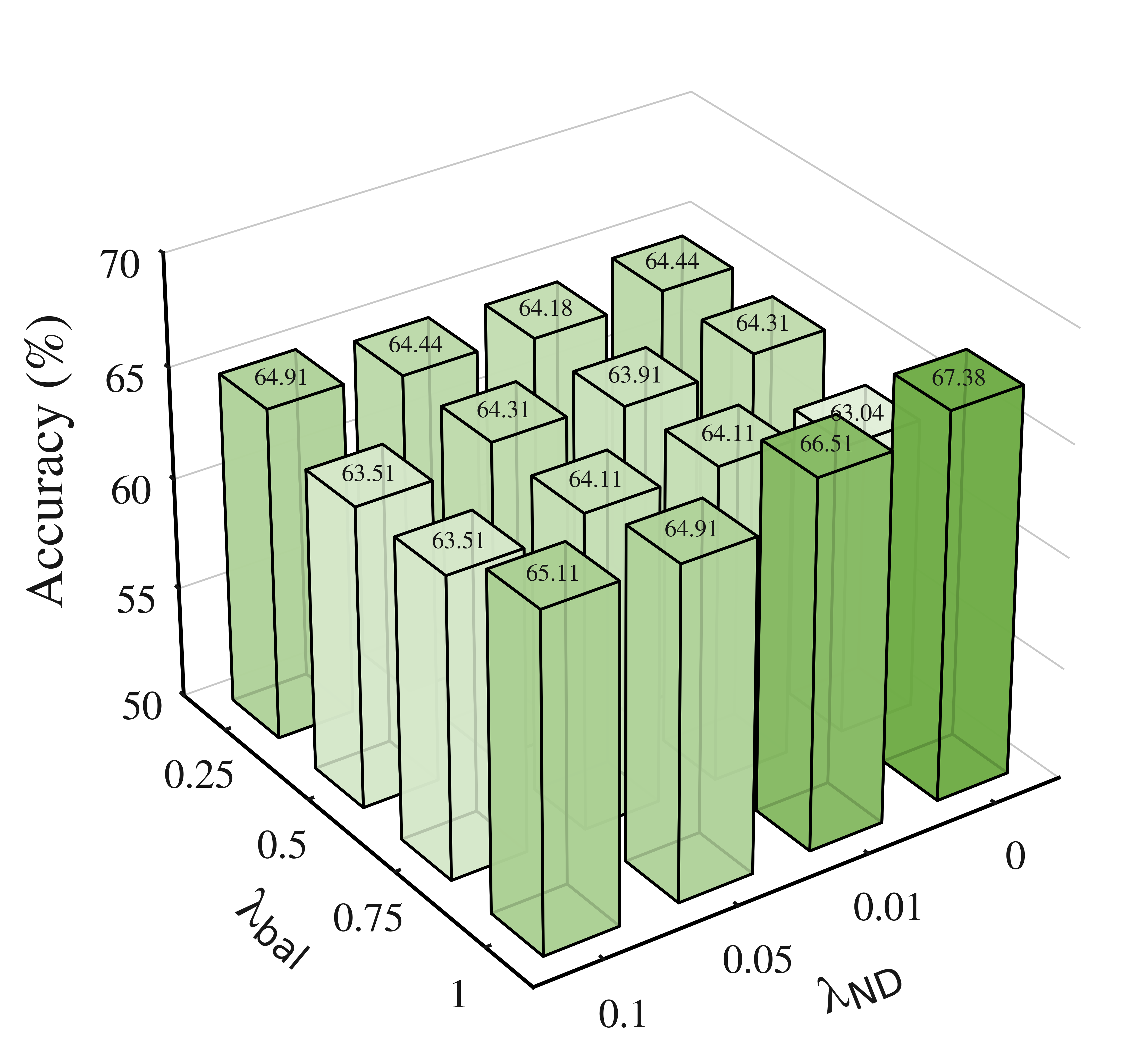} &
        \includegraphics[width=0.322\linewidth]{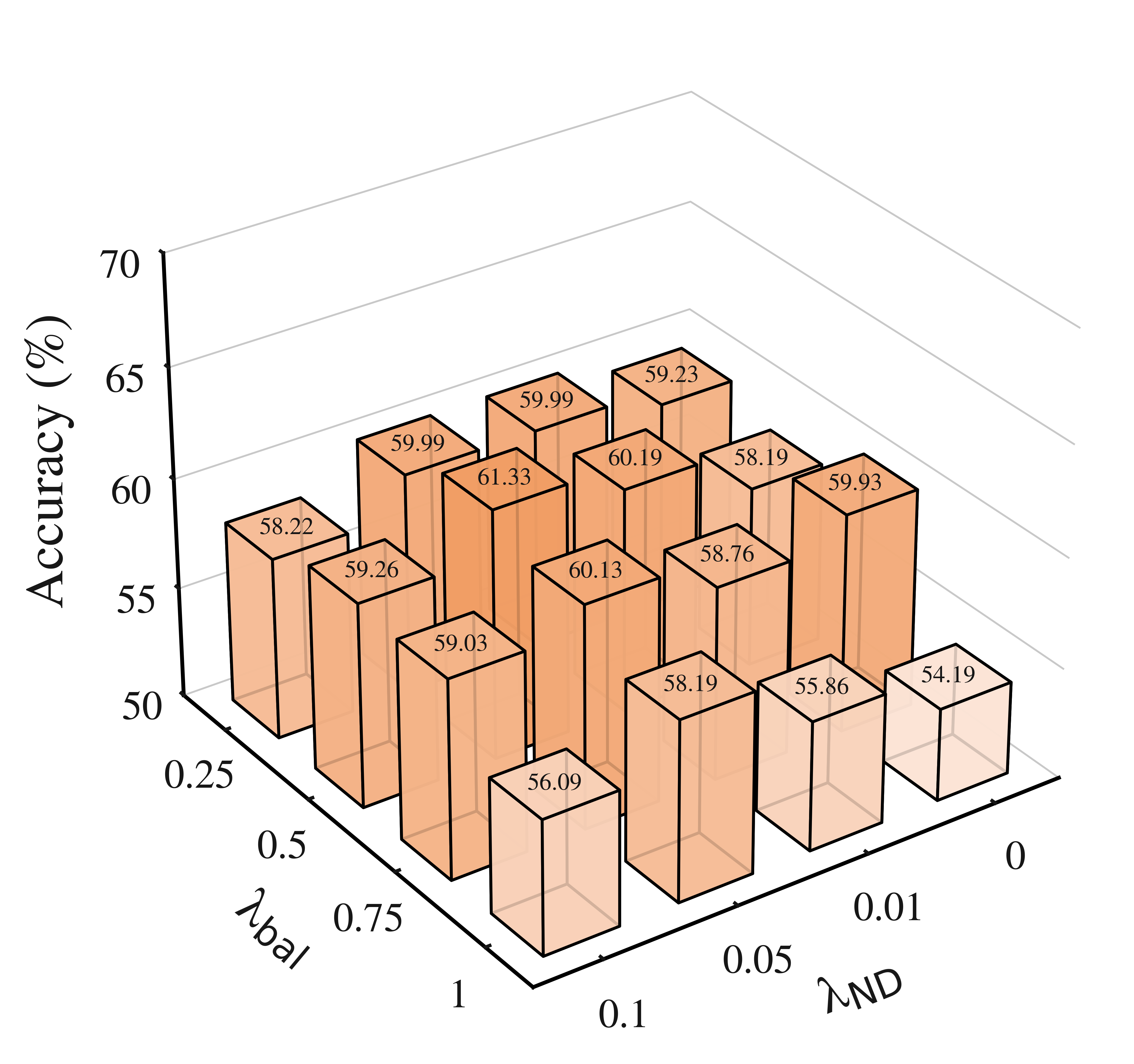} \\
        {\scriptsize (a) $\texttt{All}_{\mathrm{GH}}$} & {\scriptsize (b) $\texttt{Old}_{\mathrm{GH}}$} & {\scriptsize (c) $\texttt{New}_{\mathrm{GH}}$} \\
        \includegraphics[width=0.322\linewidth]{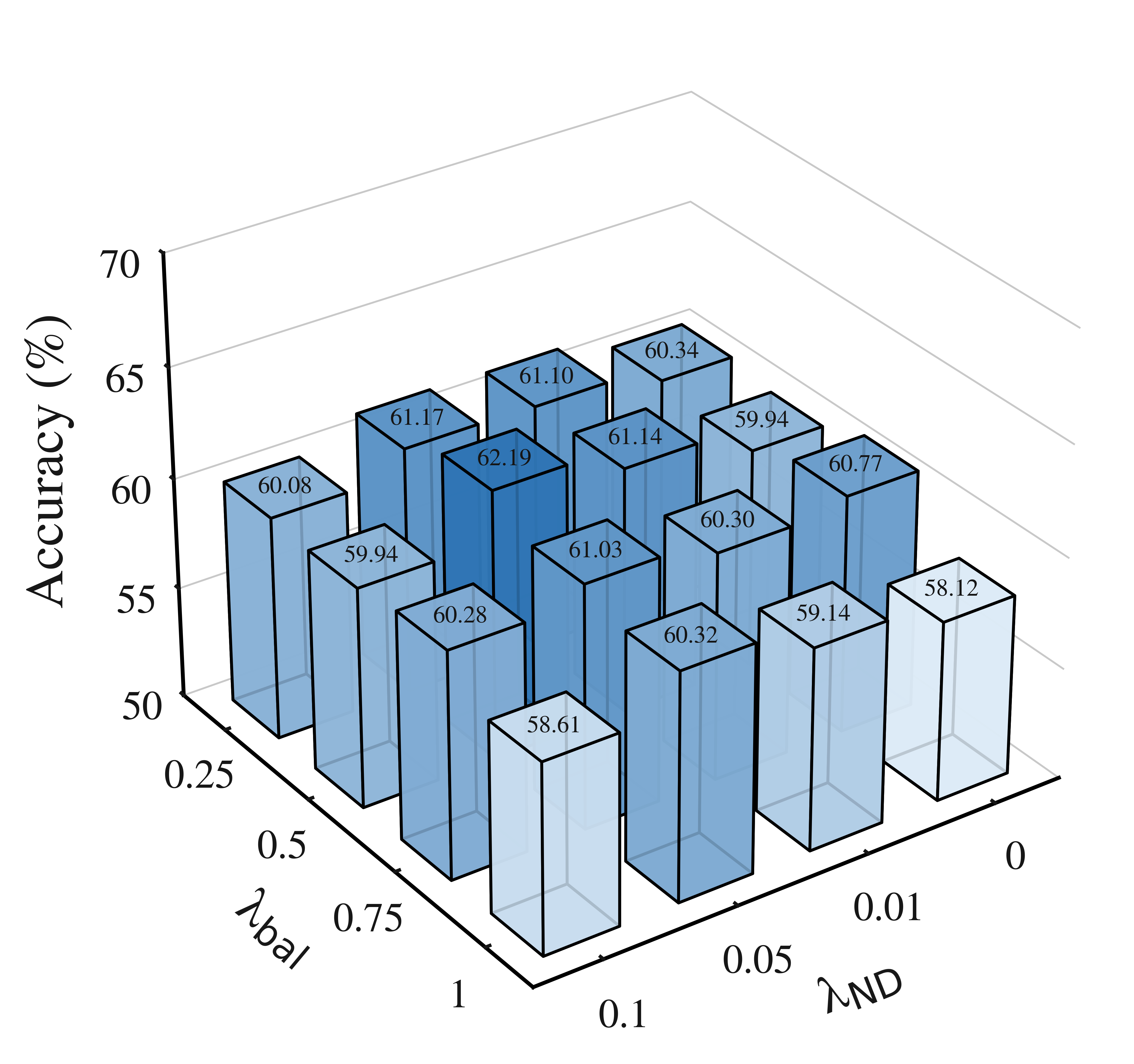} &
        \includegraphics[width=0.322\linewidth]{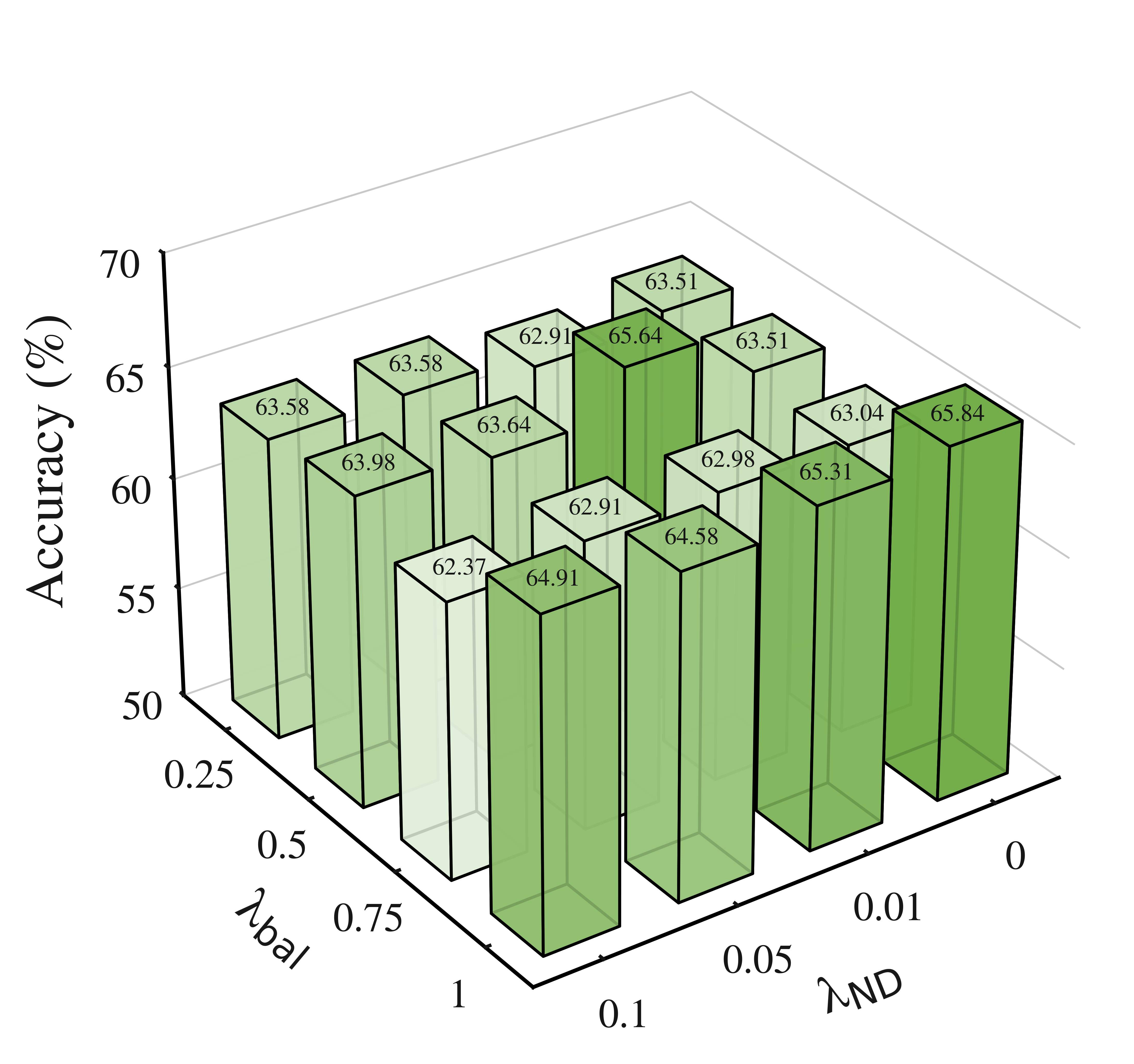} &
        \includegraphics[width=0.322\linewidth]{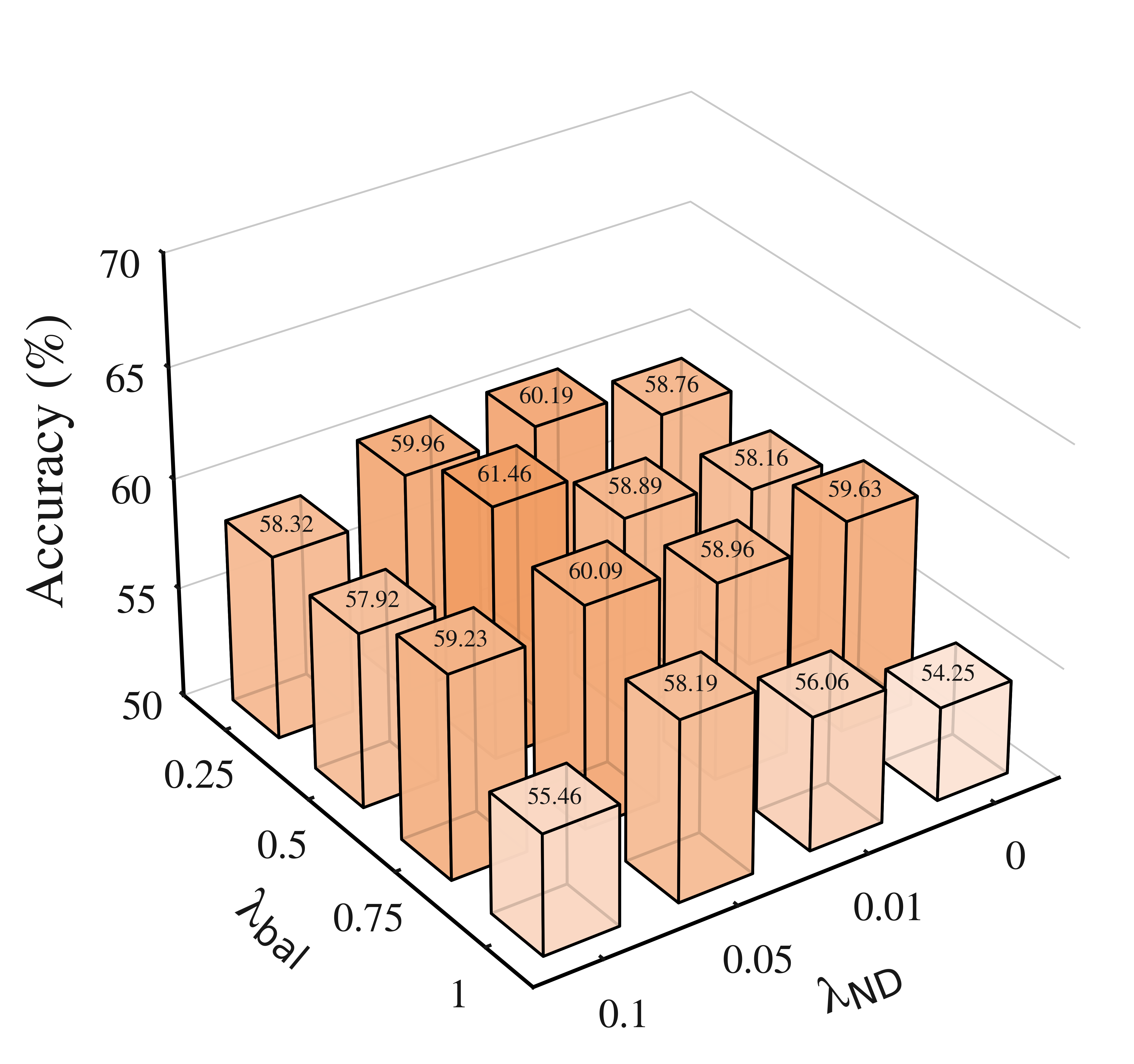} \\
        {\scriptsize (d) $\texttt{All}_{\mathrm{SPE}}$} & {\scriptsize (e) $\texttt{Old}_{\mathrm{SPE}}$} & {\scriptsize (f) $\texttt{New}_{\mathrm{SPE}}$}
    \end{tabular}
    \caption{Joint sensitivity of SimGCD--DINO on CUB to the balancing coefficient $\lambda_{\mathrm{RB}}$ and the NDL coefficient $\lambda_{\mathrm{ND}}$. The top and bottom rows report GH and SPE, respectively; the columns from left to right show All, Old, and New accuracy.}
    \label{fig:cub_hyperparameter_sensitivity}
\end{figure}

\textbf{Overall effectiveness under both protocols.}
Tab.\ref{tab:main_results} compares all seven baselines with their MMO variants on four fine-grained benchmarks. MMO improves new-category accuracy under both GH and SPE in $23$ of the $28$ configurations. For example, on CUB, MMO improves the new-category accuracy of SimGCD--CLIP by $5.57$ and $6.10$ points under GH and SPE, respectively. Overall, the average new-accuracy gain is $2.55$ points under GH and $6.48$ points under SPE. The larger gains under SPE support the alignment of MMO with our proposed evaluation protocol. Additional results on the coarse-grained ImageNet-100 benchmark are provided in App.\ref{app:coarse_grained_ours}. 

\textbf{Mitigating errors hidden by semantic mismapping.}
The largest SPE gains occur on Herbarium 19, where the overestimation is most significant. MMO raises $\texttt{New}_{\mathrm{SPE}}$ by $22.17$, $30.93$, and $26.76$ points for SimGCD--DINO, SimGCD--CLIP, and GET--CLIP, respectively. Their full-set gaps simultaneously decrease from $9.03$, $12.13$, and $10.44$ points to $0.85$, $0.56$, and $0.75$ points. Recall that the full-set GH--SPE gap is determined by the summed cycle margins through Eq.\eqref{eq:cycle_gap_identity}, as verified in Sec.\ref{sec:exp_protocol_importance}. This result is consistent with the theoretical bounds in Sec.\ref{sec:mmo_theory} and validates the effectiveness of MMO in reducing the semantic-mismapping errors.

\textbf{Better old--new trade-off of MMO.}
As shown in Fig.\ref{fig:old_new_tradeoff}, the $12$ CLIP-based configurations all improve new accuracy under both protocols, while old-category accuracy can decrease. For example, on CUB, SimGCD--CLIP gains $6.10$ points in $\texttt{New}_{\mathrm{SPE}}$ and $3.27$ points in $\texttt{H}_{\mathrm{SPE}}$, with a $0.34$-point decrease in $\texttt{Old}_{\mathrm{SPE}}$. This improvement shows that MMO helps achieve a better balance between old-category recognition and novel-category discovery. App.\ref{app:clip_tradeoffs} provides more CLIP-based trade-off plots.

\begin{table}[!t]
    \centering
    \scriptsize
    \caption{Large-scale comparison on ImageNet-1K. Adding MMO improves both new-category and overall accuracy under GH and SPE with DINO and CLIP backbones, demonstrating its effectiveness on the large-scale benchmark, as the answer to \textbf{RQ2}.}
    \label{tab:large_scale_results}
    \setlength{\tabcolsep}{1.5pt}
    \renewcommand{\arraystretch}{1.08}
    \resizebox{\columnwidth}{!}{%
    \begin{tabular}{@{}llrrrrrr@{}}
        \toprule
        Method & Backbone
        & $\texttt{All}_{\mathrm{GH}}$ & $\texttt{Old}_{\mathrm{GH}}$ & $\texttt{New}_{\mathrm{GH}}$
        & $\texttt{All}_{\mathrm{SPE}}$ & $\texttt{Old}_{\mathrm{SPE}}$ & $\texttt{New}_{\mathrm{SPE}}$ \\
        \midrule
        SimGCD & DINO & 57.04 & 77.14 & 46.94 & 54.81 & 85.15 & 39.58 \\
        \rowcolor{black!6}\quad\textbf{+ Ours} & DINO & 58.69 & 77.29 & 49.34 & 56.88 & 85.28 & 42.60 \\
        SimGCD & CLIP & 57.99 & 77.06 & 48.41 & 55.98 & 85.26 & 41.27 \\
        \rowcolor{black!6}\quad\textbf{+ Ours} & CLIP & 59.03 & 77.63 & 49.68 & 57.35 & 85.10 & 43.40 \\
        \bottomrule
    \end{tabular}}
\end{table}

\textbf{Large-scale evaluation.}
We further extend the comparison to ImageNet-1K in Tab.\ref{tab:large_scale_results}. Again, MMO improves both overall and new accuracy under GH and SPE with both backbones. For DINO, the overall/new gains are $1.65/2.40$ points under GH and $2.07/3.02$ points under SPE; for CLIP, they are $1.04/1.27$ and $1.37/2.13$ points, respectively. Complete results are provided in App.\ref{app:large_scale_details}.

\textbf{Sensitivity to the labeling fraction.}
On the imbalanced benchmark such as Herbarium 19, nearly equal numbers of old and new categories do not imply equal sample proportions~\cite{wang_imbalanced_2023,wang_localization_2026,yang_dirmixe_2024,li_focalsam_2025}. RBL therefore uses the labeling fraction $\rho$, or its estimate $\widehat\rho$, to infer the old-region sample prior through Eq.\eqref{eq:pao_region_prior}. Fig.\ref{fig:rho_sensitivity} compares manually specified estimates with the data-driven TIcE estimate for SimGCD--CLIP, while the true labeling fraction remains fixed at $0.50$.

The results show that increasing $\widehat\rho$ from $0.40$ to $0.70$ lowers the inferred old-region prior and shifts the accuracy trade-off toward new categories: $\texttt{Old}_{\mathrm{SPE}}$ decreases from $67.50\%$ to $37.89\%$, while $\texttt{New}_{\mathrm{SPE}}$ increases from $41.04\%$ to $45.37\%$. This trend agrees with RBL's role in controlling probability allocation across output regions. Among the manually specified settings, $\widehat\rho=0.40$ yields the highest $\texttt{All}_{\mathrm{SPE}}$ and $\texttt{H}_{\mathrm{SPE}}$, at $50.30\%$ and $51.04\%$. This suggests that slight perturbations to the regional prior may improve robustness and classification performance.

TIcE estimates $\widehat\rho=0.3575$ and $\widehat p_{\mathrm{old}}=0.6287$, yielding $\texttt{All}_{\mathrm{SPE}}=46.02\%$, $\texttt{New}_{\mathrm{SPE}}=32.72\%$, and $\texttt{H}_{\mathrm{SPE}}=44.74\%$. These scores exceed the baseline values of $35.27\%$, $12.63\%$, and $21.71\%$, respectively, although they fall below the best manually specified setting. In other words, MMO improves performance even when the regional prior is estimated automatically. 

\begin{figure*}[!b]
    \centering
    \includegraphics[width=\linewidth]{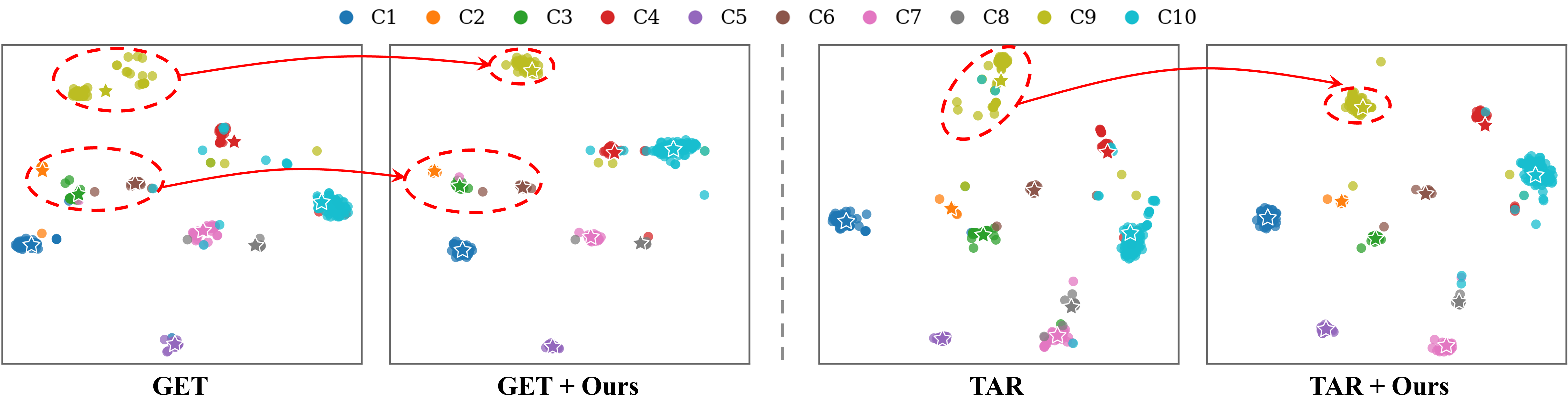}
    \caption{t-SNE visualization of ten selected new categories from Herbarium 19 for (a) GET, (b) GET\textbf{+Ours}, (c) GET-TAR, and (d) GET-TAR\textbf{+Ours}. Colored circles denote samples, and stars denote the corresponding class prototypes. Red dashed ellipses highlight representative local structures, while arrows connect the baseline and \textbf{+Ours} views for visual comparison. The complete visualizations and the definitions of the high-dimensional intra- and inter-class statistics are provided in App.\ref{app:tsne_additional}.}
    \label{fig:ndl_embedding}
\end{figure*}

\textbf{Hyperparameter sensitivity.}
Fig.\ref{fig:cub_hyperparameter_sensitivity} examines $\lambda_{\mathrm{RB}}$ and $\lambda_{\mathrm{ND}}$ jointly for SimGCD--DINO on CUB. Moderate RBL weights, $\lambda_{\mathrm{RB}}\in[0.25,0.75]$, preserve old-category accuracy, while moderate nonzero NDL weights, typically $\lambda_{\mathrm{ND}}\in[0.01,0.05]$, improve new-category recognition. These trends are consistent with the complementary roles of regional balance and new-prototype diversity. Jointly using the largest tested weights, $\lambda_{\mathrm{RB}}=1.0$ and $\lambda_{\mathrm{ND}}=0.1$, reduces $\texttt{All}_{\mathrm{GH}}/\texttt{All}_{\mathrm{SPE}}$ to $59.10/58.61$ and $\texttt{New}_{\mathrm{GH}}/\texttt{New}_{\mathrm{SPE}}$ to $56.09/55.46$, showing that stronger regularization does not necessarily improve performance. Overall, the moderate coefficient ranges provide a favorable balance between old-category recognition and novel-category discovery. Component ablation results for RBL and NDL are provided in App.\ref{app:mmo_component_ablation}.

\textbf{Visual analysis.}
Fig.\ref{fig:ndl_embedding} complements the aforementioned quantitative evaluation with t-SNE~\cite{van2008visualizing} views of selected new categories on Herbarium 19. Although GET and GET-TAR achieve only $17.10\%$ and $16.46\%$ new accuracy under SPE, samples from several new categories already form well-defined clusters. In other words, visually distinct clusters can coexist with low SPE accuracy. The large GH--SPE gaps further indicate that global reassignment can hide such prediction errors, validating the necessity of our SPE protocol.

With MMO, several displayed clusters become more compact and less overlapping, qualitatively supporting improved new-category discrimination. Meanwhile, $\texttt{New}_{\mathrm{SPE}}$ increases to $43.86\%$ for GET and $44.80\%$ for GET-TAR. The visualization illustrates changes in local feature structure, while the SPE results measure whether predictions respect the fixed old-output meanings and the exchangeability of new outputs.

\begin{table*}[!t]
    \centering
    \footnotesize
    \setlength{\abovecaptionskip}{3pt}
    \caption{Evaluation of SPE with an estimated new-category number following Vaze et al.~\cite{DBLP:conf/cvpr/VazeHVZ22}. SPE remains applicable when the estimated category number differs from the ground truth and reveals GH--SPE score differences on CUB and Stanford Cars, as the answer to \textbf{RQ3}.}
    \label{tab:estimated_cardinality_results}
    \setlength{\tabcolsep}{2.8pt}
    \renewcommand{\arraystretch}{1.25}
    \begin{tabularx}{\textwidth}{@{}lll>{\hsize=0.65\hsize\linewidth=\hsize\centering\arraybackslash}X>{\hsize=0.65\hsize\linewidth=\hsize\centering\arraybackslash}X>{\hsize=1.5\hsize\linewidth=\hsize\centering\arraybackslash}X>{\hsize=1.5\hsize\linewidth=\hsize\centering\arraybackslash}X>{\hsize=1\hsize\linewidth=\hsize\centering\arraybackslash}X>{\hsize=1\hsize\linewidth=\hsize\centering\arraybackslash}X>{\hsize=0.85\hsize\linewidth=\hsize\centering\arraybackslash}X>{\hsize=0.85\hsize\linewidth=\hsize\centering\arraybackslash}X@{}}
        \toprule
        Dataset & Method & Estimator
        & $C_{\mathrm{new}}$ & $\widehat{C}_{\mathrm{new}}$
        & $\texttt{New}_{\mathrm{GH}}$ & $\texttt{New}_{\mathrm{SPE}}$
        & $\texttt{Old}_{\mathrm{SPE}}$ & $\texttt{All}_{\mathrm{SPE}}$
        & $\Delta_{\mathrm{new}}$ & $\Delta_{\mathrm{all}}$ \\
        \midrule
        & SimGCD & DINO & 100 & 131 & 60.19 & 60.56 & 63.11 & 61.41 & \textbf{-0.37} & 0.27 \\
        \rowcolor{black!2}\cellcolor{white}& \quad\textbf{+ Ours} & DINO & 100 & 131 & \SPEgain{\textbf{60.46}}{+0.27} & \SPEgain{\textbf{60.99}}{+0.43} & \textbf{66.64} & \textbf{62.88} & -0.53 & \textbf{0.15} \\
        & SimGCD & CLIP & 100 & 112 & 67.60 & 67.23 & \textbf{76.32} & \textbf{70.26} & \textbf{0.37} & \textbf{0.13} \\
        \rowcolor{black!2}\cellcolor{white}\multirow{-4}{*}{CUB}& \quad\textbf{+ Ours} & CLIP & 100 & 112 & \SPEgain{\textbf{68.57}}{+0.97} & \SPEgain{\textbf{67.27}}{+0.04} & 74.92 & 69.82 & 1.30 & 0.71 \\
        \midrule
        & SimGCD & DINO & 98 & 132 & 44.53 & 42.60 & \textbf{72.91} & 52.47 & 1.93 & 0.41 \\
        \rowcolor{black!2}\cellcolor{white}& \quad\textbf{+ Ours} & DINO & 98 & 132 & \SPEgain{\textbf{46.68}}{+2.15} & \SPEgain{\textbf{45.84}}{+3.24} & 68.97 & \textbf{53.37} & \textbf{0.84} & \textbf{0.28} \\
        & SimGCD & CLIP & 98 & 136 & 69.30 & 65.39 & 87.26 & 72.51 & 3.91 & 1.06 \\
        \rowcolor{black!2}\cellcolor{white}\multirow{-4}{*}{Stanford Cars}& \quad\textbf{+ Ours} & CLIP & 98 & 136 & \SPEgain{\textbf{70.87}}{+1.57} & \SPEgain{\textbf{70.38}}{+4.99} & \textbf{88.31} & \textbf{76.22} & \textbf{0.48} & \textbf{0.05} \\
        \bottomrule
    \end{tabularx}
\end{table*}

\begin{table*}[!t]
    \centering
    \scriptsize
    \setlength{\abovecaptionskip}{3pt}
    \caption{Evaluation of the non-parametric SPE extension using SelEx~\cite{rastegar_selex_2024} equipped with Bures-Isotropy Alignment (BIA)~\cite{tang2026bia}. Labeled-set calibration enables SPE to evaluate non-parametric predictions and reveals GH--SPE score differences across the five benchmarks, as the answer to \textbf{RQ3}.}
    \label{tab:nonparametric_extension_results}
    \setlength{\tabcolsep}{3.0pt}
    \renewcommand{\arraystretch}{1.25}
    \definecolor{ProtocolGapPositive}{RGB}{232,151,76}
    \definecolor{ProtocolGapNegative}{RGB}{92,160,211}
    \newcommand{\ProtocolGap}[1]{%
        \ifdim#1pt<0pt
            \def\ProtocolGapFill{ProtocolGapNegative}%
            \dimen0=-#1pt\relax
        \else
            \def\ProtocolGapFill{ProtocolGapPositive}%
            \dimen0=#1pt\relax
        \fi
        \ifdim\dimen0<1pt \def\ProtocolGapShade{10}%
        \else\ifdim\dimen0<3pt \def\ProtocolGapShade{23}%
        \else\ifdim\dimen0<7pt \def\ProtocolGapShade{38}%
        \else\ifdim\dimen0<15pt \def\ProtocolGapShade{55}%
        \else \def\ProtocolGapShade{75}\fi\fi\fi\fi
        \ifdim#1pt>0pt\ifdim\dimen0<15pt\else\def\ProtocolGapShade{60}\fi\fi
        \ifdim\dimen0=0pt
            \def\ProtocolGapFill{ProtocolGapNegative}\def\ProtocolGapShade{5}%
        \fi
        \edef\ProtocolGapPaint{\noexpand\cellcolor{\ProtocolGapFill!\ProtocolGapShade!white}}%
        \ProtocolGapPaint\textcolor{black}{#1}%
    }
    \begin{tabular*}{\textwidth}{@{\extracolsep{\fill}}lrrrrrrrrrr@{}}
        \toprule
        \multirow{2}{*}{Dataset} & \multirow{2}{*}{Cali. Acc.}
        & \multicolumn{3}{c}{Old} & \multicolumn{3}{c}{New} & \multicolumn{3}{c}{All} \\
        \cmidrule(lr){3-5}\cmidrule(lr){6-8}\cmidrule(l){9-11}
        &
        & $\texttt{Old}_{\mathrm{GH}}$ & $\texttt{Old}_{\mathrm{SPE}}$ & $\Delta_{\mathrm{old}}$
        & $\texttt{New}_{\mathrm{GH}}$ & $\texttt{New}_{\mathrm{SPE}}$ & $\Delta_{\mathrm{new}}$
        & $\texttt{All}_{\mathrm{GH}}$ & $\texttt{All}_{\mathrm{SPE}}$ & $\Delta_{\mathrm{all}}$ \\
        \midrule
        CIFAR-100 & 94.77 & 81.04 & 82.15 & \ProtocolGap{-1.11} & 68.20 & 65.45 & \ProtocolGap{2.75} & 76.76 & 76.58 & \ProtocolGap{0.18} \\
        CUB & 99.93 & 67.71 & 67.58 & \ProtocolGap{0.13} & 74.04 & 73.74 & \ProtocolGap{0.30} & 71.93 & 71.69 & \ProtocolGap{0.24} \\
        Stanford Cars & 99.75 & 75.21 & 77.26 & \ProtocolGap{-2.05} & 50.88 & 48.73 & \ProtocolGap{2.15} & 58.81 & 58.02 & \ProtocolGap{0.78} \\
        FGVC-Aircraft & 99.70 & 65.01 & 65.97 & \ProtocolGap{-0.96} & 57.00 & 56.34 & \ProtocolGap{0.66} & 59.67 & 59.55 & \ProtocolGap{0.12} \\
        Herbarium 19 & 82.93 & 45.08 & 45.22 & \ProtocolGap{-0.14} & 30.10 & 28.17 & \ProtocolGap{1.93} & 35.34 & 34.13 & \ProtocolGap{1.21} \\
        \bottomrule
    \end{tabular*}
\end{table*}

\subsection{Applicability of SPE Extensions (RQ3)}
\label{sec:exp_spe_extensions}
In this part, we examine the two SPE extensions in Sec.\ref{sec:nonparametric_extension}: estimating the number of new categories and evaluating non-parametric GCD methods. Both extensions preserve the divide-and-conquer principle of SPE by fixing the old-output correspondence and restricting category matching to the new region.

\textbf{Estimated number of new categories.}
Following Vaze et al.~\cite{DBLP:conf/cvpr/VazeHVZ22}, we estimate the total category number $\widehat C$ and obtain $\widehat C_{\mathrm{new}}=\widehat C-C_{\mathrm{old}}$. SPE accommodates the resulting mismatch by padding only the new region, while preserving the anchored old-output semantics. In Tab.\ref{tab:estimated_cardinality_results}, DINO and CLIP estimate $131$ and $112$ new categories on CUB, whose true number is $100$, and $132$ and $136$ on Stanford Cars, whose true number is $98$. SPE remains applicable in all these configurations and exposes score gains from global reassignment. For example, the Stanford Cars SimGCD--CLIP baseline has $\Delta_{\mathrm{new}}=3.91$ and $\Delta_{\mathrm{all}}=1.06$. With MMO, these gaps decrease to $0.48$ and $0.05$ points, respectively, and $\texttt{All}_{\mathrm{SPE}}$ increases from $72.51\%$ to $76.22\%$. These results show that the evaluation extension supports both baseline and MMO when the new-category number is estimated. Complete results are provided in App. Tab.\ref{tab:appendix_estimated_cardinality_full}.

\textbf{Non-parametric GCD.}
Non-parametric cluster indices are initially exchangeable. We therefore establish the old-output correspondence using only the labeled set $\mathcal D_l$ and freeze this correspondence before evaluating $\mathcal D_u$, as prescribed by Eq.\eqref{eq:nonparametric_calibration}. Tab.\ref{tab:nonparametric_extension_results} evaluates this extension with SelEx~\cite{rastegar_selex_2024} equipped with BIA~\cite{tang2026bia}. We report calibration accuracy to assess how well the fixed old-output correspondence matches the labeled samples, since SPE relies on this correspondence to anchor old-category semantics. With $N_l=|\mathcal D_l|$ and the mapping $b_l^*$ defined in Eq.\eqref{eq:nonparametric_calibration},
\begin{equation}
    \mathrm{Cali.\,Acc.}
    = \frac{1}{N_l}\sum_{(\boldsymbol{x},y)\in\mathcal D_l}
    \mathbb{I}\{f(\boldsymbol{x})=b_l^*(y)\}.
    \label{eq:calibration_accuracy}
\end{equation}
This measures the fraction of labeled samples correctly assigned under the calibrated mapping. No unlabeled ground-truth categories are used to establish this mapping.

GH--SPE differences remain observable after calibration. On CUB, calibration accuracy is $99.93\%$, yet GH exceeds SPE by $0.30$ points in new accuracy and $0.24$ points overall. The new-accuracy gaps reach $2.75$ points on CIFAR-100 and $2.15$ points on Stanford Cars. On Herbarium 19, calibration accuracy is lower at $82.93\%$, and the overall gap reaches $1.21$ points. These results show that semantic mismapping also occurs in non-parametric GCD after labeled-set calibration. Across the shared benchmarks, the non-parametric method exhibits smaller GH--SPE gaps on average than the parametric methods.

Together, these experiments answer \textbf{RQ3}: SPE extends to estimated category numbers through new-region padding and to non-parametric predictions through labeled-set calibration, which supplies the fixed old-output semantics required by the protocol.

\section{Conclusion}
\label{sec:conclusion}
In this paper, we revisit the evaluation of parametric generalized category discovery and show that global matching can overestimate performance by overriding the semantic meaning of supervised old outputs. Our theoretical analysis and empirical findings pinpoint semantic mismapping as a source of misleading accuracy gains that can also affect method rankings. To address this issue, we introduce SPE to preserve anchored output semantics during evaluation, and develop MMO from this evaluation perspective to mitigate prediction errors that global matching can hide. Experiments demonstrate the applicability of SPE across different GCD settings and show that MMO generally improves new-category recognition and the balance between old and new categories. These findings support using SPE alongside GH to provide a more informative assessment of GCD models while maintaining comparability with previous work. More broadly, our study highlights the importance of respecting learned output semantics in evaluation and exploiting evaluation insights to guide model optimization.

  \bibliographystyle{IEEEtranN}
  \bibliography{IEEEabrv,citations}

  \begingroup
\makeatletter
\def\@IEEEBIOskipN{3\baselineskip}
\makeatother

\begin{IEEEbiography}
    [{\includegraphics[width=1in,height=1.25in,clip,keepaspectratio]{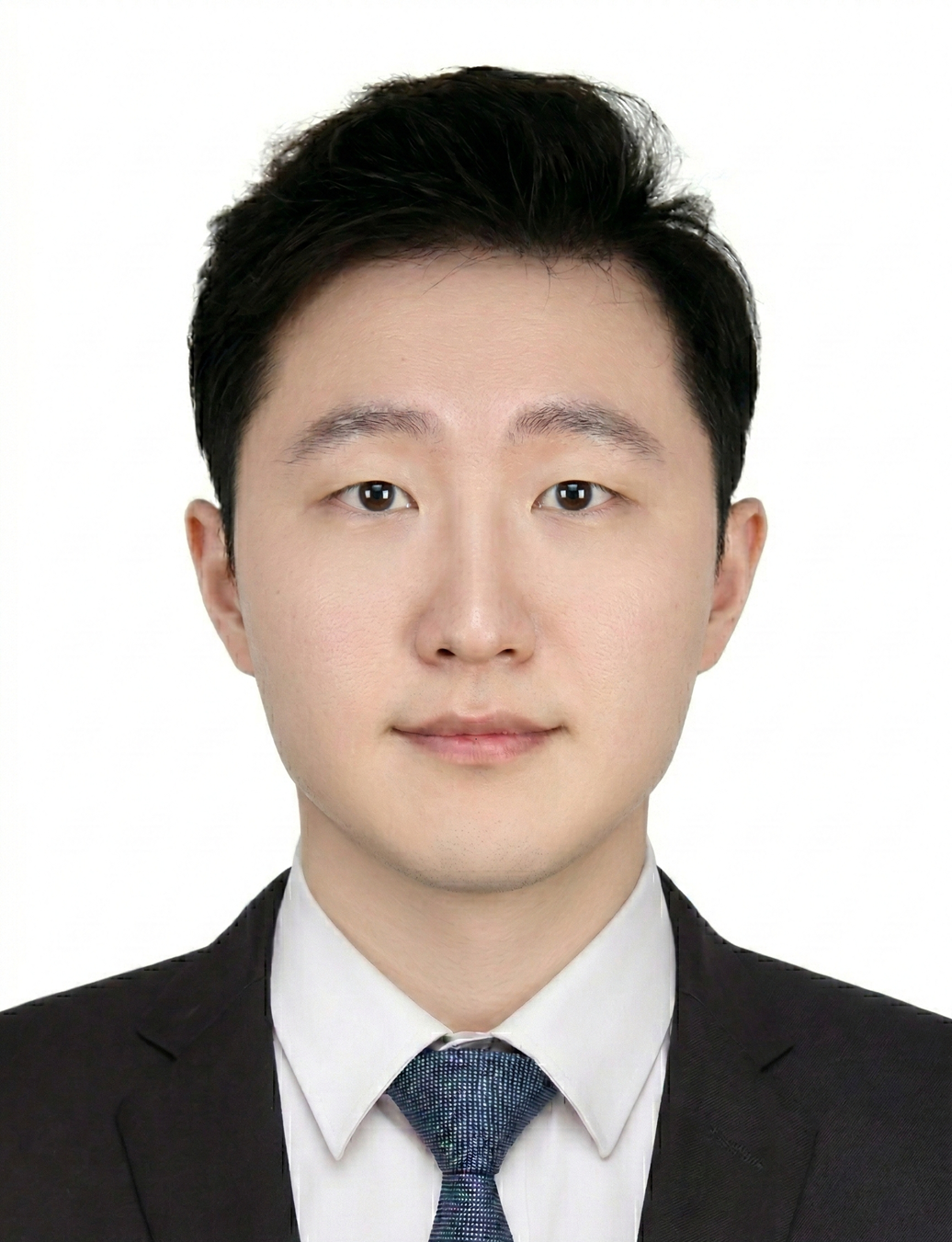}}]{Zitai Wang} received the B.E. degree in computer science and technology from Beijing Jiaotong University in 2019 and the Ph.D. degree from the University of Chinese Academy of Sciences in 2024. He is currently a post-doctoral research fellow with the Institute of Computing Technology, Chinese Academy of Sciences. His research interests include machine learning and data mining. He has authored or coauthored 20+ academic papers in top-tier international conferences and journals including IEEE T-PAMI, IJCV, ICML, NeurIPS, CVPR, AAAI, and ACM Multimedia. 
\end{IEEEbiography}

\begin{IEEEbiography}
	[{\includegraphics[width=1in,height=1.25in,clip,keepaspectratio]{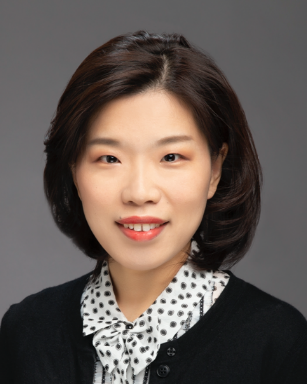}}]{Qianqian Xu} received the B.S. degree in computer science from China University of Mining and Technology in 2007 and the Ph.D. degree in computer science from the University of Chinese Academy of Sciences in 2013. She is currently a Professor with the Institute of Computing Technology, Chinese Academy of Sciences, Beijing, China. Her research interests include statistical machine learning, with applications in multimedia and computer vision. She has authored or coauthored 100+ academic papers in prestigious international journals and conferences (including T-PAMI, IJCV, T-IP, NeurIPS, ICML, CVPR, AAAI, etc.). Moreover, she serves as an associate editor of IEEE Transactions on Circuits and Systems for Video Technology, IEEE Transactions on Multimedia, and ACM Transactions on Multimedia Computing, Communications, and Applications.
\end{IEEEbiography}

\begin{IEEEbiography}
	[{\includegraphics[width=1in,height=1.25in,clip,keepaspectratio]{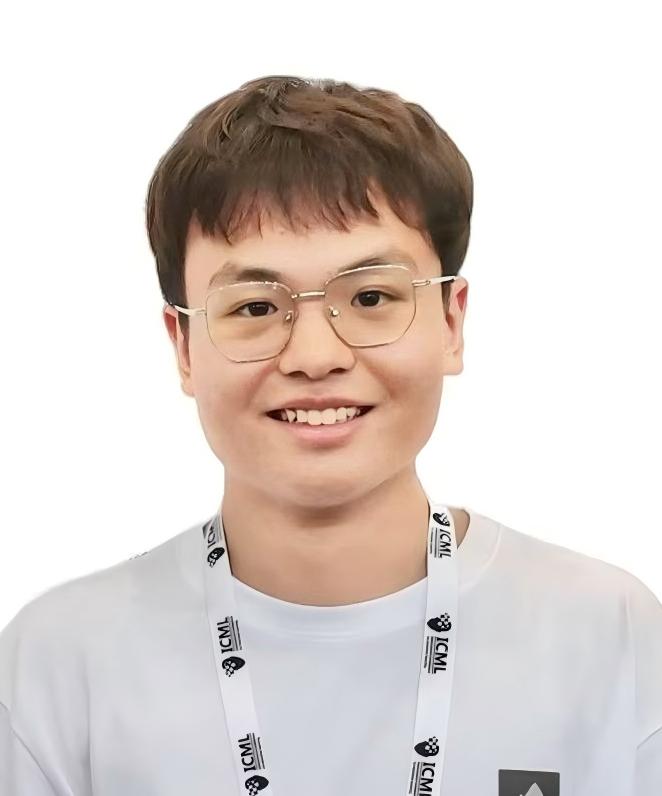}}]{Cong Hua} received the B.S. degree from Jilin University in 2023. He is currently working toward the Ph.D. degree with the Institute of Computing Technology, Chinese Academy of Sciences. His research interests are machine learning and multi-modal learning. He has authored or coauthored several academic papers in top-tier international conferences and journals including ICML, AAAI, and TNNLS. 
\end{IEEEbiography}

\begin{IEEEbiography}
	[{\includegraphics[width=1in,height=1.25in,clip,keepaspectratio]{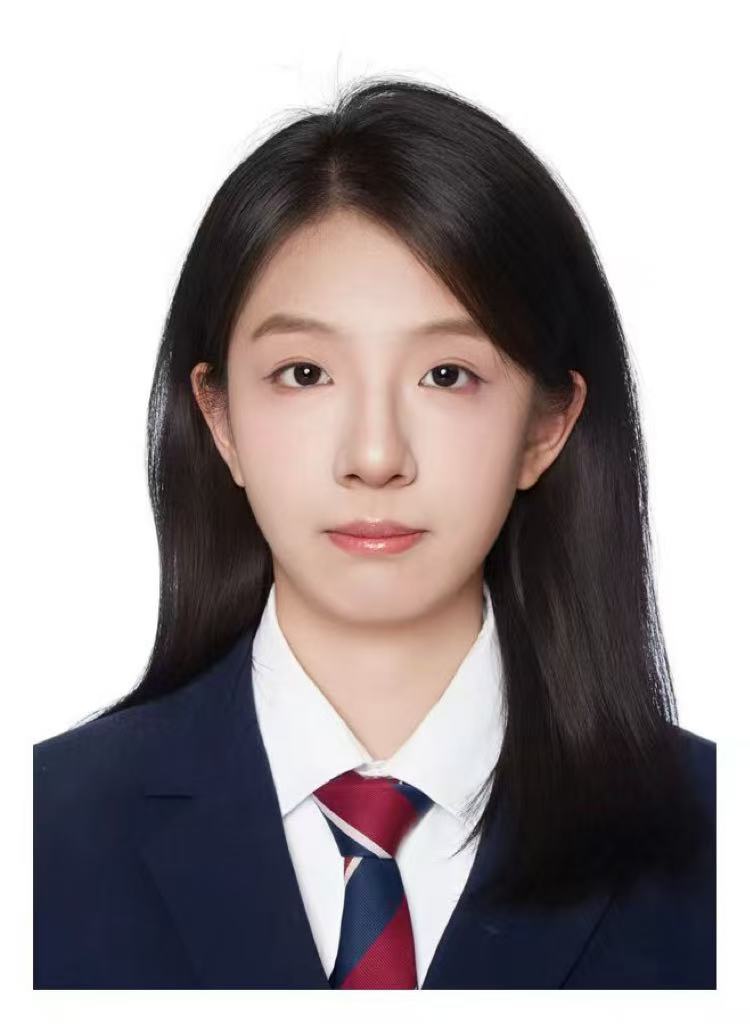}}]{Xi Chen} received the B.Eng. degree from Tsinghua University in 2026. She is currently pursuing her Ph.D. degree with the Institute of Computing Technology, Chinese Academy of Sciences. Her research interest is machine learning.
\end{IEEEbiography}

\begin{IEEEbiography}
	[{\includegraphics[width=1in,height=1.25in,clip,keepaspectratio]{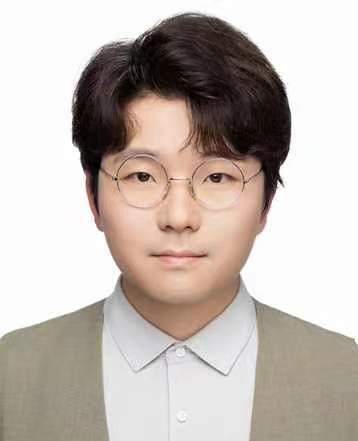}}]{Zhiyong Yang} received his M.S. degree in computer science and technology from the University of Science and Technology Beijing (USTB) in 2017 and his Ph.D. degree from the University of Chinese Academy of Sciences (UCAS) in 2021. He is currently an Associate Professor at the UCAS. His research interests include trustworthy machine learning, long-tail learning, and optimization frameworks for complex metrics. His work has been recognized with various awards, including Top 100 Baidu AI Chinese Rising Stars Around the World, Top-20 Nomination for the Baidu Fellowship, Asian Trustworthy Machine Learning (ATML) Fellowship, and the China Computer Federation (CCF) Doctoral Dissertation Award. He has authored or co-authored over 60 papers in top-tier international conferences and journals, including 30+ papers in T-PAMI, ICML, and NeurIPS. He has also served as an AC for NeurIPS/ICLR/ICML, an SPC member for IJCAI.
\end{IEEEbiography}

\begin{IEEEbiography}
	[{\includegraphics[width=1in,height=1.25in,clip,keepaspectratio]{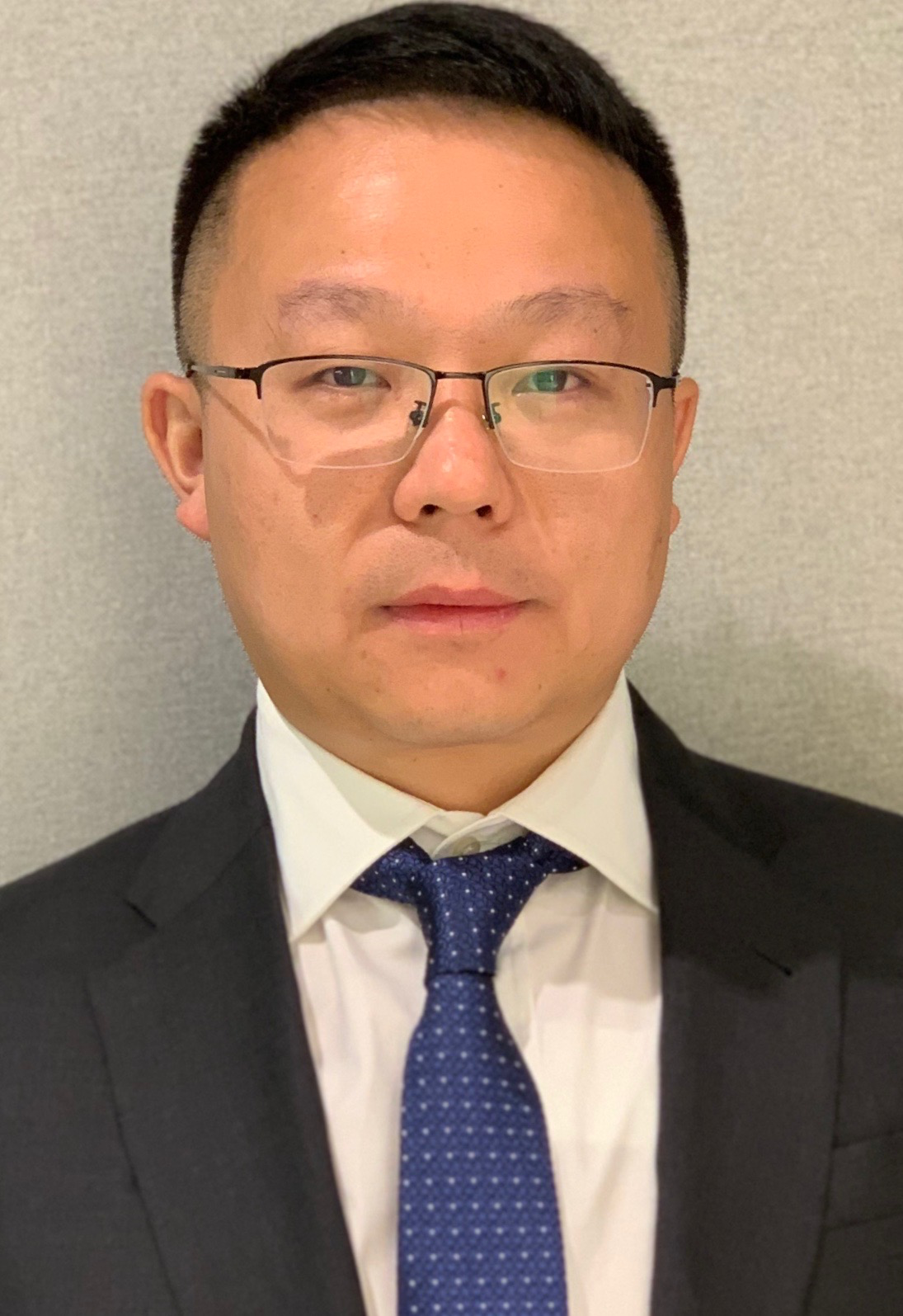}}]{Xiaochun Cao} is a Professor in the School of Cyber Science and Technology, Shenzhen Campus of Sun Yat-sen University. He received the B.E. and M.E. degrees both in computer science from Beihang University (BUAA), China, and the Ph.D. degree in computer science from the University of Central Florida, USA, with his dissertation nominated for the university-level Outstanding Dissertation Award. After graduation, he spent about three years at ObjectVideo Inc. as a Research Scientist. From 2008 to 2012, he was a professor at Tianjin University. Before joining SYSU, he was a professor at the Institute of Information Engineering, Chinese Academy of Sciences. He has authored and coauthored over 200 journal and conference papers. In 2004 and 2010, he was the recipient of the Piero Zamperoni best student paper award at the International Conference on Pattern Recognition. He is on the editorial boards of IEEE Transactions on Image Processing and IEEE Transactions on Multimedia, and was on the editorial board of IEEE Transactions on Circuits and Systems for Video Technology.
\end{IEEEbiography}

\begin{IEEEbiography}
	[{\includegraphics[width=1in,height=1.25in,clip,keepaspectratio]{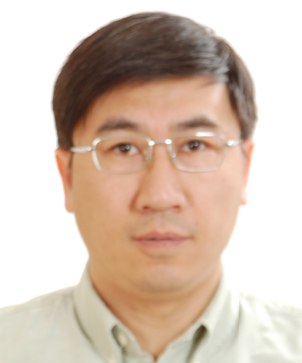}}]{Qingming Huang} is a chair professor in the University of Chinese Academy of Sciences and an adjunct research professor in the Institute of Computing Technology, Chinese Academy of Sciences. He graduated with a bachelor's degree in Computer Science in 1988 and a Ph.D. degree in Computer Engineering in 1994, both from Harbin Institute of Technology, China. His research areas include multimedia computing, image processing, computer vision and pattern recognition. He has authored or coauthored more than 400 academic papers in prestigious international journals and top-level international conferences. He was the associate editor of IEEE Trans. on CSVT and Acta Automatica Sinica, and the reviewer of various international journals including IEEE Trans. on PAMI, IEEE Trans. on Image Processing, IEEE Trans. on Multimedia, etc. He is a Fellow of IEEE and has served as general chair, program chair, area chair and TPC member for various conferences, including ACM Multimedia, CVPR, ICCV, ICME, ICMR, PCM, BigMM, PSIVT, etc.
\end{IEEEbiography}

\vfill

\endgroup

\clearpage
\onecolumn
\raggedbottom
\appendices
\startcontents[appendix]
\vspace*{12pt}
\begin{center}
    {\Large\bfseries Appendix Contents}
\end{center}
\medskip
\printcontents[appendix]{}{1.3}{\setcounter{tocdepth}{2}}
\clearpage

\section{Proofs of the Theoretical Results}
\label{app:proof_mismapping_margins}

\subsection{Semantic Mismapping Margin}

\begin{theoremstatementbox}
\textbf{Thm.\ref{thm:semantic_mismapping_margin} (Semantic mismapping margin).}
\begingroup
\itshape
Let $C_{y\rightarrow\hat{y}}$ be the number of samples from true category $y$ predicted as output index $\hat{y}\in\widehat{\mathcal{Y}}$. We define the highest-scoring assignment that preserves the supervised old-output meanings by
\begin{equation*}
\begin{aligned}
\mathcal{A}_0
&\in
\arg\max_{\mathcal{A}\in\Pi}
\sum_{\hat{y}\in\widehat{\mathcal{Y}}}
C_{\mathcal{A}(\hat{y})\rightarrow\hat{y}} \\
\textnormal{s.t.}\quad
&\mathcal{A}(\hat{y}_o)=y_o,
\quad \forall y_o\in\mathcal{Y}_{\mathrm{old}},\\
&\mathcal{A}(\widehat{\mathcal{Y}}_{\mathrm{new}})
=\mathcal{Y}_{\mathrm{new}}.
\end{aligned}
\end{equation*}
For different categories $y_1,\ldots,y_k$ with $k\geq2$, set $\hat{y}_i=\mathcal{A}_0^{-1}(y_i)$, $y_{k+1}=y_1$, and $\mathcal{Y}_{\gamma}:=\{y_1,\ldots,y_k\}$. The permutation cycle $\gamma=(y_1,\ldots,y_k)$ defines an assignment $\mathcal{A}_{\gamma}$ that maps $\hat{y}_i$ to $y_{i+1}$ and agrees with $\mathcal{A}_0$ elsewhere. Define
\begin{equation}
    m_{\gamma}
    :=
    \sum_{i=1}^{k}
    \left(
    C_{y_i\rightarrow\hat{y}_i}
    -C_{y_{i+1}\rightarrow\hat{y}_i}
    \right).
\end{equation}
The assignment $\mathcal{A}_{\gamma}$ improves the global matching score if and only if $m_{\gamma}<0$, and ties when $m_{\gamma}=0$. Define
\begin{equation}
    m_{\mathrm{sem}}
    :=
    \min_{\gamma:\,\mathcal{Y}_{\gamma}\cap\mathcal{Y}_{\mathrm{old}}\neq\varnothing}
    m_{\gamma}.
\end{equation}
Then $\mathcal{A}_0$ is globally optimal over the entire category space if and only if $m_{\mathrm{sem}}\geq0$.
\endgroup
\end{theoremstatementbox}

\begin{proof}
We first examine how a single cycle changes the matching score, then extend the argument to an arbitrary assignment. Let
\begin{equation}
    \mathcal{S}(\mathcal{A})
    :=
    \sum_{\hat y\in\widehat{\mathcal Y}}
    C_{\mathcal A(\hat y)\rightarrow\hat y}
\end{equation}
be the number of correct predictions under assignment $\mathcal A$. Along a cycle $\gamma=(y_1,\ldots,y_k)$, $\mathcal A_\gamma$ assigns output $\hat y_i$ to $y_{i+1}$ instead of $y_i$. All other outputs keep their assignments, so their contributions cancel when we subtract the two scores:
\begin{equation}
    \mathcal{S}(\mathcal{A}_{\gamma})
    -\mathcal{S}(\mathcal{A}_0)
    =
    \sum_{i=1}^{k}
    \left(
    C_{y_{i+1}\rightarrow\hat{y}_i}
    -C_{y_i\rightarrow\hat{y}_i}
    \right)
    =-m_{\gamma}.
\end{equation}
Thus, a negative margin means that the cycle gains more correct predictions than it loses. The score increases exactly when $m_\gamma<0$ and remains unchanged when $m_\gamma=0$.

To compare $\mathcal A_0$ with any global assignment, let $\mathcal A\in\Pi$, where $\Pi$ is the set of all bijections from output indices $\widehat{\mathcal Y}$ to categories $\mathcal Y$. Their relative reassignment is described by the category permutation
\begin{equation}
    \pi_{\mathcal{A}}
    :=
    \mathcal{A}\circ\mathcal{A}_0^{-1}.
\end{equation}
For each category $y$, this permutation gives the category that $\mathcal A$ assigns to the output previously assigned to $y$ by $\mathcal A_0$. Decompose it into disjoint nontrivial cycles $\gamma_1,\ldots,\gamma_r$. Since these cycles affect disjoint outputs and fixed points leave assignments unchanged, the total score change is
\begin{equation}
    \mathcal{S}(\mathcal{A})
    -\mathcal{S}(\mathcal{A}_0)
    =
    -\sum_{j=1}^{r}m_{\gamma_j}.
\end{equation}
A cycle involving only new categories leaves all old-output assignments unchanged. Since $\mathcal A_0$ already maximizes the score with these assignments fixed, such a cycle cannot increase the score and thus has a nonnegative margin. Hence, any improvement must come from a cycle involving an old category, and $m_{\mathrm{sem}}$ is exactly the minimum margin over these cycles. If $m_{\mathrm{sem}}\geq0$, all cycle margins are nonnegative, so no assignment improves on $\mathcal A_0$; otherwise, a negative-margin cycle yields a higher score. Hence, $\mathcal A_0$ is globally optimal if and only if $m_{\mathrm{sem}}\geq0$.
\end{proof}

\begingroup

\clearpage
\subsection{Theoretical Analysis of MMO}
\label{app:mmo_bound}
\begin{assumption}
\label{ass:mmo_analysis}
We have the following assumptions.
\par
\begin{itemize}[leftmargin=*,topsep=3pt,itemsep=3pt,parsep=0pt]
\item \textbf{Evaluation setting.} We consider a fixed model on an unlabeled set $\mathcal D_u$ of size $N_u$ containing at least one sample from every category. The output and category sets have equal cardinality, with at least one old category and $C_{\mathrm{new}}\geq2$ new categories.

\item \textbf{Prediction--similarity consistency.} Whenever the prediction $\hat y_i\in\arg\max_{j\in\widehat{\mathcal Y}}s_{i,j}$ lies in the new region, its prototype also maximizes cosine similarity among new outputs:
\begin{equation}
    \hat y_i\in\widehat{\mathcal Y}_{\mathrm{new}}
    \quad\Longrightarrow\quad
    \boldsymbol z_i^{\top}\widetilde{\mathbf w}_{\hat y_i}
    \geq\boldsymbol z_i^{\top}\widetilde{\mathbf w}_u,
    \quad\forall u\in\widehat{\mathcal Y}_{\mathrm{new}},
    \label{eq:mmo_cosine_head}
\end{equation}
where $\boldsymbol z_i$ and $\widetilde{\mathbf w}_u$ denote unit-norm features and new-output prototypes, respectively.
\end{itemize}
\end{assumption}

\begin{remark}
These assumptions are mild in the common GCD setting.
\par
\begin{itemize}[leftmargin=*,topsep=3pt,itemsep=3pt,parsep=0pt]
\item For the evaluation setting, the conditions hold as long as the full evaluation set covers every category and the head uses the given category count, with at least two new categories.

\item For prediction--similarity consistency, the condition is mild for normalized cosine heads because it only requires consistency with similarity ordering, not correct predictions or well-separated prototypes. The cosine classifiers used in SimGCD~\cite{DBLP:conf/iccv/WenZQ23}, LegoGCD~\cite{DBLP:conf/cvpr/CaoZWYS0L024}, and ProtoGCD~\cite{ma2025protogcd} satisfy this condition: their logits are inner products between normalized features and prototypes, so the predicted output naturally maximizes cosine similarity. More generally, the condition holds for any shared strictly increasing transformation of cosine similarity. For example, logits $s_{i,u}=a_i\boldsymbol z_i^{\top}\widetilde{\mathbf w}_u+b_i$, with $a_i>0$ and $a_i,b_i$ shared across new outputs, satisfy Eq.\eqref{eq:mmo_cosine_head}, and a shared positive temperature $\tau$ corresponds to $a_i=1/\tau$ and $b_i=0$.
\end{itemize}
\end{remark}

\begin{lemma}[Prediction and prototype distance]
\label{lem:mmo_prediction_distance}
Under Assump.\ref{ass:mmo_analysis}, suppose sample $i$ is predicted as a new output $v$. For any other new output $u\neq v$, let $c_{uv}:=\widetilde{\mathbf w}_u^{\top}\widetilde{\mathbf w}_v$. Then,
\begin{equation}
    1-c_{uv}\leq2\|\boldsymbol z_i-\widetilde{\mathbf w}_u\|_2^2.
    \label{eq:mmo_prediction_distance}
\end{equation}
\end{lemma}

\begin{proof}
Prediction--similarity consistency gives
\begin{equation}
    \begin{aligned}
        0
        &\geq\boldsymbol{z}_i^{\top}
        (\widetilde{\mathbf{w}}_u-\widetilde{\mathbf{w}}_v)\\
        &=1-c_{uv}+(\boldsymbol{z}_i-\widetilde{\mathbf{w}}_u)^{\top}
        (\widetilde{\mathbf{w}}_u-\widetilde{\mathbf{w}}_v)\\
        &\geq1-c_{uv}
        -\|\boldsymbol{z}_i-\widetilde{\mathbf{w}}_u\|_2
        \sqrt{2(1-c_{uv})}.
    \end{aligned}
\end{equation}
The first inequality follows from Assump.\ref{ass:mmo_analysis}: predicting $v$ implies $\boldsymbol{z}_i^{\top}\widetilde{\mathbf{w}}_v\geq\boldsymbol{z}_i^{\top}\widetilde{\mathbf{w}}_u$. The equality adds and subtracts $\widetilde{\mathbf{w}}_u$ and uses its unit norm. The final inequality applies Cauchy--Schwarz and $\|\widetilde{\mathbf{w}}_u-\widetilde{\mathbf{w}}_v\|_2^2=2(1-c_{uv})$. For $c_{uv}<1$, rearranging and squaring gives
\begin{equation}
    \|\boldsymbol{z}_i-\widetilde{\mathbf{w}}_u\|_2
    \geq\frac{1-c_{uv}}{\sqrt{2(1-c_{uv})}}
    \quad\Longrightarrow\quad
    \|\boldsymbol{z}_i-\widetilde{\mathbf{w}}_u\|_2^2
    \geq\frac{1-c_{uv}}{2}.
\end{equation}
When $c_{uv}=1$, the resulting inequality $1-c_{uv}\leq2\|\boldsymbol{z}_i-\widetilde{\mathbf{w}}_u\|_2^2$ also holds since its left-hand side is zero.
\end{proof}

\begin{theoremstatementbox}
\textbf{Thm.\ref{thm:mmo_mismapping_bound} (Upper bound on semantic mismapping).}
\begingroup
\itshape
Under Assump.\ref{ass:mmo_analysis}, define
\begin{equation*}
    B:=\sqrt{\frac{\mathbb{KL}_{\mathrm{RB}}(p\Vert\bar q)}{2}}
    +\alpha\mathcal{L}_{\mathrm{ND}}+\epsilon_{\mathrm{res}},
\end{equation*}
where $\alpha:=\pi_{\mathrm{new}}^{*}C_{\mathrm{new}}(C_{\mathrm{new}}-1)$, $\pi_{\mathrm{new}}^{*}:=\max_{y\in\mathcal{Y}_{\mathrm{new}}}\pi_y$, and $\pi_y$ denotes the sample proportion of category $y$.
Then, the matching gain satisfies
\begin{equation*}
    \Delta_{\mathrm{all}}\leq\min\bigl\{B,[2B-\pi_*]_+\bigr\},
\end{equation*}
where $[x]_+:=\max\{x,0\}$ and
\[
    \pi_*:=\min_{c\in\mathcal{Y}_{\mathrm{old}},\,y\in\mathcal{Y}\setminus\{c\}}(\pi_c+\pi_y).
\]
Furthermore, for every permutation cycle $\gamma$, we have
\begin{equation*}
    \frac{m_\gamma}{N_u}
    \geq\sum_{y\in\mathcal{Y}_\gamma}\pi_y-2B.
\end{equation*}
\endgroup
\end{theoremstatementbox}

\begin{proof}
\textbf{Setup and residual errors.}
Let $\mathcal{A}_0$ be the highest-scoring semantics-preserving assignment from Thm.\ref{thm:semantic_mismapping_margin}, and set $t(y):=\mathcal{A}_0^{-1}(y)$.
Define the empirical category proportions by
\begin{equation}
    \pi_y:=\frac{1}{N_u}\sum_{i=1}^{N_u}\mathbb{I}\{y_i=y\},
    \qquad
    \pi_{\mathrm{old}}:=\sum_{y\in\mathcal{Y}_{\mathrm{old}}}\pi_y.
\end{equation}
Minimizing RBL and NDL alone does not guarantee correct category predictions, so the error bound also includes four residual terms:
\begin{equation}
    \begin{aligned}
        \epsilon_{\mathrm{old}}&:=\frac{1}{N_u}
        \sum_{\substack{1\leq i\leq N_u\\y_i\in\mathcal{Y}_{\mathrm{old}}}}
        \mathbb{I}\{\hat y_i\neq t(y_i)\},
        &\epsilon_{\mathrm{prior}}&:=|p_{\mathrm{old}}-\pi_{\mathrm{old}}|,\\
        \epsilon_{\mathrm{route}}&:=\frac{1}{N_u}\sum_{i=1}^{N_u}
        \left|q_i^{\mathrm{old}}-
        \mathbb{I}\{\hat y_i\in\widehat{\mathcal{Y}}_{\mathrm{old}}\}\right|,
        &\epsilon_{\mathrm{align}}&:=\frac{1}{N_u}
        \sum_{\substack{1\leq i\leq N_u\\y_i\in\mathcal{Y}_{\mathrm{new}}}}
        \|\boldsymbol{z}_i-\widetilde{\mathbf{w}}_{t(y_i)}\|_2^2.
    \end{aligned}
    \label{eq:mmo_residual_components}
\end{equation}
We explain below what each residual measures and why minimizing the two regularizers alone does not force it to vanish.
\begin{itemize}[leftmargin=*,topsep=3pt,itemsep=3pt,parsep=0pt]
\item $\epsilon_{\mathrm{old}}$ measures errors relative to old outputs. RBL controls the averaged regional probability and NDL controls new-output prototypes. Hence, neither directly enforces correct old-category predictions. Instead, supervision from the labeled set can help reduce this term.

\item $\epsilon_{\mathrm{prior}}$ measures the discrepancy between the estimated old-category proportion $p_{\mathrm{old}}$ and its actual value $\pi_{\mathrm{old}}$. RBL uses $p_{\mathrm{old}}$ as a fixed target, and neither regularizer corrects its estimation error. More accurate estimation of $p_{\mathrm{old}}$, for example using TIcE~\cite{BekkerDavis2018TIcE} as in Rem.\ref{rem:rbl_prior}, can help reduce this term.

\item $\epsilon_{\mathrm{route}}$ measures the average discrepancy between the soft old-region probability $q_i^{\mathrm{old}}$ and the hard prediction indicator $\mathbb{I}\{\hat y_i\in\widehat{\mathcal{Y}}_{\mathrm{old}}\}$. RBL matches average probability mass, and neither regularizer directly enforces agreement with individual argmax decisions. Prediction sharpening is a potential way to reduce this discrepancy, which we leave for future work.

\item $\epsilon_{\mathrm{align}}$ measures feature--prototype misalignment of new categories under $\mathcal{A}_0$. NDL separates prototypes and RBL controls regional probability totals; neither directly aligns features with their corresponding prototypes. Self-distillation with reliable pseudo-labels, as used in SimGCD~\cite{DBLP:conf/iccv/WenZQ23}, can help reduce this term.
\end{itemize}
The following derivation shows how each term enters the prediction-error bound.

\textbf{Step 0: prediction-error decomposition.}
Under $\mathcal A_0$, every incorrect prediction belongs to one of three disjoint groups: errors on old-category samples, new-category samples predicted in the old region, or new-category samples assigned to an incorrect new output. Their proportions among all $N_u$ samples are $\epsilon_{\mathrm{old}}$, $e_{\mathrm{new}\to\mathrm{old}}$, and $e_{\mathrm{nn}}$, respectively. Thus, we have
\begin{equation}
    e_{\mathrm{SPE}}:=1-\texttt{All}_{\mathrm{SPE}}
    =\underbrace{\epsilon_{\mathrm{old}}}_{\text{Errors on old samples}}+\underbrace{e_{\mathrm{new}\to\mathrm{old}}+e_{\mathrm{nn}}}_{\text{Errors on novel samples}}.
    \label{eq:mmo_total_error_decomposition}
\end{equation}
Next, we bound $e_{\mathrm{new}\to\mathrm{old}}$ in Step~1 and $e_{\mathrm{nn}}$ in Step~2, then use these bounds to prove Eqs.\eqref{eq:mmo_gap_bound} and~\eqref{eq:mmo_cycle_bound} in Steps~3.1 and~3.2, respectively.

\textbf{Step 1: regional balance and cross-region errors.}
Let $e_{\mathrm{old}\to\mathrm{new}}$ denote the fraction of all samples that belong to old categories but are predicted in the new region. Samples predicted in the old region come from two groups: samples from old categories predicted within this region and samples from new categories incorrectly predicted in this region. Thus,
\begin{equation*}
    \frac{1}{N_u}\sum_{i=1}^{N_u}
    \mathbb{I}\{\hat y_i\in\widehat{\mathcal{Y}}_{\mathrm{old}}\}
    =\pi_{\mathrm{old}}-e_{\mathrm{old}\to\mathrm{new}}
    +e_{\mathrm{new}\to\mathrm{old}}.
\end{equation*}
Subtracting $\pi_{\mathrm{old}}$ from both sides gives
\begin{equation}
    e_{\mathrm{new}\to\mathrm{old}}-e_{\mathrm{old}\to\mathrm{new}}
    =\frac{1}{N_u}\sum_{i=1}^{N_u}
    \mathbb{I}\{\hat y_i\in\widehat{\mathcal{Y}}_{\mathrm{old}}\}-\pi_{\mathrm{old}}.
\end{equation}
Furthermore, we have
\begin{equation}
    \begin{aligned}
        &|e_{\mathrm{new}\to\mathrm{old}}-e_{\mathrm{old}\to\mathrm{new}}|\\
        &\quad=\Bigg|\frac{1}{N_u}\sum_{i=1}^{N_u}
        \left(\mathbb{I}\{\hat y_i\in\widehat{\mathcal{Y}}_{\mathrm{old}}\}-q_i^{\mathrm{old}}\right)
        +(\bar q^{\,\mathrm{old}}-p_{\mathrm{old}})+(p_{\mathrm{old}}-\pi_{\mathrm{old}})\Bigg|\\
        &\quad\leq\frac{1}{N_u}\sum_{i=1}^{N_u}
        \left|\mathbb{I}\{\hat y_i\in\widehat{\mathcal{Y}}_{\mathrm{old}}\}-q_i^{\mathrm{old}}\right|
        +|\bar q^{\,\mathrm{old}}-p_{\mathrm{old}}|+|p_{\mathrm{old}}-\pi_{\mathrm{old}}|\\
        &\quad=\epsilon_{\mathrm{route}}+|\bar q^{\,\mathrm{old}}-p_{\mathrm{old}}|+\epsilon_{\mathrm{prior}}\\
        &\quad\leq\epsilon_{\mathrm{route}}+\sqrt{\frac{\mathbb{KL}_{\mathrm{RB}}(p\Vert\bar q)}{2}}+\epsilon_{\mathrm{prior}}.
    \end{aligned}
    \label{eq:mmo_routing_bound}
\end{equation}
The first equality adds and subtracts $\bar q^{\,\mathrm{old}}$ and $p_{\mathrm{old}}$, using $\bar q^{\,\mathrm{old}}=N_u^{-1}\sum_{i=1}^{N_u}q_i^{\mathrm{old}}$. The first inequality applies the triangle inequality, and the second equality substitutes the definitions of $\epsilon_{\mathrm{route}}$ and $\epsilon_{\mathrm{prior}}$. The final inequality follows from Pinsker's inequality for Bernoulli distributions, $\mathbb{KL}_{\mathrm{RB}}(p\Vert\bar q)\geq2(p_{\mathrm{old}}-\bar q^{\,\mathrm{old}})^2$, with natural logarithms. More specifically, the fixed-prior RBL in Eq.\eqref{eq:pao_balance_loss}, evaluated over $\mathcal D_u$, gives $\mathbb{KL}_{\mathrm{RB}}(p\Vert\bar q)=\mathcal L_{\mathrm{RB}}-H(p)$, where $H(p):=-\sum_r p_r\log p_r$ is the entropy of $p$.

Every sample from an old category predicted in the new region is also incorrectly assigned relative to its anchored old output. Thus, these errors have been included in $\epsilon_{\mathrm{old}}$, giving $e_{\mathrm{old}\to\mathrm{new}}\leq\epsilon_{\mathrm{old}}$. Combining this observation with Eq.\eqref{eq:mmo_routing_bound} yields
\begin{equation}
    \begin{aligned}
        e_{\mathrm{new}\to\mathrm{old}}
        &\leq e_{\mathrm{old}\to\mathrm{new}}
        +\left|e_{\mathrm{new}\to\mathrm{old}}-e_{\mathrm{old}\to\mathrm{new}}\right|\\
        &\leq\epsilon_{\mathrm{old}}+\epsilon_{\mathrm{prior}}+\epsilon_{\mathrm{route}}
        +\sqrt{\frac{\mathbb{KL}_{\mathrm{RB}}(p\Vert\bar q)}{2}}.
    \end{aligned}
    \label{eq:mmo_false_old_bound}
\end{equation}

\textbf{Step 2: prototype correlations and within-new-category errors.}
For distinct new outputs $u,v\in\widehat{\mathcal Y}_{\mathrm{new}}$, define the error proportion
\begin{equation}
    e_{uv}:=\frac{1}{N_u}\sum_{i=1}^{N_u}
    \mathbb{I}\{y_i=\mathcal A_0(u),\ \hat y_i=v\}.
    \label{eq:mmo_new_error_proportion}
\end{equation}
The corresponding error groups are disjoint and together account for all errors within the new region, so $e_{\mathrm{nn}}=\sum_{u\neq v}e_{uv}$. Unless otherwise specified, all sums over $u\neq v$ below run over distinct outputs in $\widehat{\mathcal Y}_{\mathrm{new}}$. Then, we have
\begin{equation}
    \begin{aligned}
    e_{\mathrm{nn}}
    &=\sum_{u\neq v}e_{uv}
    =\sum_{u\neq v}e_{uv}\bigl[(1-c_{uv})+c_{uv}\bigr]\\
    &=\underbrace{\sum_{u\neq v}e_{uv}(1-c_{uv})}_{(\ast)}
    +\underbrace{\sum_{u\neq v}e_{uv}c_{uv}}_{(\ast\ast)}.
    \end{aligned}
    \label{eq:mmo_error_decomposition}
\end{equation}
The first equality uses the definition of $e_{\mathrm{nn}}$. We then write $1=(1-c_{uv})+c_{uv}$ and distribute the sum, where $c_{uv}$ is the prototype cosine similarity defined in Lem.\ref{lem:mmo_prediction_distance}.

To bound $(\ast)$, we expand the sum over individual misclassified samples:
\begin{equation}
    \begin{aligned}
    (\ast)
    &=\frac{1}{N_u}
    \sum_{\substack{1\leq i\leq N_u,\ y_i\in\mathcal Y_{\mathrm{new}}\\
    \hat y_i\in\widehat{\mathcal Y}_{\mathrm{new}},\ \hat y_i\neq t(y_i)}}
    (1-c_{t(y_i),\hat y_i})\\
    &\leq\frac{2}{N_u}
    \sum_{\substack{1\leq i\leq N_u,\ y_i\in\mathcal Y_{\mathrm{new}}\\
    \hat y_i\in\widehat{\mathcal Y}_{\mathrm{new}},\ \hat y_i\neq t(y_i)}}
    \|\boldsymbol z_i-\widetilde{\mathbf w}_{t(y_i)}\|_2^2\\
    &\leq\frac{2}{N_u}
    \sum_{\substack{1\leq i\leq N_u\\y_i\in\mathcal Y_{\mathrm{new}}}}
    \|\boldsymbol z_i-\widetilde{\mathbf w}_{t(y_i)}\|_2^2
    =2\epsilon_{\mathrm{align}}.
    \end{aligned}
    \label{eq:mmo_alignment_error_bound}
\end{equation}
The first inequality applies Lem.\ref{lem:mmo_prediction_distance} with $u=t(y_i)$ and $v=\hat y_i$. The second includes all new-category samples in the sum. The final equality uses the definition of $\epsilon_{\mathrm{align}}$.

To bound $(\ast\ast)$, we use weighted Cauchy--Schwarz and NDL:
\begin{equation}
    \begin{aligned}
        (\ast\ast)
        &\leq\sqrt{e_{\mathrm{nn}}\sum_{u\neq v}e_{uv}c_{uv}^2}\\
        &\leq\sqrt{e_{\mathrm{nn}}\pi_{\mathrm{new}}^{*}\sum_{u\neq v}c_{uv}^2}\\
        &=\sqrt{e_{\mathrm{nn}}\alpha\mathcal{L}_{\mathrm{ND}}}.
    \end{aligned}
    \label{eq:mmo_correlation_bound}
\end{equation}
For the first inequality, write $e_{uv}c_{uv}=\sqrt{e_{uv}}(\sqrt{e_{uv}}c_{uv})$ and apply Cauchy--Schwarz to these two factors:
\begin{equation*}
    \left|\sum_{u\neq v}\sqrt{e_{uv}}\bigl(\sqrt{e_{uv}}c_{uv}\bigr)\right|
    \leq\sqrt{\sum_{u\neq v}e_{uv}}\,
    \sqrt{\sum_{u\neq v}e_{uv}c_{uv}^2}
    =\sqrt{e_{\mathrm{nn}}\sum_{u\neq v}e_{uv}c_{uv}^2}.
\end{equation*}
Since $(\ast\ast)\leq|(\ast\ast)|$, this also bounds the original correlation sum.
For the second inequality, all misclassified samples contributing to $e_{uv}$ belong to category $\mathcal A_0(u)$. Their proportion cannot exceed that of the entire category, so $e_{uv}\leq\pi_{\mathcal A_0(u)}$. In turn, $\pi_{\mathcal A_0(u)}\leq\pi_{\mathrm{new}}^{*}:=\max_{y\in\mathcal Y_{\mathrm{new}}}\pi_y$, the proportion of samples belonging to the largest new category. Replacing each weight $e_{uv}$ by $\pi_{\mathrm{new}}^{*}$ gives the inequality.
For the final equality, NDL in Eq.\eqref{eq:pao_novel_diversity} is the average squared cosine similarity over the $C_{\mathrm{new}}(C_{\mathrm{new}}-1)$ ordered choices of distinct new outputs. Thus, $\sum_{u\neq v}c_{uv}^2=C_{\mathrm{new}}(C_{\mathrm{new}}-1)\mathcal L_{\mathrm{ND}}$. Multiplying by $\pi_{\mathrm{new}}^{*}$ and using $\alpha=\pi_{\mathrm{new}}^{*}C_{\mathrm{new}}(C_{\mathrm{new}}-1)$ gives the equality.

Substituting the bounds on $(\ast)$ and $(\ast\ast)$ in Eqs.\eqref{eq:mmo_alignment_error_bound} and~\eqref{eq:mmo_correlation_bound} into Eq.\eqref{eq:mmo_error_decomposition} gives
\begin{equation}
    \begin{aligned}
    e_{\mathrm{nn}}&\leq2\epsilon_{\mathrm{align}}
    +\sqrt{e_{\mathrm{nn}}\alpha\mathcal L_{\mathrm{ND}}}\\
    &\leq2\epsilon_{\mathrm{align}}
    +\frac{e_{\mathrm{nn}}+\alpha\mathcal{L}_{\mathrm{ND}}}{2}.
    \end{aligned}
\end{equation}
The second inequality uses $\sqrt{xy}\leq(x+y)/2$ for nonnegative $x,y$. Subtracting $e_{\mathrm{nn}}/2$ from both sides and multiplying by two yields
\begin{equation}
    e_{\mathrm{nn}}
    \leq4\epsilon_{\mathrm{align}}
    +\alpha\mathcal{L}_{\mathrm{ND}}.
    \label{eq:mmo_novel_error_bound}
\end{equation}

\textbf{Step 3.1: matching-gain bound $\rightarrow$ Eq.\eqref{eq:mmo_gap_bound}.}
Substituting the new-to-old error bound in Eq.\eqref{eq:mmo_false_old_bound} and the within-new-category error bound in Eq.\eqref{eq:mmo_novel_error_bound} into the decomposition in Eq.\eqref{eq:mmo_total_error_decomposition} gives
\begin{equation}
    \begin{aligned}
    e_{\mathrm{SPE}}
    &\leq\sqrt{\frac{\mathbb{KL}_{\mathrm{RB}}(p\Vert\bar q)}{2}}
    +\alpha\mathcal L_{\mathrm{ND}}
    +\underbrace{2\epsilon_{\mathrm{old}}+\epsilon_{\mathrm{prior}}
    +\epsilon_{\mathrm{route}}+4\epsilon_{\mathrm{align}}}_{\epsilon_{\mathrm{res}}}.
    \end{aligned}
    \label{eq:mmo_residual}
\end{equation}
Thus, $e_{\mathrm{SPE}}\leq B$.
Since GH accuracy cannot exceed one, its improvement over SPE satisfies
\begin{equation*}
    \Delta_{\mathrm{all}}
    =\texttt{All}_{\mathrm{GH}}-\texttt{All}_{\mathrm{SPE}}
    \leq1-\texttt{All}_{\mathrm{SPE}}
    =e_{\mathrm{SPE}}\leq B.
\end{equation*}

To sharpen this matching-gain bound, we examine which reassignments can increase the score. Following Thm.\ref{thm:semantic_mismapping_margin}, we express the changes from $\mathcal A_0$ to a GH-optimal assignment as disjoint cycles. With the old-output assignments fixed, $\mathcal A_0$ already gives the highest score by matching new outputs to new categories. A cycle involving only new categories merely changes this new-output matching and therefore cannot increase the score. Any cycle that increases the score must consequently change at least one old-output assignment and thus contain an old category. To quantify the gain from these cycles, define the number of errors on category $y$ under $\mathcal A_0$ as
\begin{equation}
    E_y:=\sum_{i=1}^{N_u}\mathbb I\{y_i=y,\ \hat y_i\neq t(y)\}.
    \label{eq:mmo_category_error_count}
\end{equation}
Category $y$ contributes $N_u\pi_y-E_y$ correct predictions under $\mathcal A_0$, but at most $E_y$ when matched to any output $v\neq t(y)$. When $\Delta_{\mathrm{all}}>0$, let $S$ collect the categories in all score-increasing cycles. Since $S$ contains an old category and at least one other category, the definition of $\pi_*$ gives $\sum_{y\in S}\pi_y\geq\pi_*$. Summing the gains over these disjoint cycles then yields
\begin{equation}
    \begin{aligned}
        \Delta_{\mathrm{all}}
        &\leq\frac{1}{N_u}\left[\sum_{y\in S}E_y-\sum_{y\in S}(N_u\pi_y-E_y)\right]\\
        &=\frac{2}{N_u}\sum_{y\in S}E_y-\sum_{y\in S}\pi_y\\
        &\leq2e_{\mathrm{SPE}}-\sum_{y\in S}\pi_y\\
        &\leq2e_{\mathrm{SPE}}-\pi_*\leq2B-\pi_*.
    \end{aligned}
\end{equation}
The first inequality omits cycles that do not increase the score and bounds the gain on the remaining categories. Before reassignment, these categories contribute $\sum_{y\in S}(N_u\pi_y-E_y)$ correct predictions. After reassignment, each category $y$ is matched to an output other than $t(y)$ and contributes at most $E_y$ correct predictions. Their difference, divided by $N_u$, gives the inequality. Since the cycles are disjoint, each category in $S$ is counted once; collecting the terms gives the next equality. The subsequent inequalities use $\sum_{y\in S}E_y\leq N_u e_{\mathrm{SPE}}$, $\sum_{y\in S}\pi_y\geq\pi_*$, and $e_{\mathrm{SPE}}\leq B$, respectively. If $\Delta_{\mathrm{all}}=0$, the bound $\Delta_{\mathrm{all}}\leq[2B-\pi_*]_+$ holds directly. 

Combining both cases with $\Delta_{\mathrm{all}}\leq B$ proves Eq.\eqref{eq:mmo_gap_bound}, which bounds the matching gain characterized by the cycle decomposition in Thm.\ref{thm:semantic_mismapping_margin}.
If $2B<\pi_*$, GH and SPE have the same accuracy because $\Delta_{\mathrm{all}}=0$. When $2B=\pi_*$, a reassignment may instead tie with $\mathcal A_0$, allowing semantic mismapping without changing accuracy.

\textbf{Step 3.2: cycle-margin bound $\rightarrow$ Eq.\eqref{eq:mmo_cycle_bound}.}
We next bound the margin of an arbitrary cycle. For a cycle $\gamma=(y_1,\ldots,y_k)$ with $y_{k+1}=y_1$,
\begin{equation}
    \begin{aligned}
        m_\gamma
        &=\sum_{j=1}^{k}\left(C_{y_j\rightarrow t(y_j)}-C_{y_{j+1}\rightarrow t(y_j)}\right)\\
        &\geq\sum_{y\in\mathcal Y_\gamma}(N_u\pi_y-E_y)
        -\sum_{y\in\mathcal Y_\gamma}E_y\\
        &= N_u\sum_{y\in\mathcal{Y}_\gamma}\pi_y
        -2\sum_{y\in\mathcal{Y}_\gamma}E_y\\
        &\geq N_u\left(\sum_{y\in\mathcal Y_\gamma}\pi_y-2e_{\mathrm{SPE}}\right)\\
        &\geq N_u\left(\sum_{y\in\mathcal{Y}_\gamma}\pi_y-2B\right),
    \end{aligned}
\end{equation}
The first equality recalls that the cycle margin compares the numbers of correct predictions before and after reassignment. Under $\mathcal A_0$, output $t(y_j)$ correctly classifies $C_{y_j\rightarrow t(y_j)}=N_u\pi_{y_j}-E_{y_j}$ samples. The cycle reassigns this output to category $y_{j+1}$, so its correct predictions become $C_{y_{j+1}\rightarrow t(y_j)}$. Because $t(y_j)\neq t(y_{j+1})$, these samples were errors under $\mathcal A_0$, and their number is at most $E_{y_{j+1}}$. Substituting these counts gives the first inequality. As the cycle visits each category once, $\sum_{j=1}^{k}E_{y_{j+1}}=\sum_{y\in\mathcal Y_\gamma}E_y$, so the two error sums combine into $2\sum_{y\in\mathcal Y_\gamma}E_y$. This sum can then be bounded using the total number of errors, since $\sum_{y\in\mathcal Y_\gamma}E_y\leq\sum_{y\in\mathcal Y}E_y=N_u e_{\mathrm{SPE}}$. Substituting this upper bound into the subtracted term, followed by $e_{\mathrm{SPE}}\leq B$, gives the final two inequalities.

Finally, dividing by $N_u$ proves Eq.\eqref{eq:mmo_cycle_bound}, which bounds the cycle margin in Thm.\ref{thm:semantic_mismapping_margin}.

\end{proof}

\endgroup

\begingroup
\def\IncludeCompleteBenchmarkTables{}
\def\IncludeFullOptimizationTables{}
\newcommand{\AppendixShade}[2]{%
    \ifdim#2pt<0pt
        \def\AppFill{SPELossFill}\dimen0=-#2pt\relax
    \else
        \def\AppFill{SPEGainFill}\dimen0=#2pt\relax
    \fi
    \ifdim\dimen0<1pt \def\AppShade{10}%
    \else\ifdim\dimen0<3pt \def\AppShade{23}%
    \else\ifdim\dimen0<7pt \def\AppShade{38}%
    \else\ifdim\dimen0<15pt \def\AppShade{55}%
    \else \def\AppShade{75}\fi\fi\fi\fi
    \ifdim#2pt>0pt\ifdim\dimen0<15pt\else\def\AppShade{60}\fi\fi
    \ifdim\dimen0=0pt \def\AppFill{SPELossFill}\def\AppShade{5}\fi
    \edef\AppPaint{\noexpand\cellcolor{\AppFill!\AppShade!white}}%
    \AppPaint #1%
}

\ifdefined\IncludeCompleteBenchmarkTables
\section{Additional Experimental Setup}
\label{app:additional_experimental_setup}
\label{app:complete_benchmark}

\subsection{Datasets}
\label{app:additional_datasets}

We evaluate our method on eight standard GCD datasets: CIFAR-10, CIFAR-100, ImageNet-100, ImageNet-1K, CUB-200-2011, Stanford Cars, FGVC-Aircraft, and Herbarium 19. They cover coarse-grained recognition, fine-grained recognition, and large-scale category discovery; their detailed properties and the GCD splits used in our experiments are introduced below.

\textbf{CIFAR-10.}
CIFAR-10~\cite{cifar10} contains $32\times32$ natural images from 10 coarse-grained object categories. Following the standard GCD split, 12.5K labeled images come from 5 old categories, while 37.5K unlabeled images cover all 10 categories and therefore include both old and new categories.

\textbf{CIFAR-100.}
CIFAR-100~\cite{cifar10} extends CIFAR-10 to 100 object categories and provides a more challenging coarse-grained benchmark. We use 20.0K labeled images from 80 old categories and 30.0K unlabeled images from all 100 categories, leaving 20 categories to be discovered as new categories.

\textbf{ImageNet-100.}
ImageNet-100 is a 100-category subset of ImageNet~\cite{imagenet,DBLP:journals/ijcv/RussakovskyDSKS15} with greater visual diversity and higher-resolution images than CIFAR. Its GCD split contains 31.9K labeled images from 50 old categories and 95.3K unlabeled images covering all 100 categories.

\textbf{ImageNet-1K.}
ImageNet-1K~\cite{imagenet,DBLP:journals/ijcv/RussakovskyDSKS15} contains 1,000 object categories and serves as our large-scale benchmark. We use 321K labeled images from 500 old categories and 960K unlabeled images spanning all 1,000 categories, which tests the scalability of GCD methods when many new categories must be identified.

\textbf{CUB-200-2011.}
CUB-200-2011 (CUB)~\cite{cub} is a fine-grained dataset containing 200 bird species whose distinctions often depend on subtle shape, color, and texture cues. The adopted split uses 1.5K labeled images from 100 old species and 4.5K unlabeled images from all 200 species.

\textbf{Stanford Cars.}
Stanford Cars~\cite{stanfordcars} contains 196 fine-grained car categories defined by make, model, and year. We use 2.0K labeled images from 98 old categories and 6.1K unlabeled images covering all 196 categories, making the dataset useful for evaluating discovery under small inter-class differences.

\textbf{FGVC-Aircraft.}
FGVC-Aircraft~\cite{fgvcair} comprises 100 aircraft variants with highly similar global appearances and discriminative local details. Its GCD split includes 1.7K labeled images from 50 old variants and 5.0K unlabeled images from all 100 variants.

\textbf{Herbarium 19.}
Herbarium 19~\cite{HerbariumChallenge2019} contains 683 plant species and exhibits both fine-grained visual differences and substantial class imbalance. We use 8.9K labeled images from 341 old species and 25.4K unlabeled images from all 683 species, providing a challenging test of GCD under a naturally long-tailed distribution.

\subsection{Evaluated Methods}
\label{app:additional_compared_methods}

We evaluate and compare seven method--backbone configurations. The four DINO-based baselines are LegoGCD--DINO, SimGCD-BIA--DINO, ProtoGCD--DINO, and SimGCD--DINO; the three CLIP-based baselines are SimGCD--CLIP, GET--CLIP, and GET-TAR--CLIP. These configurations cover representative parametric classifiers, prototype-based learning, regularized GCD methods, and multimodal approaches. We apply MMO to every baseline while preserving its original objective, backbone, and training pipeline. The individual baselines are introduced below.

\noindent\colorbox{black!8}{\textit{DINO-based baselines.}}

\textbf{LegoGCD--DINO.}
LegoGCD~\cite{DBLP:conf/cvpr/CaoZWYS0L024} builds on SimGCD and focuses on reducing catastrophic forgetting of old categories. With a DINO backbone, it introduces local entropy regularization and dual-view prediction consistency to preserve old-category knowledge while learning new categories, without changing the basic classifier architecture.

\textbf{SimGCD-BIA--DINO.}
This configuration augments the DINO-based SimGCD baseline with Bures-Isotropy Alignment (BIA)~\cite{tang2026bia}. BIA regularizes the feature geometry toward an isotropic structure on unlabeled data, reducing dimensional collapse and improving the separation of old and new categories while preserving the original SimGCD training framework.

\textbf{ProtoGCD--DINO.}
ProtoGCD~\cite{ma2025protogcd} uses a unified prototype-learning framework to represent both old and new categories. Its debiased prototype learning reduces the imbalance between supervised old-category signals and unsupervised new-category signals, with DINO features providing the visual initialization in this configuration.

\textbf{SimGCD--DINO.}
SimGCD~\cite{DBLP:conf/iccv/WenZQ23} is a parametric GCD baseline that jointly uses labeled old-category supervision and self-distillation on unlabeled data. In the DINO configuration, a learnable classifier is trained on DINO-pretrained visual features to predict a shared output space containing both old and new categories.

\noindent\colorbox{black!8}{\textit{CLIP-based baselines.}}

\textbf{SimGCD--CLIP.}
This configuration uses the SimGCD classifier, objective, and optimization strategy but replaces the DINO backbone with the CLIP visual encoder~\cite{DBLP:conf/icml/RadfordKHRGASAM21}. It allows us to isolate how the representation learned through vision--language pretraining affects parametric GCD performance.

\textbf{GET--CLIP.}
GET~\cite{wang2025get} is designed to exploit the multimodal knowledge of CLIP for generalized category discovery. It introduces textual semantics into the discovery process so that visual samples can be organized with richer semantic guidance than that available from image-only supervision.

\textbf{GET-TAR--CLIP.}
GET-TAR applies token-aware refinement~\cite{yang2026tar} to the CLIP-based GET training pipeline~\cite{wang2025get}. The refinement suppresses distracting attention patterns and strengthens informative local cues, which is particularly helpful for recognizing and discovering visually similar fine-grained categories.

\subsection{Implementation Details}
\label{app:baseline_hyperparameters}

\noindent\colorbox{black!8}{\textit{General implementation details.}}

Our implementation follows standard practices in recent GCD methods~\cite{DBLP:conf/iccv/WenZQ23,wang2025get}. Image preprocessing, including random cropping, random horizontal flipping, and normalization, follows the settings of the corresponding baselines. We train for 200 epochs (per stage for GET-based methods) and use the final checkpoint for evaluation. For MMO, we select $\lambda_{\mathrm{RB}}$ and $\lambda_{\mathrm{ND}}$ via grid search over $\lambda_{\mathrm{RB}}\in\{0.25,0.5,0.75\}$ and $\lambda_{\mathrm{ND}}\in\{0.01,0.05,0.1\}$. We use the same random seeds as the original baseline implementations. All experiments are conducted on NVIDIA GeForce RTX 3090 GPUs.

\noindent\colorbox{black!8}{\textit{Baseline-Specific Hyperparameters.}}

\textbf{SimGCD.}
Following SimGCD~\cite{DBLP:conf/iccv/WenZQ23}, we set the supervised--unsupervised balancing coefficient $\lambda$ to $0.35$; the supervised and unsupervised terms receive weights $\lambda$ and $1-\lambda$, respectively, in both representation learning and classification. The self-supervised and supervised contrastive temperatures are $\tau_u=0.07$ and $\tau_c=1.0$, respectively. In the classification objective, $\tau_s=0.1$ controls the student prediction distribution, while $\tau_t$ controls the teacher pseudo-target distribution. The teacher temperature follows a linear warm-up schedule from $0.07$ to $0.04$ during the first 30 epochs. SimGCD--CLIP uses the same settings with the CLIP visual encoder.

\textbf{LegoGCD.}
LegoGCD~\cite{DBLP:conf/cvpr/CaoZWYS0L024} preserves the SimGCD losses and adds local entropy regularization (LER) to protect potential old-class samples, together with a dual-view Kullback--Leibler (DKL) constraint in the classification loss. Its objective weights the SimGCD representation and classification losses by $\alpha=1$ and LER by $\beta$. The original paper selects $\beta=2$ on CUB and $\beta=1$ on CIFAR-100; it also reports a confidence threshold $\delta=0.85$ for selecting potential old-class samples on Stanford Cars and CIFAR-100. These are LegoGCD-specific parameters, distinct from the MMO regularization weights.

\textbf{SimGCD-BIA.}
Bures-Isotropy Alignment (BIA)~\cite{tang2026bia} preserves the SimGCD objective and adds a geometry regularizer on the projected class tokens of the unlabeled minibatch. Under the trace constraint described in the BIA paper, this term maximizes the nuclear norm of the stacked tokens; its implementation subtracts $\lambda_{\mathrm{BIA}}\|Z_u\|_*$ from the baseline loss. The paper uses $\lambda_{\mathrm{BIA}}=0.004$ and otherwise follows the underlying GCD method's optimization settings.

\textbf{ProtoGCD.}
Following ProtoGCD~\cite{ma2025protogcd}, the supervised component is weighted by $\lambda_{\mathrm{sup}}=0.35$. Its entropy and prototype-separation regularizers use $\lambda_{\mathrm{entropy}}=2$ and $\lambda_{\mathrm{sep}}=0.1$, respectively. The base and separation temperatures are $\tau_{\mathrm{base}}=\tau_{\mathrm{sep}}=0.1$, the pseudo-label sharpening temperature is $\tau_{\mathrm{sharp}}=0.05$, and the dual-level pseudo-label schedule has a linear 100-epoch ramp-up.

\textbf{GET.}
Following GET~\cite{wang2025get}, the Text Embedding Synthesizer (TES) maps visual embeddings to seven pseudo-text tokens, which are passed to the CLIP text encoder. The coefficient $\lambda=0.35$ balances the supervised and unsupervised objectives in each modality branch. The coefficient $\lambda_c=1$ weights the cross-modal instance consistency loss, and $\tau_a=0.01$ is the temperature in the visual--textual alignment loss used to train TES. The remaining contrastive and student/teacher temperatures follow the corresponding SimGCD settings.

\textbf{GET-TAR.}
GET-TAR~\cite{yang2026tar} builds on GET and applies token reweighting, consistency-aware token sampling, and gated residual refinement before classification. The official GET-TAR implementation preserves the GET optimization settings, selects the top half of visual tokens ($\texttt{top\_m\_ratio}=0.5$), and weights the additional sampling-distribution KL loss by $0.001$. 
\fi

\ifdefined\IncludeFullOptimizationTables
\section{Additional Experimental Results}
\label{app:additional_experimental_results}
\label{app:optimization_results}

\subsection{Additional Comparison between GH and SPE (RQ1)}
\label{app:detailed_gh_spe_comparison}

Tab.\ref{tab:appendix_gh_spe_complete} provides the complete old-category, new-category, and overall results under GH and SPE. We define every difference as $\Delta_r=\texttt{Acc}_{\mathrm{GH}}^r-\texttt{Acc}_{\mathrm{SPE}}^r$.
It is important to distinguish $\Delta_{\mathrm{new}}$ from $\Delta_{\mathrm{all}}$. The new-category gap $\Delta_{\mathrm{new}}=\texttt{New}_{\mathrm{GH}}-\texttt{New}_{\mathrm{SPE}}$ can be positive, zero, or negative because GH maximizes accuracy over all categories rather than over the new subset alone: a global assignment may improve old-category accuracy while sacrificing some new-category predictions. By contrast, $\Delta_{\mathrm{all}}=\texttt{All}_{\mathrm{GH}}-\texttt{All}_{\mathrm{SPE}}\geq 0$, since the assignment space optimized by GH contains the semantics-preserving assignments allowed by SPE.

\begin{table}[!htbp]
    \centering
    \footnotesize
    \caption{Detailed comparison between GH and SPE on coarse- and fine-grained benchmarks. GH--SPE differences are orange (positive) or blue (negative/zero).}
    \label{tab:appendix_gh_spe_complete}
    \setlength{\tabcolsep}{2.1pt}
    \renewcommand{\arraystretch}{1.12}
    \fitTable{%
    \begin{tabular}{@{}lllrrrrrrrrr@{}}
        \toprule
        \multirow{2}{*}{Dataset} & \multirow{2}{*}{Method} & \multirow{2}{*}{Backbone}
        & \multicolumn{3}{c}{Old categories}
        & \multicolumn{3}{c}{New categories}
        & \multicolumn{3}{c}{All categories} \\
        \cmidrule(lr){4-6}\cmidrule(lr){7-9}\cmidrule(lr){10-12}
        & & & $\texttt{Old}_{\mathrm{GH}}$ & $\texttt{Old}_{\mathrm{SPE}}$ & $\Delta_{\mathrm{old}}$
        & $\texttt{New}_{\mathrm{GH}}$ & $\texttt{New}_{\mathrm{SPE}}$ & $\Delta_{\mathrm{new}}$
        & $\texttt{All}_{\mathrm{GH}}$ & $\texttt{All}_{\mathrm{SPE}}$ & $\Delta_{\mathrm{all}}$ \\
        \midrule
        \multicolumn{12}{@{}l}{\textit{(a) Coarse-grained benchmarks}} \\
        \multirow{7}{*}{CIFAR-10} & LegoGCD & DINO & 95.76 & 95.76 & \AppendixShade{0.00}{0.00} & 97.93 & 97.93 & \AppendixShade{0.00}{0.00} & 97.21 & 97.21 & \AppendixShade{0.00}{0.00} \\
         & SimGCD-BIA & DINO & 84.25 & 84.25 & \AppendixShade{0.00}{0.00} & 98.74 & 98.74 & \AppendixShade{0.00}{0.00} & 93.91 & 93.91 & \AppendixShade{0.00}{0.00} \\
         & ProtoGCD & DINO & 81.94 & 81.94 & \AppendixShade{0.00}{0.00} & 98.94 & 98.94 & \AppendixShade{0.00}{0.00} & 93.27 & 93.27 & \AppendixShade{0.00}{0.00} \\
         & SimGCD & DINO & 93.71 & 93.71 & \AppendixShade{0.00}{0.00} & 98.48 & 98.48 & \AppendixShade{0.00}{0.00} & 96.89 & 96.89 & \AppendixShade{0.00}{0.00} \\
         & SimGCD & CLIP & 94.75 & 94.75 & \AppendixShade{0.00}{0.00} & 98.11 & 98.11 & \AppendixShade{0.00}{0.00} & 96.99 & 96.99 & \AppendixShade{0.00}{0.00} \\
         & GET-TAR & CLIP & 94.70 & 94.70 & \AppendixShade{0.00}{0.00} & 98.42 & 98.42 & \AppendixShade{0.00}{0.00} & 97.18 & 97.18 & \AppendixShade{0.00}{0.00} \\
         & GET & CLIP & 94.87 & 94.87 & \AppendixShade{0.00}{0.00} & 98.36 & 98.36 & \AppendixShade{0.00}{0.00} & 97.19 & 97.19 & \AppendixShade{0.00}{0.00} \\
        \cmidrule(lr){2-12}
        \multirow{7}{*}{CIFAR-100} & LegoGCD & DINO & 81.71 & 81.71 & \AppendixShade{0.00}{0.00} & 85.99 & 85.99 & \AppendixShade{0.00}{0.00} & 83.13 & 83.13 & \AppendixShade{0.00}{0.00} \\
         & SimGCD-BIA & DINO & 75.50 & 75.50 & \AppendixShade{0.00}{0.00} & 78.60 & 78.60 & \AppendixShade{0.00}{0.00} & 76.53 & 76.53 & \AppendixShade{0.00}{0.00} \\
         & ProtoGCD & DINO & 77.82 & 77.82 & \AppendixShade{0.00}{0.00} & 82.49 & 82.49 & \AppendixShade{0.00}{0.00} & 79.38 & 79.38 & \AppendixShade{0.00}{0.00} \\
         & SimGCD & DINO & 81.69 & 81.69 & \AppendixShade{0.00}{0.00} & 74.02 & 74.02 & \AppendixShade{0.00}{0.00} & 79.13 & 79.13 & \AppendixShade{0.00}{0.00} \\
         & SimGCD & CLIP & 84.05 & 84.90 & \AppendixShade{-0.85}{-0.85} & 73.75 & 71.68 & \AppendixShade{2.07}{2.07} & 80.62 & 80.50 & \AppendixShade{0.12}{0.12} \\
         & GET-TAR & CLIP & 85.68 & 85.68 & \AppendixShade{0.00}{0.00} & 71.69 & 71.69 & \AppendixShade{0.00}{0.00} & 81.01 & 81.01 & \AppendixShade{0.00}{0.00} \\
         & GET & CLIP & 85.69 & 85.69 & \AppendixShade{0.00}{0.00} & 71.12 & 71.12 & \AppendixShade{0.00}{0.00} & 80.83 & 80.83 & \AppendixShade{0.00}{0.00} \\
        \cmidrule(lr){2-12}
        \multirow{7}{*}{ImageNet-100} & LegoGCD & DINO & 94.88 & 94.88 & \AppendixShade{0.00}{0.00} & 77.76 & 77.76 & \AppendixShade{0.00}{0.00} & 83.49 & 83.49 & \AppendixShade{0.00}{0.00} \\
         & SimGCD-BIA & DINO & 86.05 & 86.05 & \AppendixShade{0.00}{0.00} & 78.70 & 78.70 & \AppendixShade{0.00}{0.00} & 81.16 & 81.16 & \AppendixShade{0.00}{0.00} \\
         & ProtoGCD & DINO & 91.00 & 91.00 & \AppendixShade{0.00}{0.00} & 78.36 & 78.36 & \AppendixShade{0.00}{0.00} & 82.59 & 82.59 & \AppendixShade{0.00}{0.00} \\
         & SimGCD & DINO & 92.58 & 92.58 & \AppendixShade{0.00}{0.00} & 81.26 & 81.26 & \AppendixShade{0.00}{0.00} & 85.04 & 85.04 & \AppendixShade{0.00}{0.00} \\
         & SimGCD & CLIP & 94.76 & 94.76 & \AppendixShade{0.00}{0.00} & 87.80 & 87.80 & \AppendixShade{0.00}{0.00} & 90.13 & 90.13 & \AppendixShade{0.00}{0.00} \\
         & GET-TAR & CLIP & 95.43 & 95.43 & \AppendixShade{0.00}{0.00} & 88.28 & 88.28 & \AppendixShade{0.00}{0.00} & 90.67 & 90.67 & \AppendixShade{0.00}{0.00} \\
         & GET & CLIP & 95.69 & 95.69 & \AppendixShade{0.00}{0.00} & 92.02 & 92.02 & \AppendixShade{0.00}{0.00} & 93.25 & 93.25 & \AppendixShade{0.00}{0.00} \\
        \midrule
        \multicolumn{12}{@{}l}{\textit{(b) Fine-grained benchmarks}} \\
        \multirow{7}{*}{Herbarium 19} & LegoGCD & DINO & 58.47 & 69.02 & \AppendixShade{-10.55}{-10.55} & 37.48 & 17.98 & \AppendixShade{19.50}{19.50} & 44.82 & 35.83 & \AppendixShade{8.99}{8.99} \\
         & SimGCD-BIA & DINO & 55.30 & 65.47 & \AppendixShade{-10.17}{-10.17} & 35.65 & 16.34 & \AppendixShade{19.31}{19.31} & 42.53 & 33.53 & \AppendixShade{9.00}{9.00} \\
         & ProtoGCD & DINO & 58.28 & 68.04 & \AppendixShade{-9.76}{-9.76} & 35.92 & 18.52 & \AppendixShade{17.40}{17.40} & 43.74 & 35.84 & \AppendixShade{7.90}{7.90} \\
         & SimGCD & DINO & 57.47 & 67.09 & \AppendixShade{-9.62}{-9.62} & 36.72 & 17.66 & \AppendixShade{19.06}{19.06} & 43.98 & 34.95 & \AppendixShade{9.03}{9.03} \\
         & SimGCD & CLIP & 64.39 & 77.34 & \AppendixShade{-12.95}{-12.95} & 38.26 & 12.63 & \AppendixShade{25.63}{25.63} & 47.40 & 35.27 & \AppendixShade{12.13}{12.13} \\
         & GET-TAR & CLIP & 65.08 & 78.59 & \AppendixShade{-13.51}{-13.51} & 37.77 & 16.46 & \AppendixShade{21.31}{21.31} & 47.33 & 38.20 & \AppendixShade{9.13}{9.13} \\
         & GET & CLIP & 63.75 & 78.52 & \AppendixShade{-14.77}{-14.77} & 41.11 & 17.10 & \AppendixShade{24.01}{24.01} & 49.03 & 38.59 & \AppendixShade{10.44}{10.44} \\
        \cmidrule(lr){2-12}
        \multirow{7}{*}{Stanford Cars} & LegoGCD & DINO & 75.11 & 79.41 & \AppendixShade{-4.30}{-4.30} & 43.83 & 40.72 & \AppendixShade{3.11}{3.11} & 54.02 & 53.32 & \AppendixShade{0.70}{0.70} \\
         & SimGCD-BIA & DINO & 71.21 & 74.91 & \AppendixShade{-3.70}{-3.70} & 45.16 & 42.19 & \AppendixShade{2.97}{2.97} & 53.65 & 52.85 & \AppendixShade{0.80}{0.80} \\
         & ProtoGCD & DINO & 66.97 & 69.07 & \AppendixShade{-2.10}{-2.10} & 45.21 & 43.59 & \AppendixShade{1.62}{1.62} & 52.29 & 51.89 & \AppendixShade{0.40}{0.40} \\
         & SimGCD & DINO & 72.01 & 74.11 & \AppendixShade{-2.10}{-2.10} & 44.61 & 42.70 & \AppendixShade{1.91}{1.91} & 53.53 & 52.93 & \AppendixShade{0.60}{0.60} \\
         & SimGCD & CLIP & 83.71 & 89.01 & \AppendixShade{-5.30}{-5.30} & 64.06 & 59.38 & \AppendixShade{4.68}{4.68} & 70.46 & 69.03 & \AppendixShade{1.43}{1.43} \\
         & GET-TAR & CLIP & 82.51 & 95.75 & \AppendixShade{-13.24}{-13.24} & 72.80 & 60.51 & \AppendixShade{12.29}{12.29} & 75.96 & 71.99 & \AppendixShade{3.97}{3.97} \\
         & GET & CLIP & 88.01 & 92.30 & \AppendixShade{-4.29}{-4.29} & 74.58 & 70.75 & \AppendixShade{3.83}{3.83} & 78.96 & 77.77 & \AppendixShade{1.19}{1.19} \\
        \cmidrule(lr){2-12}
        \multirow{7}{*}{CUB} & LegoGCD & DINO & 71.18 & 73.05 & \AppendixShade{-1.87}{-1.87} & 57.69 & 55.72 & \AppendixShade{1.97}{1.97} & 62.19 & 61.50 & \AppendixShade{0.69}{0.69} \\
         & SimGCD-BIA & DINO & 60.84 & 60.91 & \AppendixShade{-0.07}{-0.07} & 62.00 & 60.89 & \AppendixShade{1.11}{1.11} & 61.61 & 60.90 & \AppendixShade{0.71}{0.71} \\
         & ProtoGCD & DINO & 64.64 & 64.78 & \AppendixShade{-0.14}{-0.14} & 59.13 & 58.09 & \AppendixShade{1.04}{1.04} & 60.97 & 60.32 & \AppendixShade{0.65}{0.65} \\
         & SimGCD & DINO & 66.64 & 65.04 & \AppendixShade{1.60}{1.60} & 58.09 & 58.39 & \AppendixShade{-0.30}{-0.30} & 60.94 & 60.61 & \AppendixShade{0.33}{0.33} \\
         & SimGCD & CLIP & 75.38 & 76.12 & \AppendixShade{-0.74}{-0.74} & 65.10 & 64.40 & \AppendixShade{0.70}{0.70} & 68.53 & 68.31 & \AppendixShade{0.22}{0.22} \\
         & GET-TAR & CLIP & 77.79 & 78.79 & \AppendixShade{-1.00}{-1.00} & 73.01 & 72.07 & \AppendixShade{0.94}{0.94} & 74.60 & 74.31 & \AppendixShade{0.29}{0.29} \\
         & GET & CLIP & 76.45 & 76.85 & \AppendixShade{-0.40}{-0.40} & 74.97 & 74.74 & \AppendixShade{0.23}{0.23} & 75.47 & 75.44 & \AppendixShade{0.03}{0.03} \\
        \cmidrule(lr){2-12}
        \multirow{7}{*}{FGVC-Aircraft} & LegoGCD & DINO & 62.12 & 64.71 & \AppendixShade{-2.59}{-2.59} & 50.52 & 47.53 & \AppendixShade{2.99}{2.99} & 54.39 & 53.25 & \AppendixShade{1.14}{1.14} \\
         & SimGCD-BIA & DINO & 55.46 & 56.30 & \AppendixShade{-0.84}{-0.84} & 50.40 & 49.84 & \AppendixShade{0.56}{0.56} & 52.09 & 51.99 & \AppendixShade{0.10}{0.10} \\
         & ProtoGCD & DINO & 46.40 & 46.04 & \AppendixShade{0.36}{0.36} & 54.60 & 54.42 & \AppendixShade{0.18}{0.18} & 51.87 & 51.63 & \AppendixShade{0.24}{0.24} \\
         & SimGCD & DINO & 58.58 & 60.14 & \AppendixShade{-1.56}{-1.56} & 50.67 & 49.66 & \AppendixShade{1.01}{1.01} & 53.31 & 53.15 & \AppendixShade{0.16}{0.16} \\
         & SimGCD & CLIP & 59.54 & 59.54 & \AppendixShade{0.00}{0.00} & 53.61 & 53.43 & \AppendixShade{0.18}{0.18} & 55.59 & 55.47 & \AppendixShade{0.12}{0.12} \\
         & GET-TAR & CLIP & 64.59 & 66.39 & \AppendixShade{-1.80}{-1.80} & 58.26 & 57.09 & \AppendixShade{1.17}{1.17} & 60.37 & 60.19 & \AppendixShade{0.18}{0.18} \\
         & GET & CLIP & 64.59 & 67.65 & \AppendixShade{-3.06}{-3.06} & 58.17 & 55.71 & \AppendixShade{2.46}{2.46} & 60.31 & 59.69 & \AppendixShade{0.62}{0.62} \\
        \bottomrule
    \end{tabular}}
\end{table}

\textbf{Observations.}
On the coarse-grained benchmarks, GH and SPE give the same scores for almost all available runs. The only visible difference is SimGCD--CLIP on CIFAR-100, where the changes remain small. This indicates that the two protocols behave similarly when old and new categories are well separated. The fine-grained benchmarks are more sensitive to the evaluation protocol. On Herbarium 19, GH increases the reported new-category and overall scores for every method, while SPE gives higher old-category accuracy in most cases. The largest differences occur for SimGCD--CLIP and GET--CLIP. 

\textbf{Calculation of the cycle diagnostics.}
\phantomsection\label{app:cycle_audit}
Tab.\ref{tab:cycle_validation} compares GH and SPE on the \emph{same predictions}; only the output-to-category matching changes. We use the confusion counts $C_{y\rightarrow\hat y}$ and assignments $\mathcal A_0$ and $\mathcal A_{\mathrm{GH}}$ introduced in the main text to compute the cycle diagnostics as follows.

\noindent\textit{From changed assignments to cycles.}
Starting from each category $y$, we follow $y\mapsto\mathcal A_{\mathrm{GH}}(\mathcal A_0^{-1}(y))$ until the sequence returns to its starting category. Repeating this procedure for unvisited categories constructs $\Gamma(\mathcal A_{\mathrm{GH}},\mathcal A_0)$. We omit unchanged categories but include nontrivial cycles whose margins are zero. For each cycle, we compute $m_\gamma$ using its definition in Thm.\ref{thm:semantic_mismapping_margin} and sum them to obtain the total reported in Tab.\ref{tab:cycle_validation}. For brevity in the numerical examples below, write this sum as
\[
M:=\sum_{\gamma\in\Gamma(\mathcal A_{\mathrm{GH}},\mathcal A_0)}m_\gamma.
\]
We then obtain $\Delta_{\mathrm{all}}$ from Eq.\eqref{eq:cycle_gap_identity}.

\noindent\textit{Examples using the counts in Tab.\ref{tab:cycle_validation}.}
On CIFAR-10--DINO, no assignment changes, so $M=0$ and $\Delta_{\mathrm{all}}=0$. On CUB--CLIP, the four cycle margins are $0,-8,-2,0$. Therefore,
\[
M=0-8-2+0=-10,\qquad
\Delta_{\mathrm{all}}=\frac{100\times10}{4{,}496}=0.22.
\]
On Stanford Cars--CLIP, one cycle gives $M=-88$ and $\Delta_{\mathrm{all}}=(100\times88)/6{,}144=1.43$. On Herbarium 19--CLIP, three cycles give
\[
M=-1{,}391-1{,}644-41=-3{,}076,\qquad
\Delta_{\mathrm{all}}=\frac{100\times3{,}076}{25{,}356}=12.13.
\]

\noindent\textit{What the Type~\textnormal{(I)} and Type~\textnormal{(II)} columns count.}
For each anchored old output, we check its GH-assigned category. Type~\textnormal{(I)} counts outputs reassigned to a \emph{different old} category; Type~\textnormal{(II)} counts outputs reassigned to a \emph{new} category. These are counts of output indices, not samples or cycles. For example, Herbarium 19 has $4$ Type~\textnormal{(I)} and $119$ Type~\textnormal{(II)} outputs, for $4+119=123$ reassigned old outputs.

\FloatBarrier
\subsection{Additional Results on the Coarse-Grained Benchmark (RQ2)}
\label{app:coarse_grained_ours}

We evaluate the effectiveness of our method on the coarse-grained ImageNet-100 benchmark. The GH and SPE scores are identical for every reported baseline and +Ours run, indicating no observable GH-induced overestimation in these experiments. Nevertheless, MMO improves performance across all five configurations, with gains of $0.15$--$3.05$ points in overall accuracy and $0.02$--$3.30$ points in new accuracy. Thus, the method also generalizes to a coarse-grained setting where GH-induced overestimation is not observed. 
\begin{table}[H]
    \centering
    \footnotesize
    \caption{Baseline and \textbf{+Ours} on ImageNet-100 under GH and SPE. Orange New/$H$ cells indicate gains from the baseline.}
    \label{tab:appendix_imagenet100_ours}
    \setlength{\tabcolsep}{2.5pt}
    \renewcommand{\arraystretch}{1.08}
    \fitTable{%
    \begin{tabular}{@{}lllrrrrrrrr@{}}
        \toprule
        \multirow{2}{*}{Dataset} & \multirow{2}{*}{Method} & \multirow{2}{*}{Backbone}
        & \multicolumn{4}{c}{GH}
        & \multicolumn{4}{c}{SPE} \\
        \cmidrule(lr){4-7}\cmidrule(lr){8-11}
        & & & $\texttt{All}_{\mathrm{GH}}$ & $\texttt{Old}_{\mathrm{GH}}$ & $\texttt{New}_{\mathrm{GH}}$ & $\texttt{H}_{\mathrm{GH}}$
        & $\texttt{All}_{\mathrm{SPE}}$ & $\texttt{Old}_{\mathrm{SPE}}$ & $\texttt{New}_{\mathrm{SPE}}$ & $\texttt{H}_{\mathrm{SPE}}$ \\
        \midrule
        & LegoGCD & DINO & 83.49 & \textbf{94.88} & 77.76 & 85.47 & 83.49 & \textbf{94.88} & 77.76 & 85.47 \\
        \rowcolor{black!2}\cellcolor{white} & \quad\textbf{+ Ours} & DINO & \textbf{84.89} & 94.87 & \AppendixShade{\textbf{79.88}}{2.12} & \AppendixShade{\textbf{86.73}}{1.26} & \textbf{84.89} & 94.87 & \AppendixShade{\textbf{79.88}}{2.12} & \AppendixShade{\textbf{86.73}}{1.26} \\
        & SimGCD-BIA & DINO & 81.16 & 86.05 & 78.70 & 82.21 & 81.16 & 86.05 & 78.70 & 82.21 \\
        \rowcolor{black!2}\cellcolor{white} & \quad\textbf{+ Ours} & DINO & \textbf{84.21} & \textbf{92.11} & \AppendixShade{\textbf{80.24}}{1.54} & \AppendixShade{\textbf{85.77}}{3.56} & \textbf{84.21} & \textbf{92.11} & \AppendixShade{\textbf{80.24}}{1.54} & \AppendixShade{\textbf{85.77}}{3.56} \\
        & ProtoGCD & DINO & 82.59 & 91.00 & 78.36 & 84.21 & 82.59 & 91.00 & 78.36 & 84.21 \\
        \rowcolor{black!2}\cellcolor{white} & \quad\textbf{+ Ours} & DINO & \textbf{82.83} & \textbf{91.68} & \AppendixShade{\textbf{78.38}}{0.02} & \AppendixShade{\textbf{84.51}}{0.30} & \textbf{82.83} & \textbf{91.68} & \AppendixShade{\textbf{78.38}}{0.02} & \AppendixShade{\textbf{84.51}}{0.30} \\
        & SimGCD & DINO & 85.04 & 92.58 & 81.26 & 86.55 & 85.04 & 92.58 & 81.26 & 86.55 \\
        \rowcolor{black!2}\cellcolor{white} & \quad\textbf{+ Ours} & DINO & \textbf{85.19} & \textbf{92.89} & \AppendixShade{\textbf{81.32}}{0.06} & \AppendixShade{\textbf{86.72}}{0.17} & \textbf{85.19} & \textbf{92.89} & \AppendixShade{\textbf{81.32}}{0.06} & \AppendixShade{\textbf{86.72}}{0.17} \\
        & SimGCD & CLIP & 90.13 & 94.76 & 87.80 & 91.15 & 90.13 & 94.76 & 87.80 & 91.15 \\
        \rowcolor{black!2}\cellcolor{white}\multirow{-10}{*}{ImageNet-100} & \quad\textbf{+ Ours} & CLIP & \textbf{92.51} & \textbf{95.31} & \AppendixShade{\textbf{91.10}}{3.30} & \AppendixShade{\textbf{93.16}}{2.01} & \textbf{92.51} & \textbf{95.31} & \AppendixShade{\textbf{91.10}}{3.30} & \AppendixShade{\textbf{93.16}}{2.01} \\
        \bottomrule
    \end{tabular}}
\end{table}

\subsection{Complete Results on the Large-Scale Benchmark (RQ1 and RQ2)}
\label{app:large_scale_details}

Tab.\ref{tab:appendix_imagenet1k_protocol} compares the three reproduced ImageNet-1K baselines. Following Tab.\ref{tab:appendix_gh_spe_complete}, every difference is defined as $\Delta_r=\texttt{Acc}_{\mathrm{GH}}^r-\texttt{Acc}_{\mathrm{SPE}}^r$. All three baselines exhibit the same protocol pattern. Compared with SPE, GH increases new-category accuracy by $5.19$--$7.36$ points and overall accuracy by $1.41$--$2.23$ points, while decreasing the reported old-category accuracy by $6.12$--$8.20$ points. Thus, the GH--SPE discrepancy persists on the large-scale benchmark and is not tied to a single backbone.

\begin{table}[H]
    \centering
    \footnotesize
    \caption{Detailed GH--SPE comparison of reproduced baselines on ImageNet-1K. GH--SPE differences are orange (positive) or blue (negative/zero).}
    \label{tab:appendix_imagenet1k_protocol}
    \setlength{\tabcolsep}{2.1pt}
    \renewcommand{\arraystretch}{1.04}
    \fitTable{%
    \begin{tabular}{@{}llrrrrrrrrr@{}}
        \toprule
        \multirow{2}{*}{Method} & \multirow{2}{*}{Backbone}
        & \multicolumn{3}{c}{Old categories}
        & \multicolumn{3}{c}{New categories}
        & \multicolumn{3}{c}{All categories} \\
        \cmidrule(lr){3-5}\cmidrule(lr){6-8}\cmidrule(lr){9-11}
        & & $\texttt{Old}_{\mathrm{GH}}$ & $\texttt{Old}_{\mathrm{SPE}}$ & $\Delta_{\mathrm{old}}$
        & $\texttt{New}_{\mathrm{GH}}$ & $\texttt{New}_{\mathrm{SPE}}$ & $\Delta_{\mathrm{new}}$
        & $\texttt{All}_{\mathrm{GH}}$ & $\texttt{All}_{\mathrm{SPE}}$ & $\Delta_{\mathrm{all}}$ \\
        \midrule
        SimGCD & DINO & 77.14 & 85.15 & \AppendixShade{-8.01}{-8.01} & 46.94 & 39.58 & \AppendixShade{7.36}{7.36} & 57.04 & 54.81 & \AppendixShade{2.23}{2.23} \\
        LegoGCD & DINO & 79.18 & 85.30 & \AppendixShade{-6.12}{-6.12} & 54.88 & 49.69 & \AppendixShade{5.19}{5.19} & 63.01 & 61.60 & \AppendixShade{1.41}{1.41} \\
        SimGCD & CLIP & 77.06 & 85.26 & \AppendixShade{-8.20}{-8.20} & 48.41 & 41.27 & \AppendixShade{7.14}{7.14} & 57.99 & 55.98 & \AppendixShade{2.01}{2.01} \\
        \bottomrule
    \end{tabular}}
\end{table}

\begin{table}[H]
    \centering
    \footnotesize
    \caption{ImageNet-1K comparison of SimGCD and \textbf{+Ours}. Orange New cells indicate gains from the baseline; higher scores are in \textbf{bold}.}
    \label{tab:appendix_imagenet1k_ours}
    \setlength{\tabcolsep}{2.1pt}
    \renewcommand{\arraystretch}{1.06}
    \fitTable{%
    \begin{tabular}{@{}llrrrrrr@{}}
        \toprule
        Method & Backbone
        & $\texttt{All}_{\mathrm{GH}}$ & $\texttt{Old}_{\mathrm{GH}}$ & $\texttt{New}_{\mathrm{GH}}$
        & $\texttt{All}_{\mathrm{SPE}}$ & $\texttt{Old}_{\mathrm{SPE}}$ & $\texttt{New}_{\mathrm{SPE}}$ \\
        \midrule
        SimGCD & DINO & 57.04 & 77.14 & 46.94 & 54.81 & 85.15 & 39.58 \\
        \rowcolor{black!6}\quad\textbf{+ Ours} & DINO & \textbf{58.69} & \textbf{77.29} & \AppendixShade{\textbf{49.34}}{2.40} & \textbf{56.88} & \textbf{85.28} & \AppendixShade{\textbf{42.60}}{3.02} \\
        \cmidrule(lr){1-8}
        SimGCD & CLIP & 57.99 & 77.06 & 48.41 & 55.98 & \textbf{85.26} & 41.27 \\
        \rowcolor{black!6}\quad\textbf{+ Ours} & CLIP & \textbf{59.03} & \textbf{77.63} & \AppendixShade{\textbf{49.68}}{1.27} & \textbf{57.35} & 85.10 & \AppendixShade{\textbf{43.40}}{2.13} \\
        \bottomrule
    \end{tabular}}
\end{table}

Under both GH and SPE, +Ours improves the overall and new-category scores for both DINO and CLIP. The gains in $\texttt{New}$ are $2.40/3.02$ points for DINO and $1.27/2.13$ points for CLIP under GH/SPE, respectively. The corresponding $\texttt{All}$ gains are $1.65/2.07$ and $1.04/1.37$ points. These consistent improvements demonstrate that MMO remains effective when extended to the large-scale benchmark.

\subsection{Complete \texorpdfstring{$\rho$}{rho}-Sensitivity Results on an Imbalanced Benchmark (RQ2)}
\label{app:rho_sensitivity_full}

Herbarium 19 is a naturally long-tailed, class-imbalanced benchmark. Its nearly equal numbers of old and new categories do not imply equal sample frequencies, so the old-region sample prior in the unlabeled set cannot be recovered from category counts alone. We therefore require $\rho$ (or $\widehat\rho$) to infer $p_{\mathrm{old}}$ through Eq.\eqref{eq:pao_region_prior}. Tab.\ref{tab:appendix_rho_sensitivity_full} reports the complete GH and SPE metrics for the runs with manually specified $\widehat\rho$ and the TIcE estimate, which is described in Sec.\ref{app:rbl_prior_estimation}. 

\begin{table}[H]
    \centering
    \footnotesize
    \caption{Complete sensitivity results on the imbalanced Herbarium 19 benchmark with SimGCD--CLIP. The true labeling fraction is $\rho=0.50$. TIcE estimates $\widehat\rho$ from the training data to infer the required sample-level prior $p_{\mathrm{old}}$. The best result in each metric column is in \textbf{bold}.}
    \label{tab:appendix_rho_sensitivity_full}
    \setlength{\tabcolsep}{2.0pt}
    \renewcommand{\arraystretch}{1.06}
    \fitTable{%
    \begin{tabular}{@{}lcccrrrrrrrr@{}}
        \toprule
        \multirow{2}{*}{Variant} & \multirow{2}{*}{$\rho$} & \multirow{2}{*}{$\widehat\rho$} & \multirow{2}{*}{$\widehat p_{\mathrm{old}}$}
        & \multicolumn{4}{c}{GH}
        & \multicolumn{4}{c}{SPE} \\
        \cmidrule(lr){5-8}\cmidrule(lr){9-12}
        & & & & $\texttt{All}_{\mathrm{GH}}$ & $\texttt{Old}_{\mathrm{GH}}$ & $\texttt{New}_{\mathrm{GH}}$
        & $\texttt{H}_{\mathrm{GH}}$ & $\texttt{All}_{\mathrm{SPE}}$ & $\texttt{Old}_{\mathrm{SPE}}$ & $\texttt{New}_{\mathrm{SPE}}$ & $\texttt{H}_{\mathrm{SPE}}$ \\
        \midrule
        Baseline & 0.50 & -- & -- & 47.40 & 64.39 & 38.26 & 48.00 & 35.27 & \textbf{77.34} & 12.63 & 21.71 \\
        \midrule
        Manually specified & 0.50 & 0.40 & 0.5247 & \textbf{51.83} & 64.31 & \textbf{45.12} & \textbf{53.03} & \textbf{50.30} & 67.50 & 41.04 & \textbf{51.04} \\
        Manually specified & 0.50 & 0.50 & 0.3498 & 49.14 & 56.91 & 44.96 & 50.23 & 48.58 & 57.90 & 43.56 & 49.72 \\
        Manually specified & 0.50 & 0.60 & 0.2332 & 46.28 & 48.47 & 45.11 & 46.73 & 46.00 & 47.50 & 45.20 & 46.32 \\
        Manually specified & 0.50 & 0.70 & 0.1499 & 43.26 & 40.06 & 44.99 & 42.38 & 42.76 & 37.89 & \textbf{45.37} & 41.29 \\
        \midrule
        TIcE estimate & 0.50 & 0.3575 & 0.6287 & 50.34 & \textbf{64.64} & 42.64 & 51.38 & 46.02 & 70.74 & 32.72 & 44.74 \\
        \bottomrule
    \end{tabular}}
\end{table}

On Herbarium 19, increasing $\widehat\rho$ from $0.40$ to $0.70$ raises $\texttt{New}_{\mathrm{SPE}}$ from $41.04$ to $45.37$ but lowers $\texttt{All}_{\mathrm{SPE}}$ from $50.30$ to $42.76$. A modest deviation from the true $\rho=0.50$ can even improve overall and harmonic-mean accuracy. At $\widehat\rho=0.40$, the harmonic means reach $\texttt{H}_{\mathrm{GH}}=53.03$ and $\texttt{H}_{\mathrm{SPE}}=51.04$. The baseline's large GH--SPE gaps indicate severe semantic mismapping. With the TIcE estimate, MMO improves $\texttt{All}_{\mathrm{SPE}}$ and $\texttt{New}_{\mathrm{SPE}}$ by $10.75$ and $20.09$ points over the baseline and narrows the GH--SPE gaps. Thus, MMO mitigates prediction errors hidden by semantic mismapping and remains competitive when $\rho$ is estimated automatically in a more practical setting.

\fi

\subsection{Old--New Trade-offs and Overall Accuracy (RQ2)}
\label{app:clip_tradeoffs}
We visualize the CLIP-based SimGCD, GET-TAR (TAR), and GET results and their +Ours variants from Tab.\ref{tab:main_results} on CUB, FGVC-Aircraft, Herbarium 19, and Stanford Cars. Figs.\ref{fig:app_old_new_gh} and~\ref{fig:app_old_new_spe} show old versus new-category accuracy, while Figs.\ref{fig:app_new_all_gh} and~\ref{fig:app_new_all_spe} show new-category versus overall accuracy.

\textbf{Old--new trade-off.}
The SPE plots show that MMO often trades a small decrease in old-category accuracy for a larger gain in new-category accuracy. For SimGCD on CUB, new-category accuracy rises by $6.10$ points while old-category accuracy falls by only $0.34$ points. The new-category gains reach $13.45$ points for GET-TAR on Stanford Cars and $30.93$ points for SimGCD on Herbarium 19. Together with the harmonic means in Tab.\ref{tab:main_results}, these results show that MMO generally improves the old--new balance.

\textbf{From new-category gains to overall gains.}
The New--All plots show that MMO consistently improves new-category accuracy and improves overall accuracy in most CLIP-based configurations. Under SPE, SimGCD on Stanford Cars gains $12.38$ points in $\texttt{New}_{\mathrm{SPE}}$ and $6.80$ points in $\texttt{All}_{\mathrm{SPE}}$. TAR on Herbarium 19 gains $28.34$ and $14.52$ points in the same metrics. These results show that better recognition of new categories can also improve overall performance.

\begin{figure}[!t]
\centering
\begin{minipage}{0.235\textwidth}
\centering
\includegraphics[width=\linewidth]{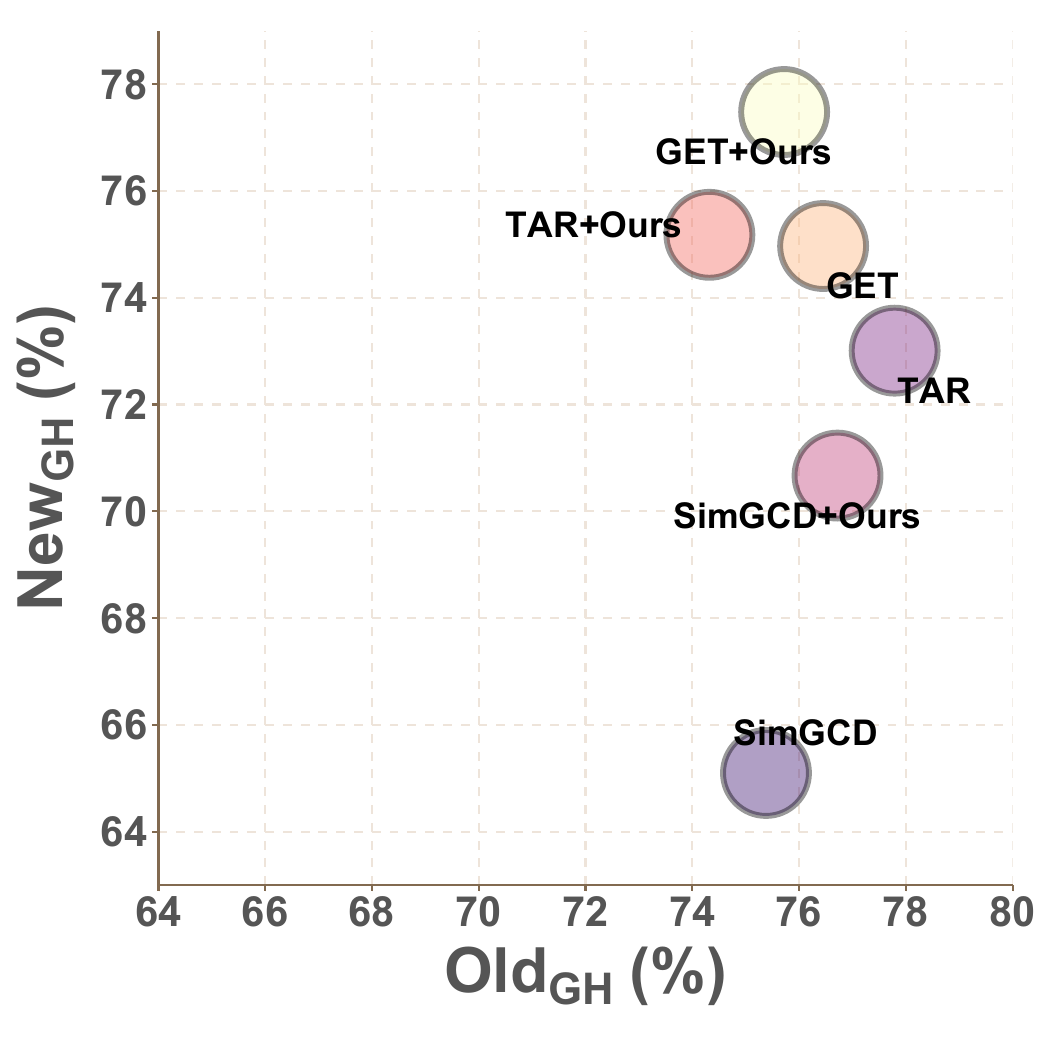}
{\small (a) CUB}
\end{minipage}\hfill
\begin{minipage}{0.235\textwidth}
\centering
\includegraphics[width=\linewidth]{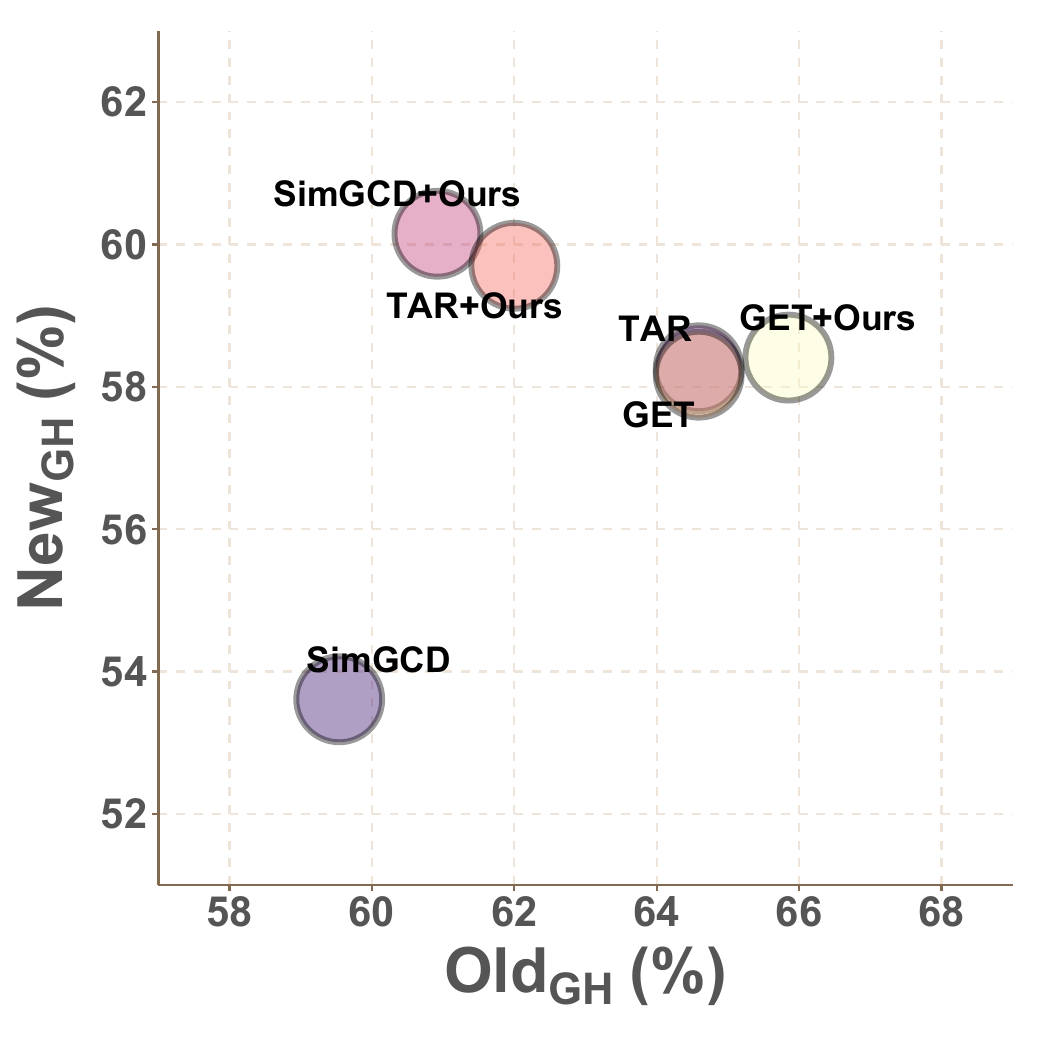}
{\small (b) Aircraft}
\end{minipage}\hfill
\begin{minipage}{0.235\textwidth}
\centering
\includegraphics[width=\linewidth]{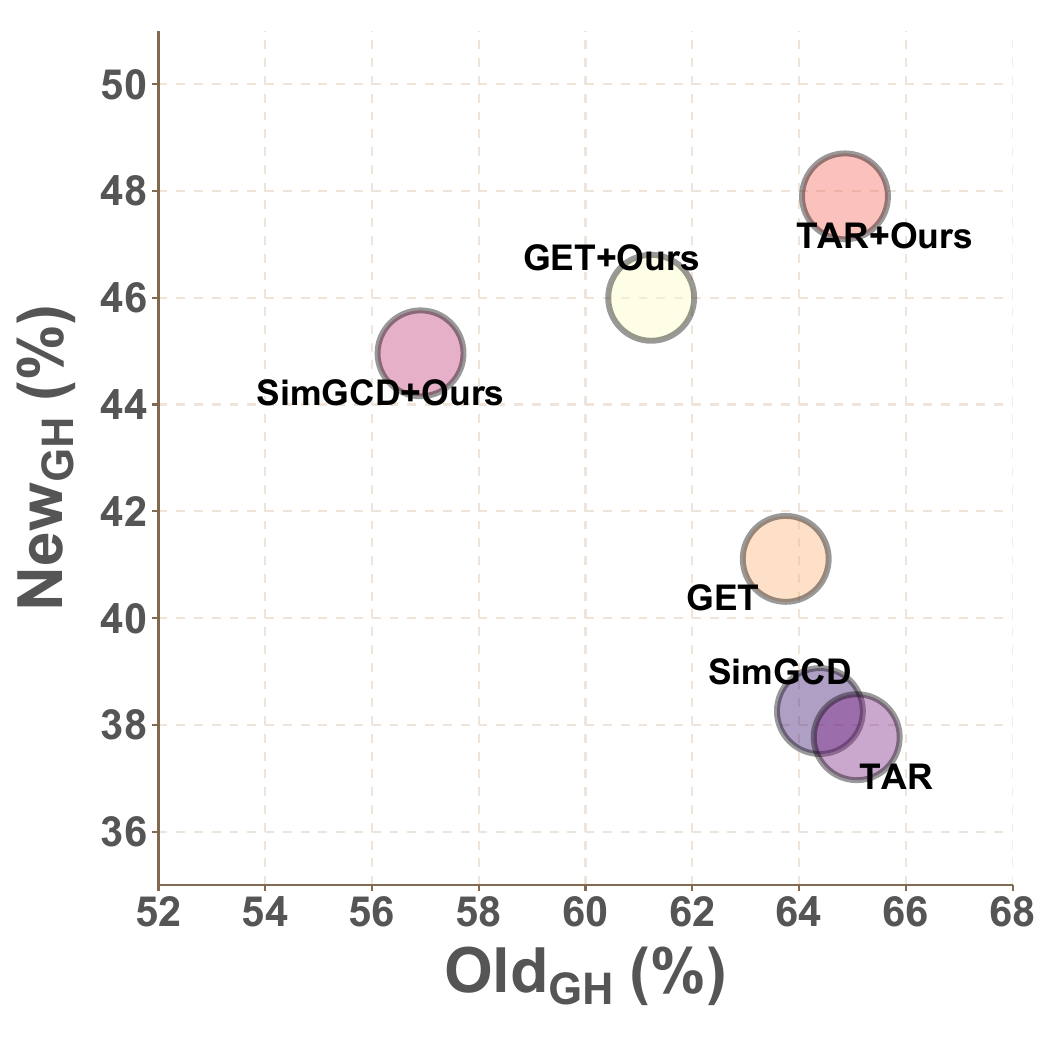}
{\small (c) H19}
\end{minipage}\hfill
\begin{minipage}{0.235\textwidth}
\centering
\includegraphics[width=\linewidth]{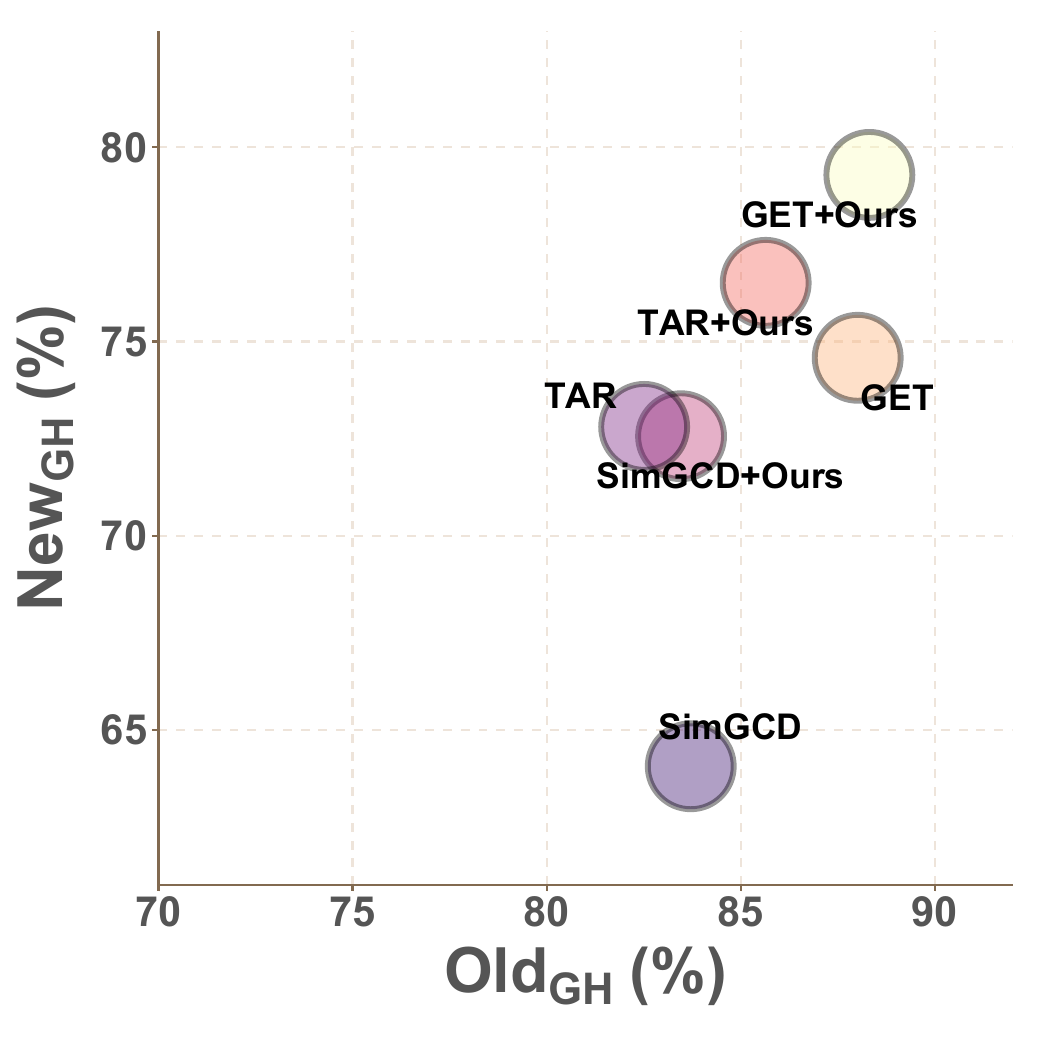}
{\small (d) Cars}
\end{minipage}
\caption{Old--new trade-offs under GH for CLIP-based methods. The axes show old-category (horizontal) and new-category (vertical) accuracy (\%). TAR denotes GET-TAR. }
\label{fig:app_old_new_gh}
\end{figure}
\begin{figure}[!t]
\centering
\begin{minipage}{0.235\textwidth}
\centering
\includegraphics[width=\linewidth]{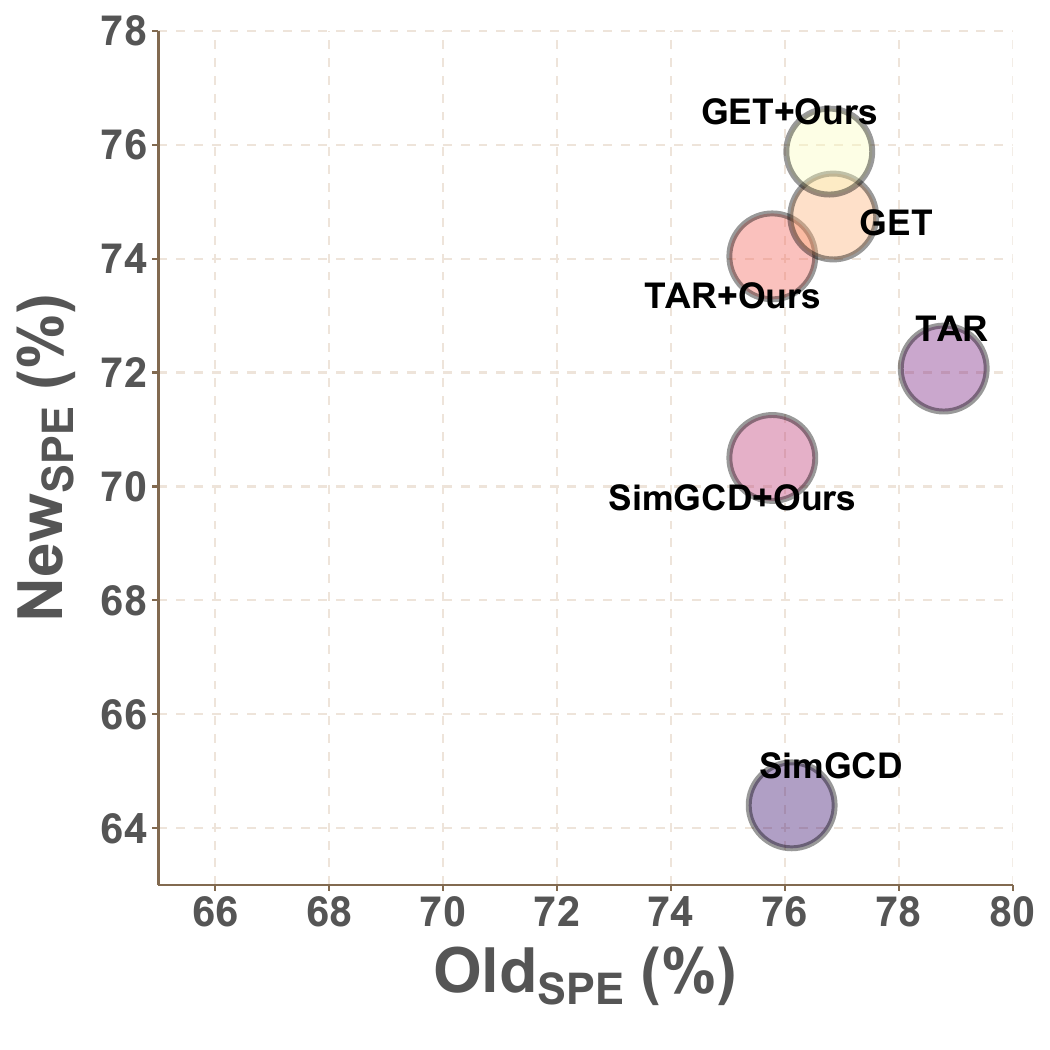}
{\small (a) CUB}
\end{minipage}\hfill
\begin{minipage}{0.235\textwidth}
\centering
\includegraphics[width=\linewidth]{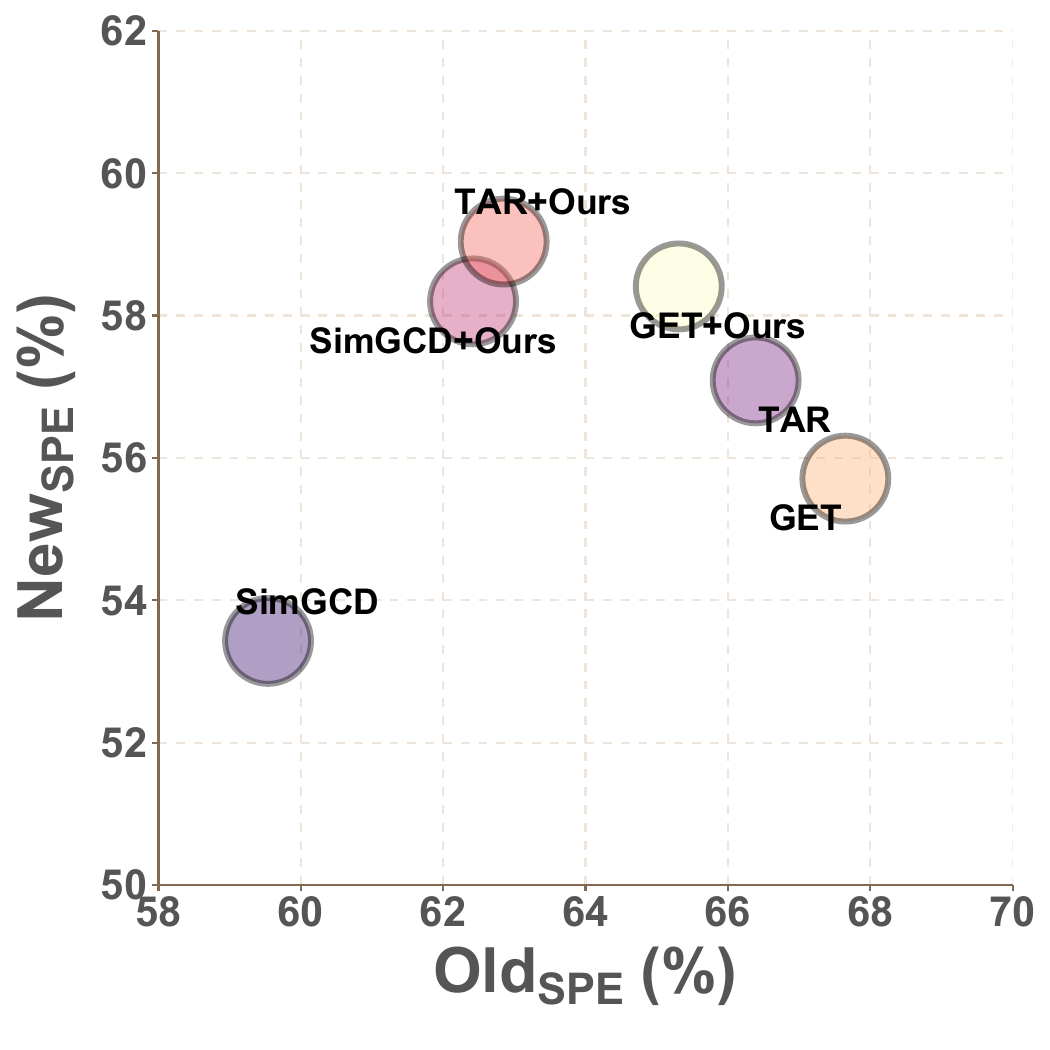}
{\small (b) Aircraft}
\end{minipage}\hfill
\begin{minipage}{0.235\textwidth}
\centering
\includegraphics[width=\linewidth]{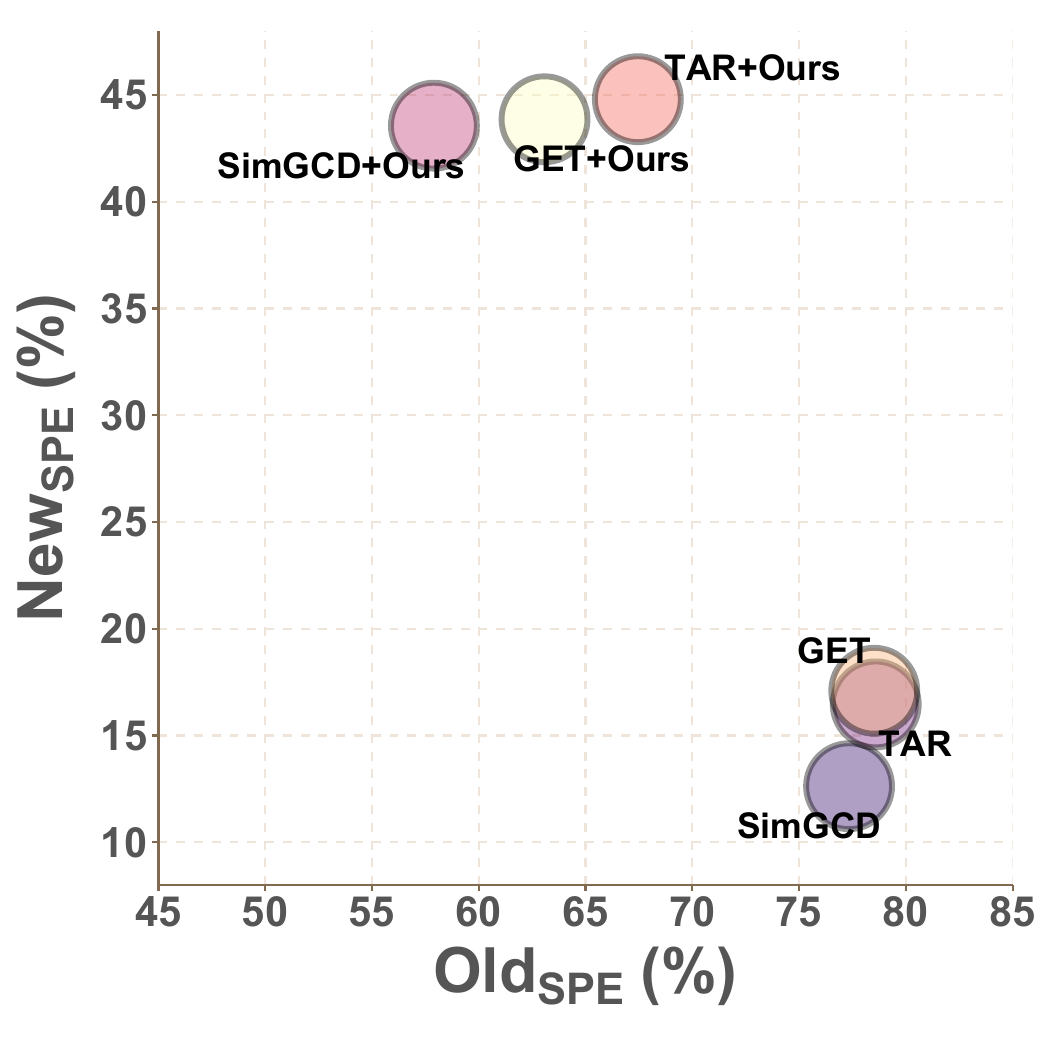}
{\small (c) H19}
\end{minipage}\hfill
\begin{minipage}{0.235\textwidth}
\centering
\includegraphics[width=\linewidth]{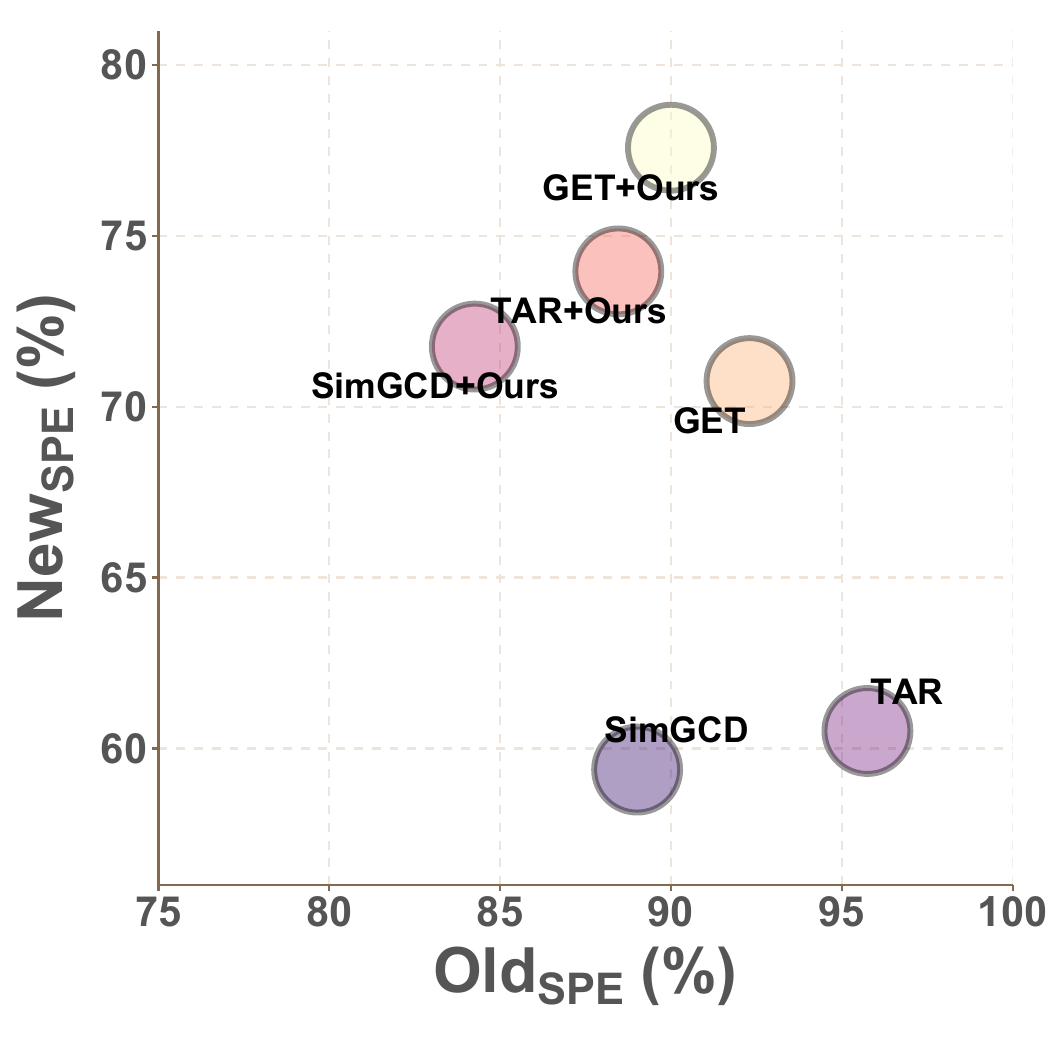}
{\small (d) Cars}
\end{minipage}
\caption{Old--new trade-offs under SPE for CLIP-based methods. The axes show old-category (horizontal) and new-category (vertical) accuracy (\%). TAR denotes GET-TAR. }
\label{fig:app_old_new_spe}
\end{figure}
\begin{figure}[!t]
\centering
\begin{minipage}{0.235\textwidth}
\centering
\includegraphics[width=\linewidth]{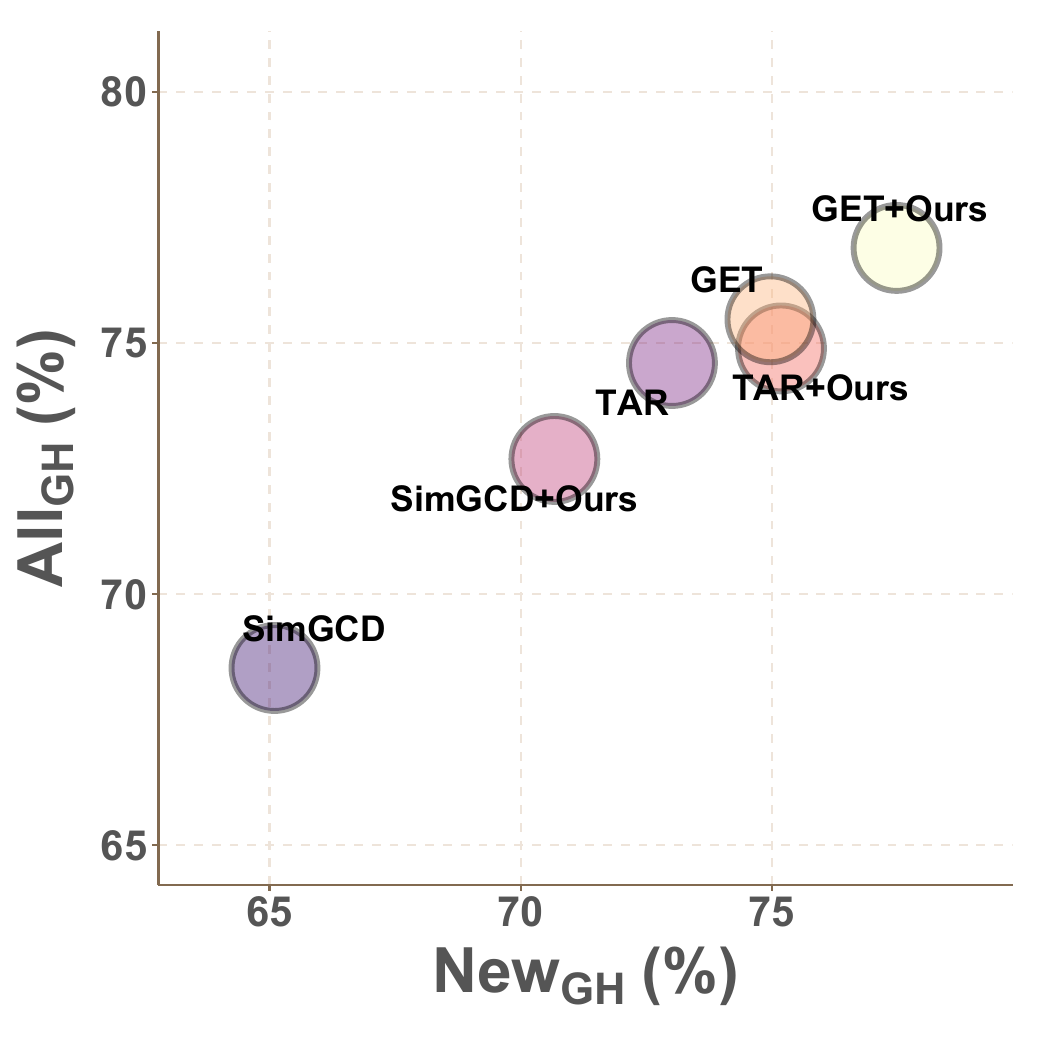}
{\small (a) CUB}
\end{minipage}\hfill
\begin{minipage}{0.235\textwidth}
\centering
\includegraphics[width=\linewidth]{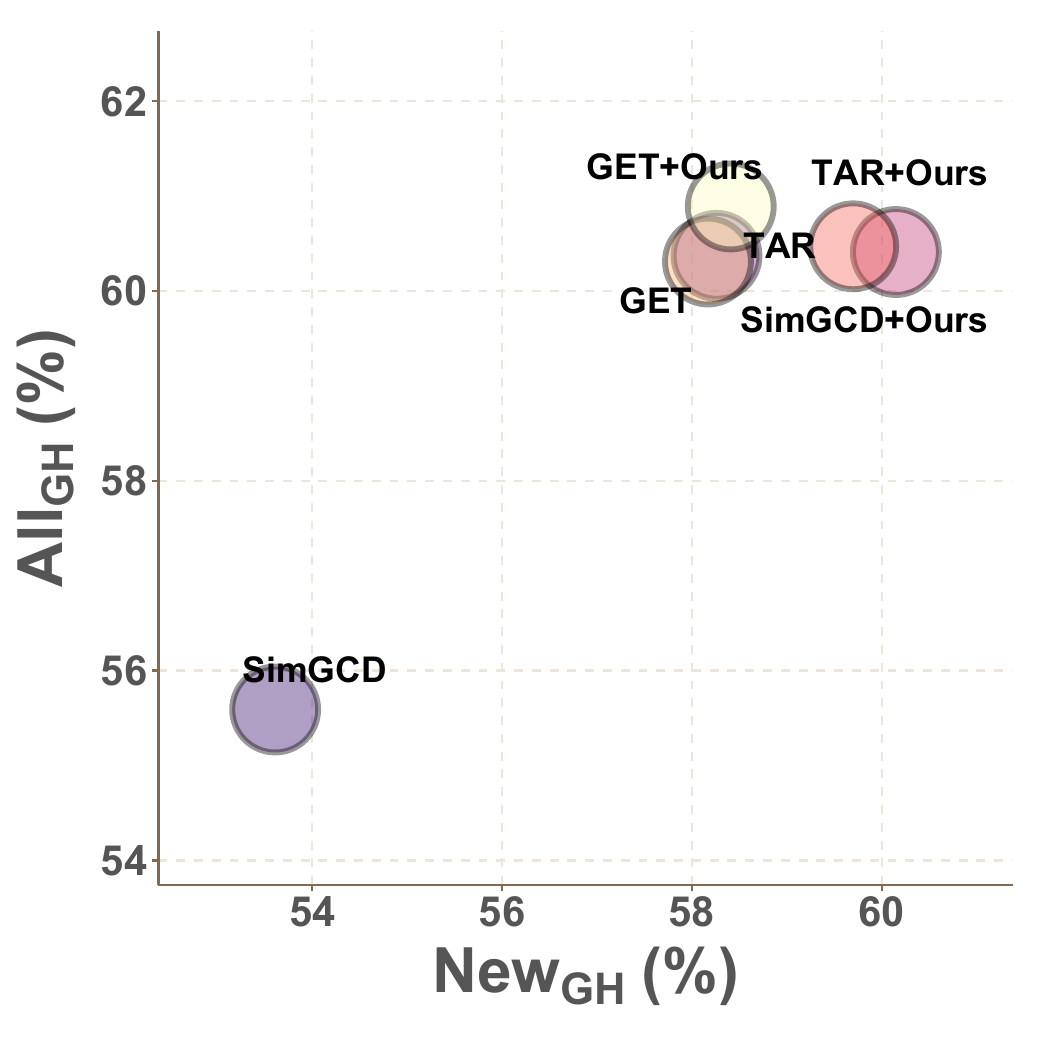}
{\small (b) Aircraft}
\end{minipage}\hfill
\begin{minipage}{0.235\textwidth}
\centering
\includegraphics[width=\linewidth]{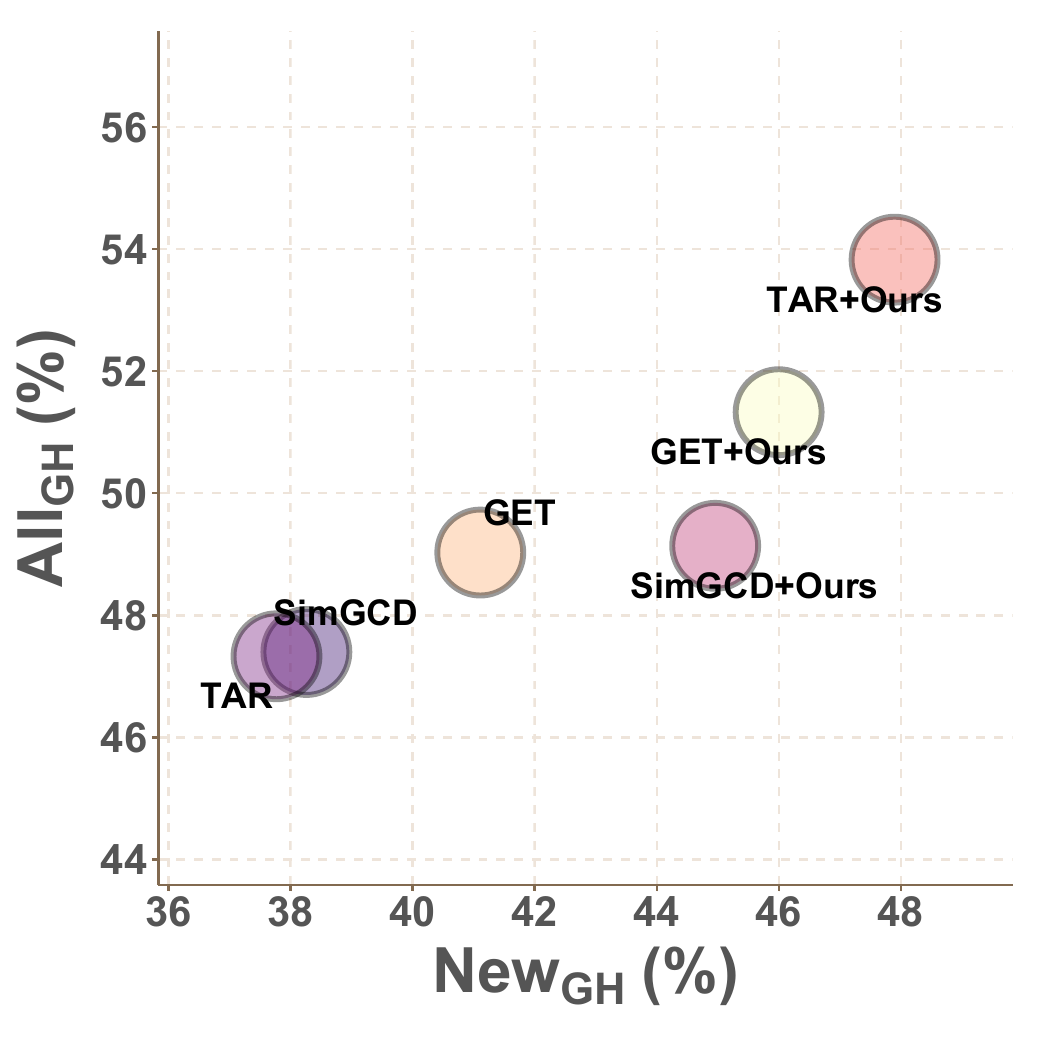}
{\small (c) H19}
\end{minipage}\hfill
\begin{minipage}{0.235\textwidth}
\centering
\includegraphics[width=\linewidth]{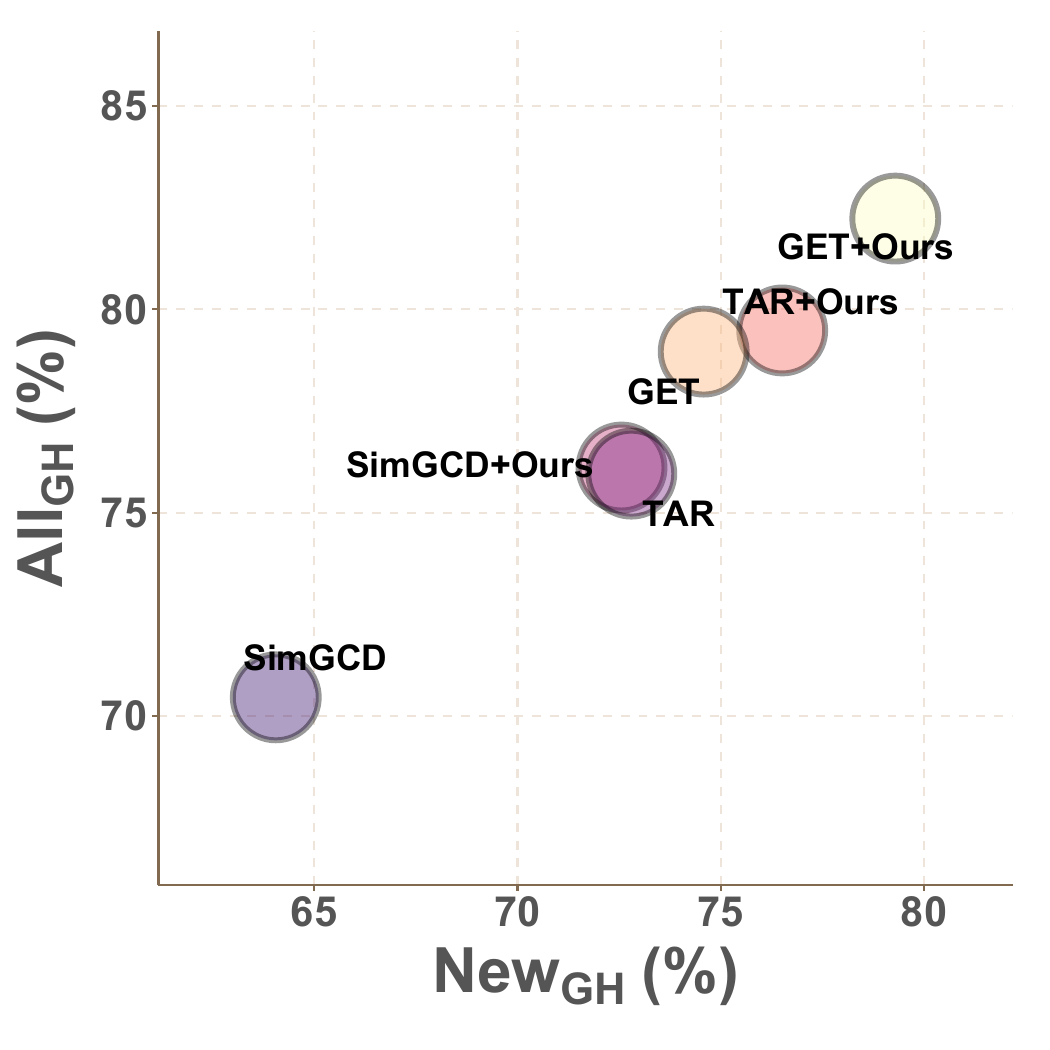}
{\small (d) Cars}
\end{minipage}
\caption{New-category versus overall accuracy under GH for CLIP-based methods. The axes show new-category (horizontal) and overall (vertical) accuracy (\%). TAR denotes GET-TAR. }
\label{fig:app_new_all_gh}
\end{figure}
\begin{figure}[!t]
\centering
\begin{minipage}{0.235\textwidth}
\centering
\includegraphics[width=\linewidth]{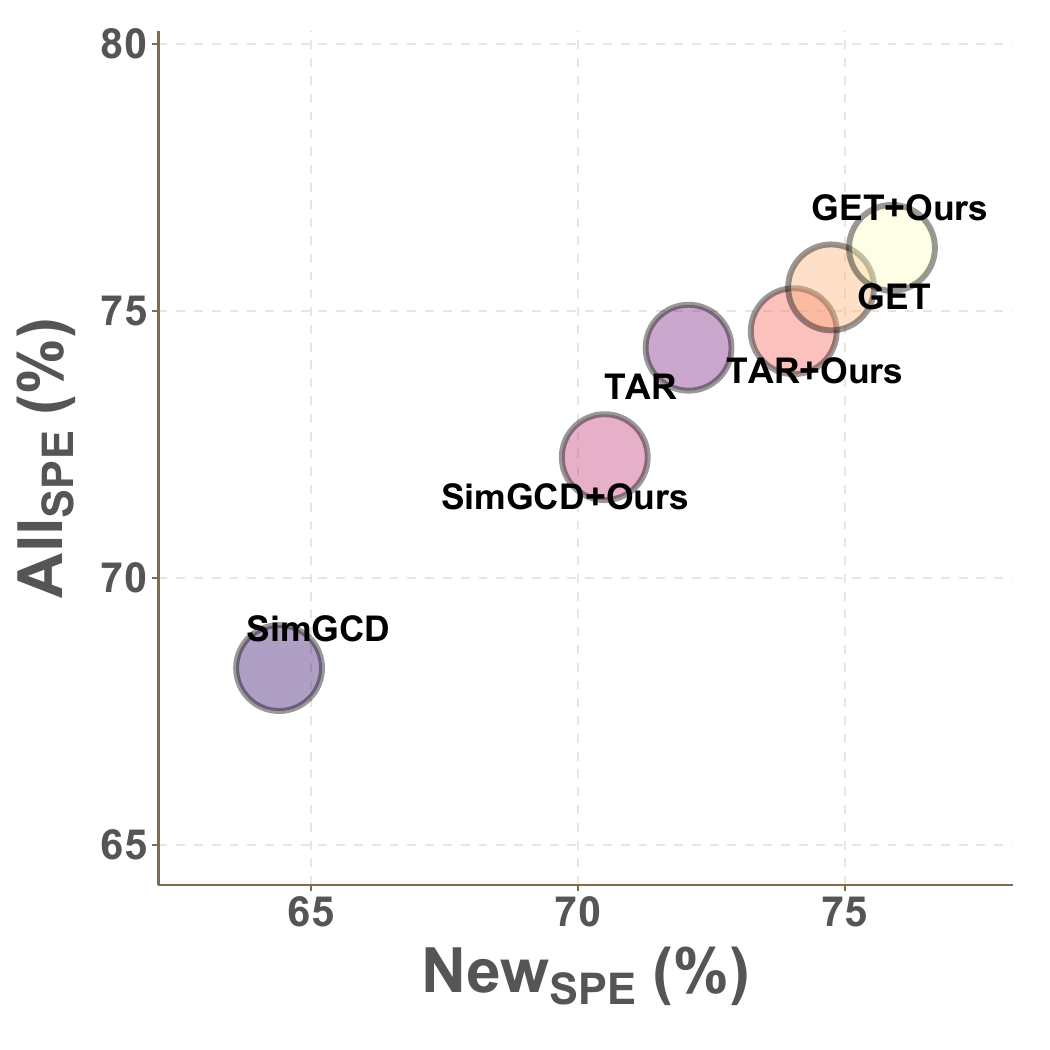}
{\small (a) CUB}
\end{minipage}\hfill
\begin{minipage}{0.235\textwidth}
\centering
\includegraphics[width=\linewidth]{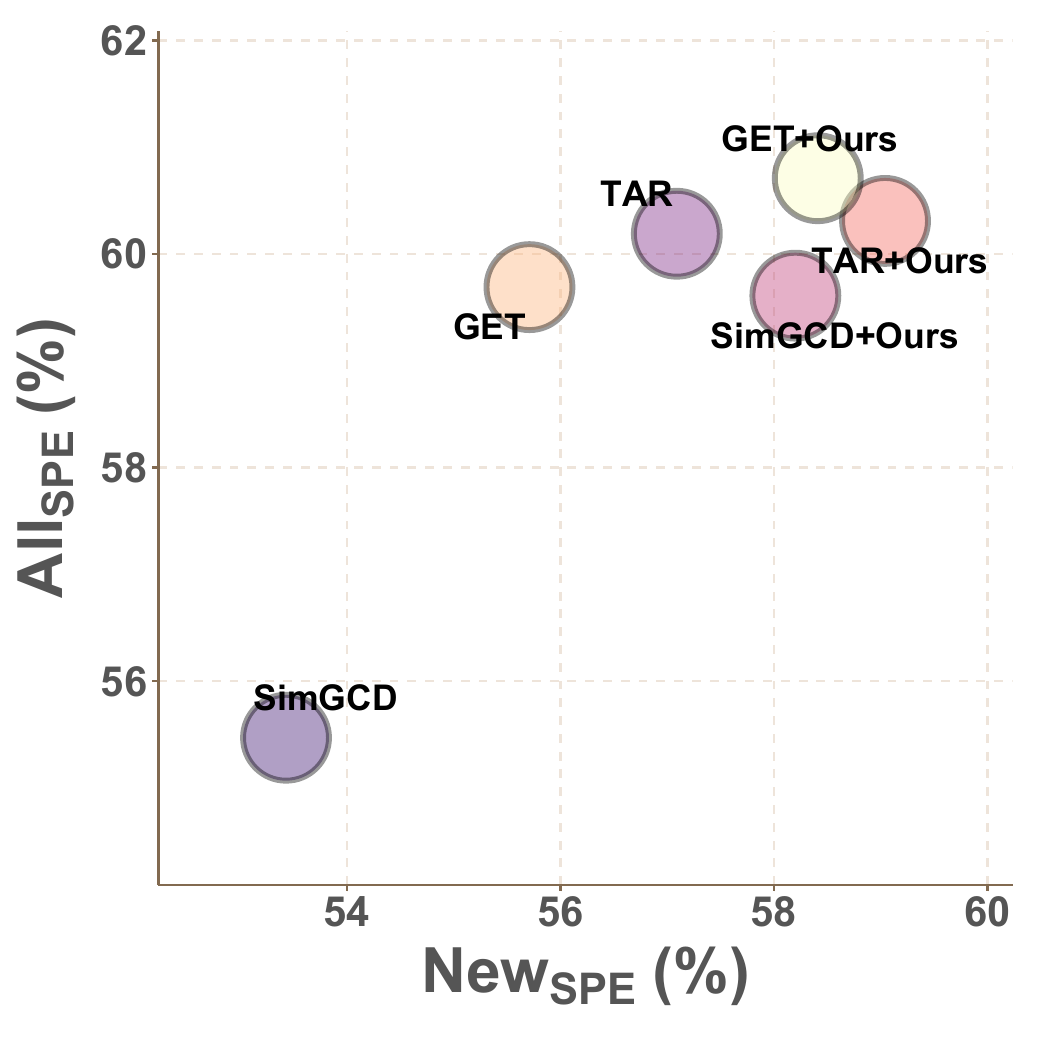}
{\small (b) Aircraft}
\end{minipage}\hfill
\begin{minipage}{0.235\textwidth}
\centering
\includegraphics[width=\linewidth]{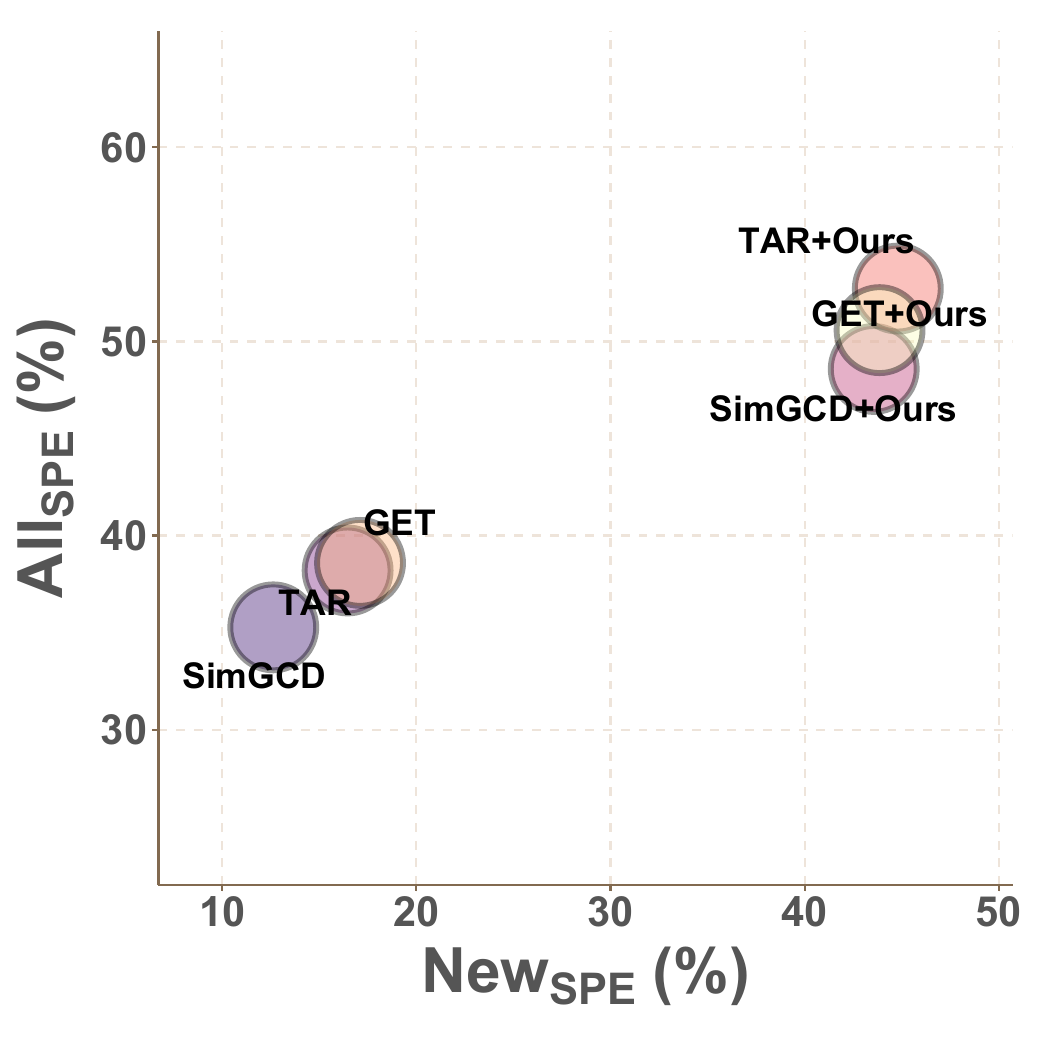}
{\small (c) H19}
\end{minipage}\hfill
\begin{minipage}{0.235\textwidth}
\centering
\includegraphics[width=\linewidth]{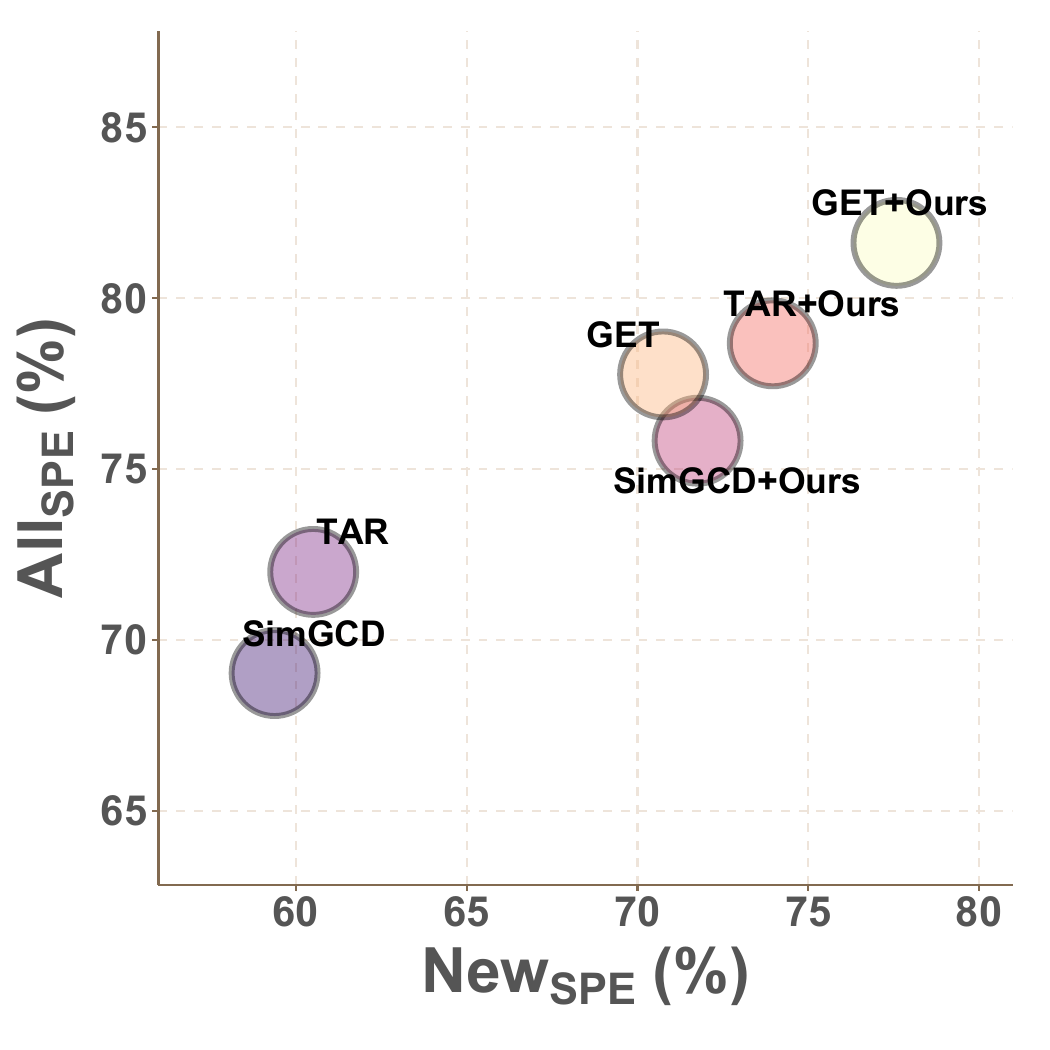}
{\small (d) Cars}
\end{minipage}
\caption{New-category versus overall accuracy under SPE for CLIP-based methods. The axes show new-category (horizontal) and overall (vertical) accuracy (\%). TAR denotes GET-TAR. }
\label{fig:app_new_all_spe}
\end{figure}
\subsection{Ablation of MMO Components (RQ2)}
\label{app:mmo_component_ablation}

Tab.\ref{tab:mmo_component_ablation} examines the effect of RBL and the additional contribution of NDL when combined with RBL. Compared with the baseline, RBL improves overall accuracy and the old--new harmonic mean on all three datasets under both GH and SPE. Under SPE, the gains in $\texttt{H}_{\mathrm{SPE}}$ are 1.76, 2.59, and 5.62 points on CUB, FGVC-Aircraft, and Stanford Cars, respectively. Adding NDL to RBL further increases $\texttt{New}_{\mathrm{SPE}}$ by 2.83, 4.38, and 0.46 points, respectively, with corresponding gains of 1.51, 1.33, and 0.65 points in $\texttt{H}_{\mathrm{SPE}}$. The same improvement in new accuracy and harmonic mean is observed under GH. These results show that RBL improves the old--new balance, while NDL provides an additional benefit in terms of new-category recognition and overall performance. 

\begin{table*}[!t]
    \centering
    \footnotesize
    \caption{Component ablation of MMO with SimGCD--CLIP on three fine-grained benchmarks. Orange New/$H$ cells indicate gains from the baseline. Best/second-best scores are bold/underlined.}
    \label{tab:mmo_component_ablation}
    \setlength{\tabcolsep}{3.5pt}
    \renewcommand{\arraystretch}{1.20}
    \begin{tabular}{@{}llcc*{8}{c}@{}}
        \toprule
        \multirow{2}{*}{Dataset} & \multirow{2}{*}{Setting} & \multirow{2}{*}{RBL} & \multirow{2}{*}{NDL}
        & \multicolumn{4}{c}{GH} & \multicolumn{4}{c}{SPE} \\
        \cmidrule(lr){5-8}\cmidrule(l){9-12}
        & & & & $\texttt{All}_{\mathrm{GH}}$ & $\texttt{Old}_{\mathrm{GH}}$ & $\texttt{New}_{\mathrm{GH}}$ & $\texttt{H}_{\mathrm{GH}}$
        & $\texttt{All}_{\mathrm{SPE}}$ & $\texttt{Old}_{\mathrm{SPE}}$ & $\texttt{New}_{\mathrm{SPE}}$ & $\texttt{H}_{\mathrm{SPE}}$ \\
        \midrule
        \multirow{3}{*}{CUB} & Baseline & -- & -- & 68.53 & 75.38 & 65.10 & 69.86 & 68.31 & \textbf{76.12} & 64.40 & 69.77 \\
        & +RBL & \cmark & -- & \underline{70.55} & \underline{76.38} & \AppendixShade{\underline{67.63}}{2.53} & \AppendixShade{\underline{71.74}}{1.88} & \underline{70.40} & \underline{75.85} & \AppendixShade{\underline{67.67}}{3.27} & \AppendixShade{\underline{71.53}}{1.76} \\
        & +RBL+NDL & \cmark & \cmark & \textbf{72.69} & \textbf{76.72} & \AppendixShade{\textbf{70.67}}{5.57} & \AppendixShade{\textbf{73.57}}{3.71} & \textbf{72.26} & 75.78 & \AppendixShade{\textbf{70.50}}{6.10} & \AppendixShade{\textbf{73.04}}{3.27} \\
        \midrule
        \multirow{3}{*}{FGVC-Aircraft} & Baseline & -- & -- & 55.59 & 59.54 & 53.61 & 56.42 & 55.47 & 59.54 & 53.43 & 56.32 \\
        & +RBL & \cmark & -- & \underline{57.79} & \textbf{63.99} & \AppendixShade{\underline{54.69}}{1.08} & \AppendixShade{\underline{58.98}}{2.56} & \underline{57.57} & \textbf{65.07} & \AppendixShade{\underline{53.82}}{0.39} & \AppendixShade{\underline{58.91}}{2.59} \\
        & +RBL+NDL & \cmark & \cmark & \textbf{60.41} & \underline{60.92} & \AppendixShade{\textbf{60.15}}{6.54} & \AppendixShade{\textbf{60.53}}{4.11} & \textbf{59.61} & \underline{62.42} & \AppendixShade{\textbf{58.20}}{4.77} & \AppendixShade{\textbf{60.24}}{3.92} \\
        \midrule
        \multirow{3}{*}{Stanford Cars} & Baseline & -- & -- & 70.46 & \textbf{83.71} & 64.06 & 72.58 & 69.03 & \textbf{89.01} & 59.38 & 71.24 \\
        & +RBL & \cmark & -- & \underline{75.50} & 82.56 & \AppendixShade{\underline{72.10}}{8.04} & \AppendixShade{\underline{76.98}}{4.40} & \underline{75.23} & 83.36 & \AppendixShade{\underline{71.30}}{11.92} & \AppendixShade{\underline{76.86}}{5.62} \\
        & +RBL+NDL & \cmark & \cmark & \textbf{76.11} & \underline{83.46} & \AppendixShade{\textbf{72.56}}{8.50} & \AppendixShade{\textbf{77.63}}{5.05} & \textbf{75.83} & \underline{84.26} & \AppendixShade{\textbf{71.76}}{12.38} & \AppendixShade{\textbf{77.51}}{6.27} \\
        \bottomrule
    \end{tabular}
\end{table*}

\subsection{Additional Results of the t-SNE Embedding Visualization (RQ2)}
\label{app:tsne_additional}

Figs.\ref{fig:simgcd_tsne}--\ref{fig:tar_tsne} provide the complete baseline and +Ours visualizations for SimGCD, GET, and GET-TAR on Herbarium 19. Each comparison uses the same ten selected categories and the same color coding, with the baseline shown on the left and its +Ours counterpart on the right. The circles represent samples and the stars represent class prototypes. Because t-SNE is a nonlinear two-dimensional projection, these plots are used only for qualitative inspection; the distances defined below are computed in the original high-dimensional embedding space.

Across the three methods, some baseline categories already form compact clusters despite low $\texttt{New}_{\mathrm{SPE}}$ scores, showing that visual compactness alone does not rule out semantic mismapping. With MMO, clusters generally become tighter and less mixed, while Intra decreases and Inter increases. The accuracy gains are much larger than the visible changes, suggesting that MMO improves feature separation and, more importantly, output--category alignment.

\textbf{Intra-class distance.}
Let $K$ denote the number of selected categories, let category $c$ contain $n_c$ samples, and let $\boldsymbol{x}_{c,i}\in\mathbb{R}^{D}$ be the high-dimensional embedding of its $i$-th sample. We first compute the centroid of category $c$ as
\begin{equation}
    \boldsymbol{\mu}_c
    =\frac{1}{n_c}\sum_{i=1}^{n_c}\boldsymbol{x}_{c,i}.
\end{equation}
The mean distance from the samples of this category to their centroid is
\begin{equation}
    d_{\mathrm{intra}}^{(c)}
    =\frac{1}{n_c}\sum_{i=1}^{n_c}
    \left\lVert\boldsymbol{x}_{c,i}-\boldsymbol{\mu}_c\right\rVert_2,
\end{equation}
and the reported class-balanced statistic is
\begin{equation}
    \mathrm{Intra}
    =\frac{1}{K}\sum_{c=1}^{K}d_{\mathrm{intra}}^{(c)}.
    \label{eq:tsne_intra}
\end{equation}
A smaller Intra value indicates that samples from the same category form a more compact cluster. We average within each category before averaging across categories, so every selected category has the same weight. This is particularly important for the long-tailed Herbarium 19 benchmark, where otherwise a few frequent categories could dominate the statistic.

\textbf{Inter-class distance.}
Using the same high-dimensional centroids, we compute the mean Euclidean distance between distinct categories, counting each combination once:
\begin{equation}
    \mathrm{Inter}
    =\frac{2}{K(K-1)}
    \sum_{1\leq c<d\leq K}
    \left\lVert\boldsymbol{\mu}_c-\boldsymbol{\mu}_d\right\rVert_2.
    \label{eq:tsne_inter}
\end{equation}
For the ten visualized categories, this is the average of $\binom{10}{2}=45$ distances between distinct centroids. A larger Inter value indicates better separation between categories. Accordingly, the desired directions are $\mathrm{Intra}\!\downarrow$ and $\mathrm{Inter}\!\uparrow$. The two measures are complementary: compact clusters can still have nearby centroids, while well-separated centroids do not by themselves guarantee small within-class dispersion.

\begin{figure*}[!htbp]
    \centering
    \includegraphics[width=0.70\linewidth]{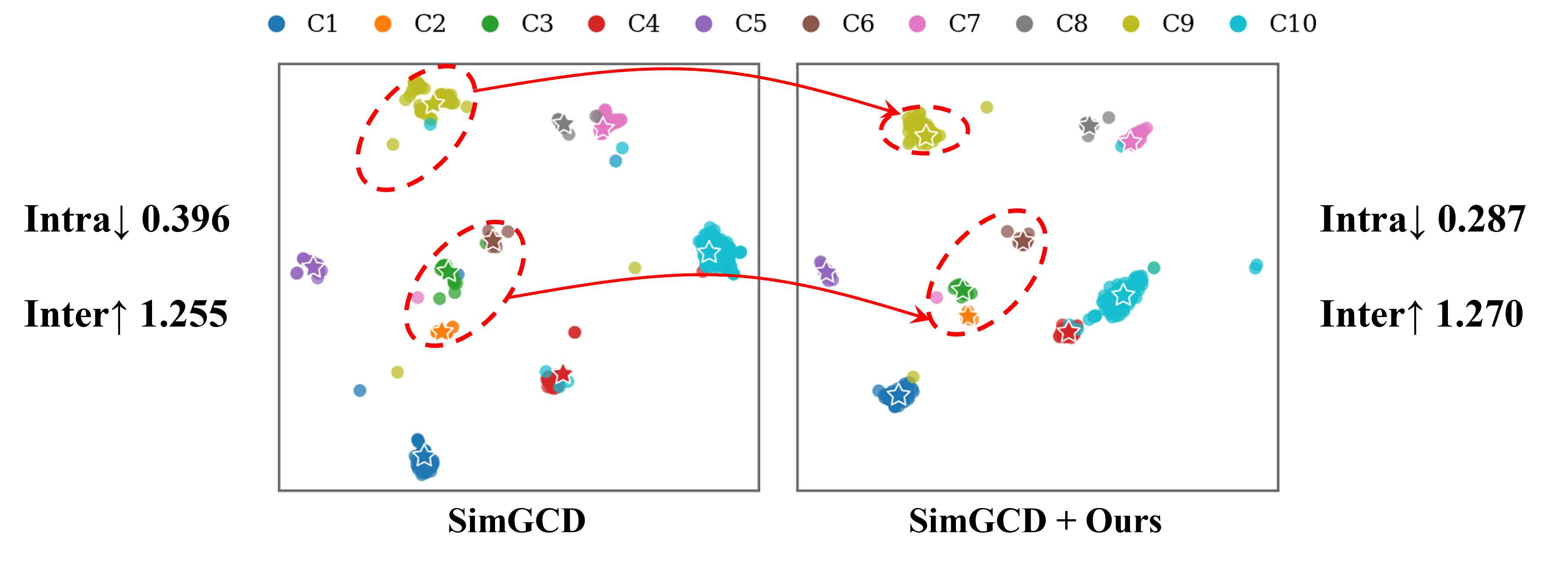}
    \caption{Comparative t-SNE visualization on Herbarium 19 for SimGCD (left) and SimGCD\textbf{+Ours} (right). Colors identify the same ten selected categories, circles denote samples, and stars denote class prototypes.}
    \label{fig:simgcd_tsne}
    \par\medskip
    \includegraphics[width=0.70\linewidth]{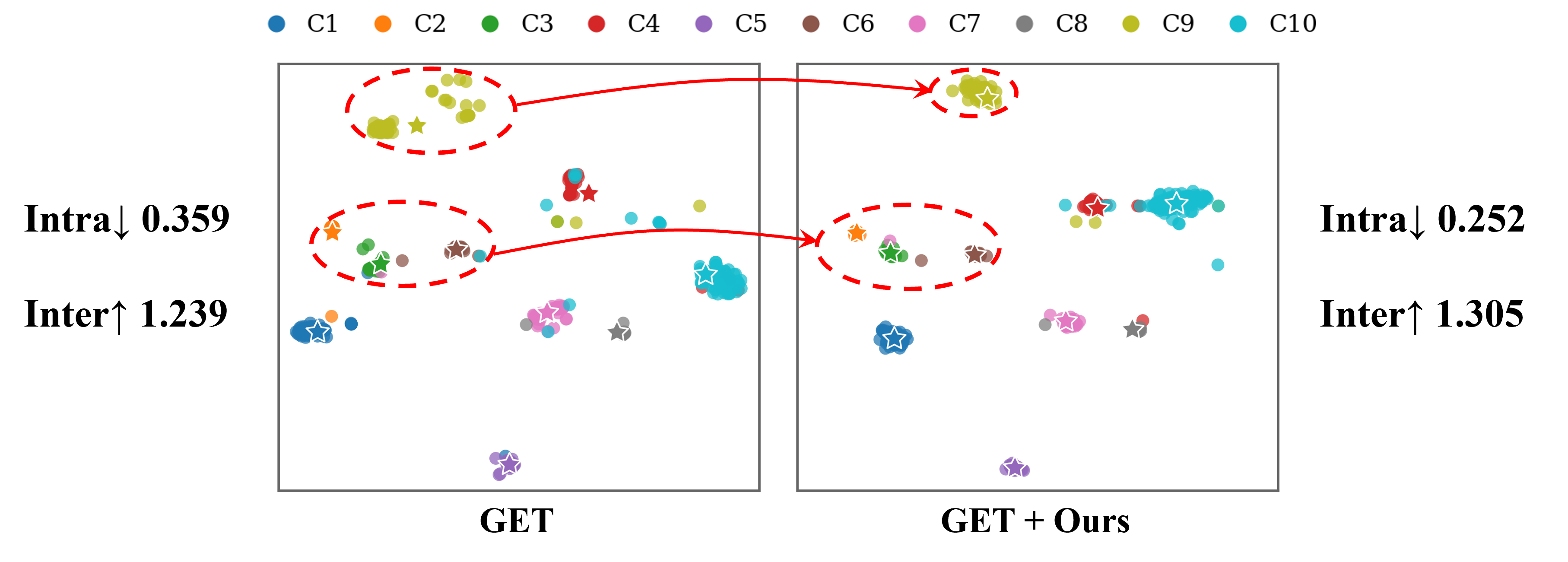}
    \caption{Comparative t-SNE visualization on Herbarium 19 for GET (left) and GET\textbf{+Ours} (right). Several baseline categories are already locally compact despite the low $\texttt{New}_{\mathrm{SPE}}$ score, while \textbf{+Ours} further tightens representative clusters.}
    \label{fig:get_tsne}
    \par\medskip
    \includegraphics[width=0.70\linewidth]{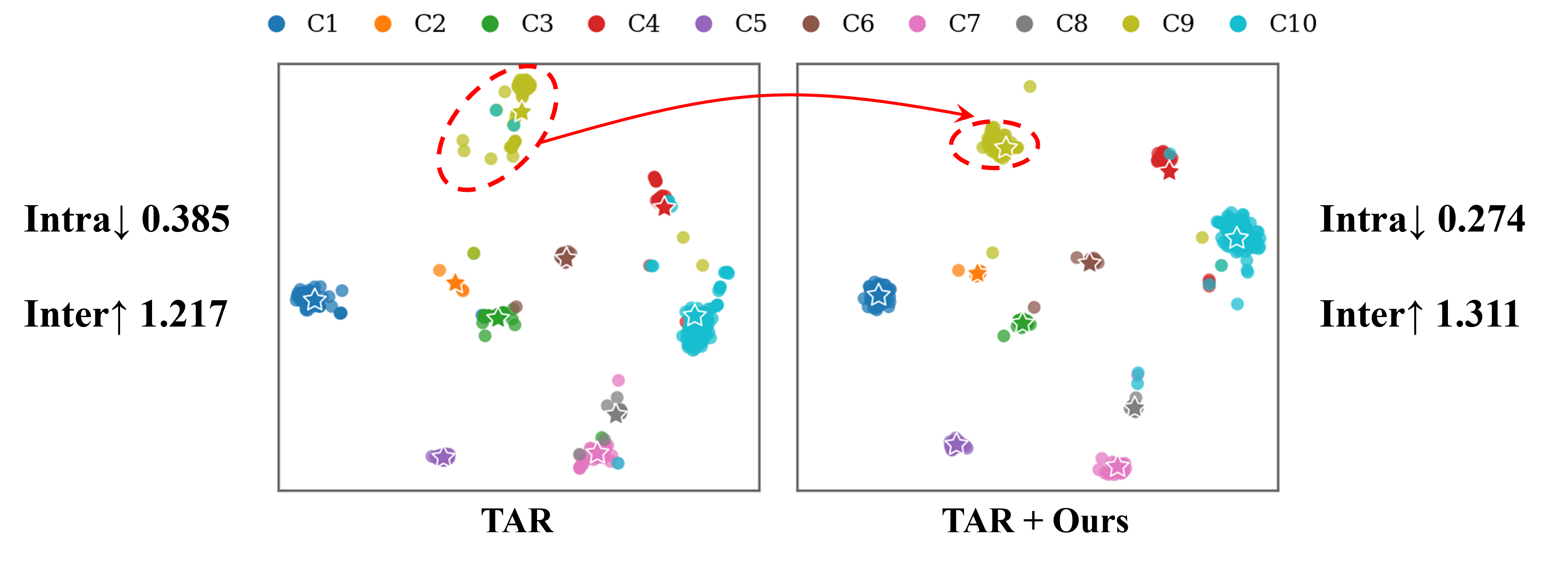}
    \caption{Comparative t-SNE visualization on Herbarium 19 for GET-TAR (left) and GET-TAR\textbf{+Ours} (right). The relatively modest visual change compared with the large $\texttt{New}_{\mathrm{SPE}}$ improvement is consistent with MMO mitigating prediction errors hidden by semantic mismapping.}
    \label{fig:tar_tsne}
\end{figure*}

\ifdefined\IncludeFullOptimizationTables
\subsection{Detailed Results with Estimated Category Numbers (RQ3)}
\label{app:detailed_estimated_cardinality}

Following Vaze et al.~\cite{DBLP:conf/cvpr/VazeHVZ22}, we estimate the total category number $\widehat C$ and derive $\widehat C_{\mathrm{new}}=\widehat C-C_{\mathrm{old}}$. Tab.\ref{tab:appendix_estimated_cardinality_full} reports the complete GH and SPE metrics under these estimates. The notation $C\!\rightarrow\!\widehat C$ gives the ground-truth and estimated total category numbers. Semantic mismapping persists when the category number is estimated. For SimGCD--CLIP on Stanford Cars, the baseline GH--SPE gap in new-category accuracy is $3.91$ points. MMO reduces this gap to $0.48$ points and improves $\texttt{New}_{\mathrm{SPE}}$, $\texttt{All}_{\mathrm{SPE}}$, and $\texttt{H}_{\mathrm{SPE}}$ by $4.99$, $3.71$, and $3.57$ points, respectively. MMO also improves new-category accuracy and the harmonic mean for both DINO configurations. Overall, the gains under estimated category numbers support the robustness and generalizability of MMO beyond the known-cardinality setting.

\begin{table}[H]
    \centering
    \footnotesize
    \caption{Complete results with category numbers estimated following Vaze et al.~\cite{DBLP:conf/cvpr/VazeHVZ22}. Colored New/$H$ cells show changes from the baseline (orange: gains; blue: decreases).}
    \label{tab:appendix_estimated_cardinality_full}
    \setlength{\tabcolsep}{2.0pt}
    \renewcommand{\arraystretch}{1.05}
    \fitTable{%
    \begin{tabular}{@{}lllrrrrrrrrr@{}}
        \toprule
        \multirow{2}{*}{Dataset ($C\!\rightarrow\!\widehat C$)}
        & \multirow{2}{*}{Method} & \multirow{2}{*}{Backbone}
        & \multicolumn{3}{c}{GH}
        & \multicolumn{4}{c}{SPE}
        & \multicolumn{2}{c}{Difference} \\
        \cmidrule(lr){4-6}\cmidrule(lr){7-10}\cmidrule(lr){11-12}
        & & & $\texttt{All}_{\mathrm{GH}}$ & $\texttt{Old}_{\mathrm{GH}}$ & $\texttt{New}_{\mathrm{GH}}$
        & $\texttt{All}_{\mathrm{SPE}}$ & $\texttt{Old}_{\mathrm{SPE}}$ & $\texttt{New}_{\mathrm{SPE}}$ & $\texttt{H}_{\mathrm{SPE}}$
        & $\Delta_{\mathrm{new}}$ & $\Delta_{\mathrm{all}}$ \\
        \midrule
        & SimGCD & DINO & 61.68 & 64.64 & 60.19 & 61.41 & 63.11 & 60.56 & 61.81 & \textbf{-0.37} & 0.27 \\
        \rowcolor{black!6}\cellcolor{white}\multirow{-2}{*}{CUB ($200\!\rightarrow\!231$)}& \quad\textbf{+ Ours} & DINO & \textbf{63.03} & \textbf{68.18} & \AppendixShade{\textbf{60.46}}{0.27} & \textbf{62.88} & \textbf{66.64} & \AppendixShade{\textbf{60.99}}{0.43} & \AppendixShade{\textbf{63.69}}{1.88} & -0.53 & \textbf{0.15} \\
        \cmidrule(lr){2-12}
        & SimGCD & CLIP & 70.40 & \textbf{75.98} & 67.60 & \textbf{70.26} & \textbf{76.32} & 67.23 & \textbf{71.49} & \textbf{0.37} & \textbf{0.13} \\
        \rowcolor{black!6}\cellcolor{white}\multirow{-2}{*}{CUB ($200\!\rightarrow\!212$)}& \quad\textbf{+ Ours} & CLIP & \textbf{70.53} & 74.45 & \AppendixShade{\textbf{68.57}}{0.97} & 69.82 & 74.92 & \AppendixShade{\textbf{67.27}}{0.04} & \AppendixShade{70.89}{-0.60} & 1.30 & 0.71 \\
        \midrule
        & SimGCD & DINO & 52.88 & \textbf{70.16} & 44.53 & 52.47 & \textbf{72.91} & 42.60 & 53.78 & 1.93 & 0.41 \\
        \rowcolor{black!6}\cellcolor{white}\multirow{-2}{*}{Stanford Cars ($196\!\rightarrow\!230$)}& \quad\textbf{+ Ours} & DINO & \textbf{53.65} & 68.07 & \AppendixShade{\textbf{46.68}}{2.15} & \textbf{53.37} & 68.97 & \AppendixShade{\textbf{45.84}}{3.24} & \AppendixShade{\textbf{55.07}}{1.29} & \textbf{0.84} & \textbf{0.28} \\
        \cmidrule(lr){2-12}
        & SimGCD & CLIP & 73.57 & 82.41 & 69.30 & 72.51 & 87.26 & 65.39 & 74.76 & 3.91 & 1.06 \\
        \rowcolor{black!6}\cellcolor{white}\multirow{-2}{*}{Stanford Cars ($196\!\rightarrow\!234$)}& \quad\textbf{+ Ours} & CLIP & \textbf{76.27} & \textbf{87.46} & \AppendixShade{\textbf{70.87}}{1.57} & \textbf{76.22} & \textbf{88.31} & \AppendixShade{\textbf{70.38}}{4.99} & \AppendixShade{\textbf{78.33}}{3.57} & \textbf{0.48} & \textbf{0.05} \\
        \bottomrule
    \end{tabular}}
\end{table}
\fi

\FloatBarrier

\endgroup
\flushbottom

\section{Implementation Details of Region-Balance Loss}
\label{app:rbl_implementation}
\subsection{Minibatch Tolerance and Numerical Stability}
The main text defines RBL using the fixed regional prior $p_r$. The tolerance-based variant used in our experiments accommodates differences between minibatches by replacing $p_r$ with an effective target $p_r^{(\delta)}$:
\begin{equation}
    \begin{aligned}
        \mathcal{I}
        &= [p_{\mathrm{old}}-\delta,p_{\mathrm{old}}+\delta]
        \cap[\epsilon,1-\epsilon], \\
        p_{\mathrm{old}}^{(\delta)}
        &=\operatorname{clip}(\bar q^{\,\mathrm{old}},\mathcal{I}),\qquad
        p_{\mathrm{new}}^{(\delta)}=1-p_{\mathrm{old}}^{(\delta)}, \\
        \mathcal{L}_{\mathrm{RB}}^{(\delta)}
        &=-\sum_{r\in\{\mathrm{old},\mathrm{new}\}}
        p_r^{(\delta)}\log\bar q^{\,r}.
    \end{aligned}
    \label{eq:rbl_effective_target}
\end{equation}
The effective target is treated as constant during backpropagation. This variant applies no corrective gradient when the batch-average regional probabilities lie inside the allowed interval and pushes them toward the nearest boundary otherwise. The underlying prior $p_r$ remains fixed; the effective target does not estimate or correct that prior. Setting $\delta=0$ recovers the fixed-prior loss, subject to numerical clipping.

For numerical stability, $p_{\mathrm{old}}$ is clipped to $[\epsilon,1-\epsilon]$ before constructing the interval, and $p_{\mathrm{new}}=1-p_{\mathrm{old}}$. Probabilities inside logarithms are also clipped to this range, with $\epsilon=10^{-6}$. We set the tolerance to $\delta=0.05$ by default.

\subsection{Estimating an Unknown Labeling Fraction}
\label{app:rbl_prior_estimation}
On imbalanced benchmarks, the old- and new-category sample proportions cannot be inferred from category counts alone. When the labeling fraction $\rho$ is unknown, we estimate it with TIcE~\cite{BekkerDavis2018TIcE} and use the estimate to obtain the old-region sample prior required by RBL. TIcE uses frozen features from $\mathcal{D}_l\cup\mathcal{D}_u$ and the observed labeled/unlabeled status, without using hidden category labels.

Let $S=1$ indicate a labeled sample, $Z=1$ an old-category sample, and $X$ its frozen feature representation. Assuming that old-category samples are selected for labeling independently of $X$ with probability $\rho$, any feature-defined subset $A$ satisfies
\begin{equation}
    \Pr(S=1\mid X\in A)
    =\rho\Pr(Z=1\mid X\in A)\leq\rho.
\end{equation}
The labeled fraction therefore approaches $\rho$ in subsets dominated by old-category samples. TIcE searches for such subsets with decision trees. It uses separate folds to select subsets and estimate their labeled fractions, avoiding evaluation on the same samples used to build the tree. For a subset $A$ with $L_A$ labeled samples among $T_A$ samples, it computes the uncertainty-adjusted score
\begin{equation}
    b_A=\frac{L_A}{T_A}
    -\sqrt{\frac{\rho_0(1-\rho_0)(1-\eta)}{\eta T_A}},
    \qquad
    \eta=\max\left\{0.025,\frac{1}{1+0.004T_{\mathrm{est}}}\right\},
\end{equation}
where $T_{\mathrm{est}}$ is the estimation-fold sample count. Starting with $\rho_0=0.5$, TIcE takes the maximum of the subset scores and the overall labeled fraction in each estimation fold, then averages these estimates across folds. It repeats this step once with the updated $\rho_0$. We substitute the resulting $\widehat\rho_{\mathrm{TIcE}}$ into Eq.\eqref{eq:pao_region_prior} to obtain $\widehat p_{\mathrm{old}}$.

\end{document}